\documentclass[12pt]{article}
\usepackage{amsmath,amsthm,amssymb,bm,mathrsfs}
\usepackage{graphicx}
\usepackage[dvipsnames]{xcolor}
\usepackage[ruled,linesnumbered]{algorithm2e}

\usepackage{caption,enumerate,listings}
\usepackage{setspace}
\usepackage[OT1]{fontenc}
\usepackage[colorlinks=true,
            linkcolor=blue,
            citecolor=blue,
            urlcolor=blue]{hyperref}
\usepackage{pifont}
\usepackage[margin=1in]{geometry}
\usepackage[protrusion=false,expansion=true]{microtype}
\usepackage[title]{appendix}
\usepackage{bbm}
\usepackage[authoryear,round]{natbib}
\usepackage{amsmath}
\usepackage{bibunits}
\usepackage{booktabs}

\makeatletter
\renewcommand\section{\@startsection{section}{1}{\z@}%
  {-1.5ex \@plus -0.5ex \@minus -.2ex}% 上间距
  {0.3ex \@plus 0.2ex \@minus .2ex}% 下间距
  {\normalfont\Large\bfseries}}
\makeatother
\usepackage{extpfeil}
\usepackage{subcaption}
\DeclareMathOperator{\csch}{csch}
\usepackage{comment}
\usepackage{enumitem}
\usepackage{multirow}
\def\spacingset#1{\renewcommand{\baselinestretch}%
{#1}\small\normalsize} \spacingset{1}
\newcommand{\argmin}{\mathop{\mathrm{argmin}}}
\newcommand{\argmax}{\mathop{\mathrm{argmax}}}

\newcommand{\sign}{\mathop{\mathrm{sign}}}

\newtheorem{lemma}{{\bf Lemma}}
\newtheorem{theorem}{{\bf Theorem}}
\newtheorem{definition}{{\bf Definition}}
\usepackage{setspace}
\usepackage{booktabs}
\theoremstyle{plain}
\usepackage{algorithm}
\usepackage{algorithmic}
\usepackage{titlesec}
\titleformat{\section}[block]{\normalfont\large\bfseries}{\thesection}{1em}{}

\begin{document}

\title{\Large \bf 
When Less Is More: Binary Feedback Can Outperform Ordinal Comparisons in Ranking Recovery
}

 \author{
 Shirong Xu\thanks{Department of Statistics and Data Science, School of Economics, Xiamen University, China}
 \thanks{Wang Yanan Institute for Studies in Economics, Xiamen University, China}, 
 Jingnan Zhang\thanks{Business for Science \& Technology, School of Management, University of Science and Technology of China}, 
 and Junhui Wang\thanks{Department of Statistics, The Chinese University of Hong Kong}
 }

\date{}
\maketitle

\begin{abstract}
Paired comparison data, where users evaluate items in pairs, play a central role in ranking and preference learning tasks. While ordinal comparison data intuitively offer richer information than binary comparisons, this paper challenges that conventional wisdom. We propose a general parametric framework for modeling ordinal paired comparisons without ties. The model adopts a generalized additive structure, featuring a link function that quantifies the preference difference between two items and a pattern function that governs the distribution over ordinal response levels. This framework encompasses classical binary comparison models as special cases, by treating binary responses as binarized versions of ordinal data. Within this framework, we show that binarizing ordinal data can significantly improve the accuracy of ranking recovery. Specifically, we prove that under the counting algorithm, the ranking error associated with binary comparisons exhibits a faster exponential convergence rate than that of ordinal data. Furthermore, we characterize a substantial performance gap between binary and ordinal data in terms of a signal-to-noise ratio (SNR) determined by the pattern function. We identify the pattern function that minimizes the SNR and maximizes the benefit of binarization. Extensive simulations and a real application on the MovieLens dataset further corroborate our theoretical findings.
\end{abstract}

\begin{bibunit}[apalike]
\spacingset{1.7} 
\newpage
\section{Introduction}
\label{Sec:Intro}
Paired comparison data arises from evaluating items in pairs and is commonly encountered across various scenarios, including college comparisons \citep{caron2014bayesian}, product evaluations \citep{bockenholt1997some,duineveld2000log}, sports tournaments \citep{buhlmann1963pairwise,li2022detecting,jadbabaie2020estimation}, and human feedback in large language models \citep{zhu2023principled,poddar2024personalizing}. Usually, it is commonly assumed that a true parameter vector $\bm{\theta}^\star=(\theta_1^\star,\ldots,\theta_n^\star)$ exists for $n$ items with $\theta_i^\star$ representing the preference of $i$-th item. A central problem resolving around paired comparison data is to estimate the ordering of $\bm{\theta}^\star$, and then a complete ranking of $n$ items can be obtained \citep{chen2019spectral,chen2022optimal,chen2022partial,wauthier2013efficient}. 

Throughout the past century, the literature has seen extensive research dedicated to the development of parametric paired comparison models. One prominent line of work focuses on modeling comparisons as \textit{binary} outcomes. Specifically, the probability that item $i$ is preferred over item $j$ under a given criterion is modeled as
\begin{align}
\label{BinaryModel}
\mathbb{P}\left( i \succ j \right) = F(\theta_i^\star - \theta_j^\star),
\end{align}
where $\succ$ denotes a ranking relationship under a specific criterion, such as preference. For instance, in the context of large language models, users may be asked to express a binary preference when comparing two textual responses. Similarly, in sports tournaments, two teams compete, and the outcome of the match reflects which team is stronger.

The function $F(\cdot)$ can take various forms, such as the logistic function, $F(x) = (1 + \exp(-x))^{-1}$, or the normal cumulative distribution function. These correspond to the Bradley-Terry-Luce (BTL) model \citep{bradley1952rank} and the Thurstone-Mosteller (TM) model \citep{thurstone1994law}, respectively. Furthermore, \citet{stern1990continuum} introduced a model in which binary comparisons arise from the comparison of two gamma-distributed random variables with different scale parameters. This framework includes the BTL and Thurstone-Mosteller models as special cases for different values of the gamma shape parameter.

Another line of research aims at modeling non-binary ordinal paired comparisons. This line of research is motivated by scenarios in which items are evaluated with varying degrees of preference. For example, consumers may express a strong preference for one product over another when comparing alternatives. One of the earliest contributions in this area extends the BTL model to account for ties in paired comparisons, effectively incorporating three distinct levels of preference \citep{glenn1960ties, rao1967ties, davidson1970extending}. Building on this, \citet{agresti1992analysis} introduces an adjacent-categories logit model that accommodates comparisons with more than three options, while naturally reducing to the BTL model when only two categories are present. Further extensions to continuous paired comparison data are presented in \citet{stern2011moderated} and \citet{han2022general}. Notably, \citet{han2022general} proposes a general paired comparison framework capable of modeling both continuous and ordinal observations.

\begin{figure}[h]
    \centering
    \includegraphics[scale=0.265]{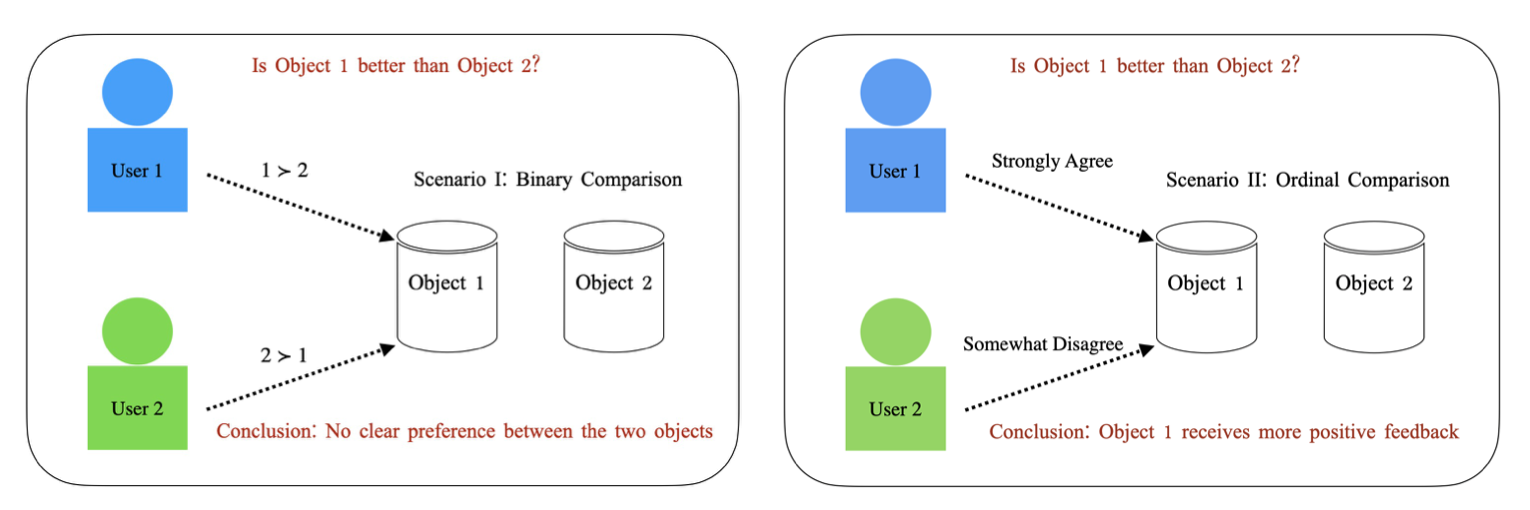}
    \caption{Two users submit pairwise comparison responses using two distinct sets of system-defined options. In Scenario I (Left), a preference ranking tie occurs. In Scenario II (Right), object 1 receives more favorable feedback.}
    \label{fig:Example}
\end{figure}

To clarify the distinction between binary and ordinal comparisons, we present an example in Figure~\ref{fig:Example}. In this example, the same two users are asked to compare the same pair of items under two differently structured systems. In the first system (Left), users provide binary responses to their comparisons, while in the second system (Right), they offer more detailed, ordinal feedback. In the first case, the preference ordering between the items is ambiguous because the two users give conflicting responses. In contrast, under the second system, item 1 receives more favorable feedback: user 1 expresses a strong preference for item 1, whereas user 2 only slightly prefers item 2. This example illustrates that when a system restricts users to binary choices, valuable information about the strength of preferences may be lost. In other words, the responses in Scenario I are binarized versions of those in Scenario II. Specifically, \emph{strongly agree} and \emph{somewhat disagree} are reduced to $1 \succ 2$ and $2 \succ 1$, respectively, resulting in significant information loss. 

The example in Figure~\ref{fig:Example} naturally gives rise to \textit{the intuition} that ordinal comparison data conveys more information about the underlying ground-truth ranking. Consequently, one might expect that using ordinal comparisons would lead to more accurate ranking estimates of items. However, in this paper, we theoretically demonstrate that this intuition does not hold within a broad class of ordinal comparison models in the asymptotic regime. In fact, we show that binarizing ordinal comparison data into binary comparisons can significantly improve the estimation of the ground-truth ranking.

To address the question of which type of comparison data is more effective for recovering item rankings, it is crucial to understand the underlying relationship between ordinal and binary comparison data. Specifically, as illustrated in Example~\ref{fig:Example}, when two different systems elicit responses on the same items from the same group of users, what is the fundamental connection between their binary and ordinal comparison responses? To some extent, binary responses can be viewed as the result of binarizing the corresponding ordinal responses. This raises a natural question: Is there a modeling framework that explicitly captures the binarization process?

To address the above questions, we begin by introducing a general class of parametric models designed to capture ordinal outcomes in pairwise comparisons without ties. This focus is motivated by the fact that ties are often absent in many real-world comparison settings, such as sports competitions and consumer preference surveys. The proposed model takes a generalized additive form, consisting of a link function that captures the preference difference between two items and a pattern function that characterizes the distribution over ordinal response levels. The contributions of this paper within the proposed framework can be summarized as follows:
\begin{itemize}
    \item[(1)] A key advantage of this framework is that it subsumes a broad family of binary comparison models, including those in~(\ref{BinaryModel}), as special cases. In particular, when ordinal comparison data are binarized, the resulting binary responses follow the model in (\ref{BinaryModel}). This connection enables a principled comparison of the ranking recovery performance between binary and ordinal comparison data.
    \item[(2)] { Within this framework, we theoretically demonstrate that binarizing ordinal comparison data can accelerate the convergence rate in recovering the ground-truth ranking using the counting algorithm \citep{busa2013top,shah2018simple}, which has been shown to be more robust and computationally efficient than maximum likelihood estimation \citep{shah2018simple}. Specifically, we demonstrate that the ranking error associated with binary comparison data exhibits a faster exponential convergence rate. This result implies that, provided a sufficiently large number of users contribute comparison data, binary comparisons consistently outperform ordinal ones in terms of ranking accuracy. This result offers valuable insight into the importance of binary comparison data, particularly given its widespread use in preference learning to enhance the performance of large language models \citep{zhu2023principled,slocum2025diverse}.}
    
    \item[(3)] We also establish the existence of a nontrivial gap in ranking error between binary and ordinal comparison data. This performance gap is governed by the signal-to-noise ratio (SNR) associated with the pattern function: the smaller the SNR, the greater the benefit of binarizing ordinal data. Furthermore, we characterize the pattern function that minimizes the SNR, thereby identifying the setting in which binarization yields the greatest improvement. This theoretical finding is further supported by extensive simulation studies, as presented in Section~\ref{SubSec:Simu}.
\end{itemize}

The remainder of this paper is organized as follows. In Section~\ref{Sec:ProModel}, we develop the proposed ordinal comparison model and provides background on the ordinal comparison graph under the proposed model. Section~\ref{Sec:OrdBetter} presents a theoretical analysis showing that binarized comparison data lead to improved performance in ranking recovery for both two-item and $n$-item ranking problems. In Section~\ref{Sec:SNR}, we identify the pattern function that minimizes the signal-to-noise ratio (SNR), thereby yielding the greatest benefit from binarizing ordinal comparison data. Section~\ref{Sec:Exp} presents extensive simulations and a real-data application to validate our theoretical findings. A brief summary is provided in Section~ \ref{Sec:Sum}. All proofs of theorems and supporting lemmas and additional discussions are provided in the supplementary file.

%\subsection{Notation}
%\label{Sec:Not}
\textbf{Notation and Definitions.} We introduce some notations used throughout the paper. For a positive integer $K$, denote $[K] = \{1, \ldots, K\}$ as the set of the first $K$ positive integers, and let $\Upsilon(K) = \{k \in \mathbb{Z} : -K \leq k \leq K\} \setminus \{0\}$. Let $\mathbb{I}(\cdot)$ denote the indicator function, where $\mathbb{I}(A) = 1$ if event $A$ is true, and 0 otherwise. For a vector $\bm{x}$, we let $\|\bm{x}\|_2$ denote its $l_2$-norm. Let $\mathrm{Geo}(\psi_{\gamma}, K)$ denote a discrete distribution, defined by $\mathbb{P}(X_\gamma = k) = \frac{e^{\psi_{\gamma}(k)}}{\sum_{j=1}^{K} e^{\psi_{\gamma}(j)}}$ for $k \in [K]$, where $X_{\gamma} \sim \mathrm{Geo}(\psi_{\gamma}, K)$ and $\psi_{\gamma}(k)$ is a discrete function depending on $\gamma$. For the random variable $X_{\gamma}$, we define its signal-to-noise ratio (SNR) as $\text{SNR}(X_{\gamma}) = \frac{[\mathbb{E}(X_{\gamma})]^2}{\text{Var}(X_{\gamma})}$. Let $\sinh(x) = \frac{e^x - e^{-x}}{2}$, $\cosh(x) = \frac{e^x + e^{-x}}{2}$, $\tanh(x) = \frac{e^x - e^{-x}}{e^x + e^{-x}}$, and $\csch(x)=\frac{2}{e^x-e^{-x}}=\frac{1}{\sinh(x)}$ denote the hyperbolic sine, cosine, tangent, and cosecant functions, respectively.

%For a random variable $X_n$ and a positive sequence $\{a_n\}_{n=1}^\infty$, $X_n = o_p(a_n)$ means $X_n / a_n \to 0$ in probability, and $X_n = O_p(a_n)$ means $X_n / a_n$ is stochastically bounded. 
\section{Proposed Method}
\label{Sec:ProModel}
In this section, we introduce a general ordinal comparison model for analyzing paired comparisons represented by symmetric, nonzero discrete values. Such comparisons frequently arise in contexts such as sports tournaments and consumer surveys. We then investigate the properties of the proposed model and highlight its connections to binary comparison models, including the Bradley–Terry–Luce (BTL) model \citep{bradley1952rank} and the Thurstone–Mosteller (TM) model \citep{thurstone1994law}, under specific choices of link functions.

\subsection{Strength Link Function}
\label{SecSub:Link}
We begin with presenting the definition of the strength link function, which serves as a fundamental building block, and allows for a range of adaptations within the framework.
\begin{definition}[Strength Link Function]
A function $\phi$ is a strength link function if $\phi$ satisfies the following properties:
\begin{itemize}
\item[(1)] Increasing Monotonicity: $\phi(x)>\phi(y)$ if $x>y$;
\item[(2)] Origin Symmetry: $\phi(x)=-\phi(-x)$ for any $x \in \mathbb{R}$.
\end{itemize}
\end{definition}

\begin{figure}[ht!]
\centering
\includegraphics[scale=0.65]{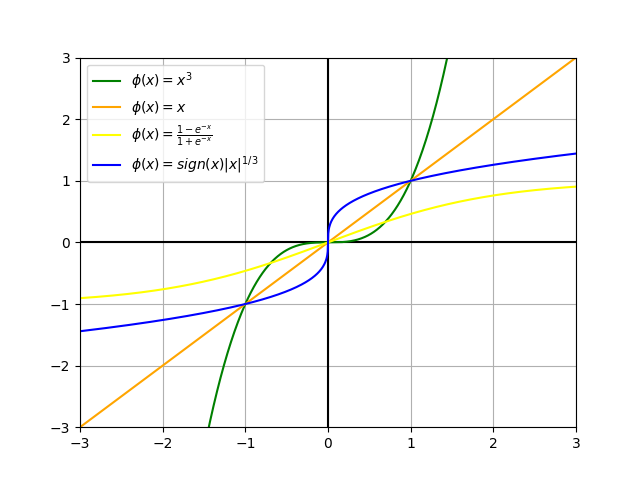}
\caption{Four examples for $\phi(x)$: (1) $\phi(x)=x^3$; (2) $\phi(x)=x$; (3) $\phi(x)=\frac{1-e^{-x}}{1+e^{-x}}$; (4) $\phi(x)=\sign(x)|x|^{1/3}$.
}\label{Fig:Exam}
\end{figure}

According to the definition above, the conditions for $\phi$ to be qualified as a strength link function encompass both increasing monotonicity and symmetry about the origin. Under the constraints imposed on $\phi$, it is evident that $\phi(0) = 0$ and $\phi(x) > 0$ for all $x > 0$. The selection of $\phi$ is notably flexible, necessitating only symmetry about the origin and increasing monotonicity. Similar conditions have also been considered in \citet{han2022general}. Figure \ref{Fig:Exam} illustrates several examples of possible choices for $\phi$.

In addition to the examples depicted in Figure \ref{Fig:Exam}, various strength link functions can be devised by leveraging the cumulative distribution functions of various continuous random variables as demonstrated in Lemma \ref{Lemma:CDF}.

\begin{lemma}
\label{Lemma:CDF}
Let $X$ be a symmetric continuous random variable centered around zero with support on $\mathbb{R}$, and let $F(X)$ denote its associated cumulative distribution function (CDF) without point masses. Then the function $\phi(x) = C\log\left(\frac{F(x)}{1-F(x)}\right)$ is a strength link function, where $C$ is any positive constant.
\end{lemma}

Lemma~\ref{Lemma:CDF} shows that for a symmetric continuous random variable $X$, a corresponding strength link function can be expressed as $\phi(x) = C\log\left(\frac{F(x)}{1 - F(x)}\right)$, where $F(x)$ is the cumulative distribution function of $X$. This result offers a flexible approach for constructing strength link functions based on commonly used distributions, such as the logistic and standard normal distributions. Furthermore, it serves as a key element in establishing connections between the proposed comparison model and classical binary paired comparison models, including the BTL and TM models. These connections will be further explored in Section~\ref{Sec:DCG}.

\subsection{Probabilities over Preference Strength Levels}
\label{SubSec:Prob}
In this section, we aim to examine the probabilities associated with different levels of preference strength using three real-world datasets from the domains of sports tournaments, recommender systems, and large language models. Understanding whether more extreme comparison outcomes are more likely to occur is crucial for developing a practical ordinal comparison model.

The first dataset we analyze is the absolute point differences of the NBA 2023-2024 season game results. Here, the absolute point differences between competing teams were used to define multiple discrete strength levels for paired comparison modeling. Specifically, the absolute score differences were grouped into intervals representing varying degrees of margin of victory. These intervals—ranging from close games (e.g., 1 to 7 points) to large blowouts (e.g., 42 or more points)—serve as ordinal strength levels that quantify the extent of dominance by the winning team. 

The second dataset we utilize is the MovieLens 100K dataset \citep{harper2015movielens}, in which user ratings for movies are recorded on a 1–5 scale. For each individual user, we compute the pairwise differences between the ratings of all movies they have rated, treating these non-zero differences as indicators of relative preference strength between items. Aggregating over all users, we regard these differences as instances of ordinal pairwise comparisons. To analyze the structure of such comparisons, we further take the absolute values of the rating differences and construct an empirical distribution, which reflects the frequency of different levels of preference intensity observed in the dataset. 

The third dataset we consider is the UltraFeedback dataset \citep{cui2023ultrafeedback}. In this dataset, the authors employed GPT-4 to assign 5-point ratings across multiple aspects to different answers generated by various LLMs for the same prompt. For each prompt, we compute the rating difference between two answers and treat the result as ordinal comparison data. The rating differences range from 0.25 to 4 and are categorized into five equally spaced intervals of width 0.75, each representing a distinct degree of preference strength.

\begin{figure}[h!]
    \centering
    \begin{subfigure}[b]{0.325\textwidth}
        \centering
        \includegraphics[width=\textwidth]{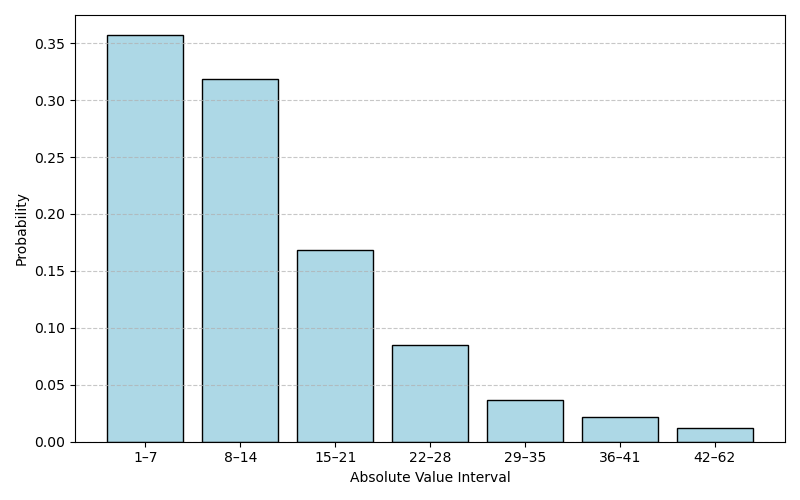}
        \caption{NBA 2023-2024 dataset}
    \end{subfigure}
    \hfill
    \begin{subfigure}[b]{0.325\textwidth}
        \centering
        \includegraphics[width=\textwidth]{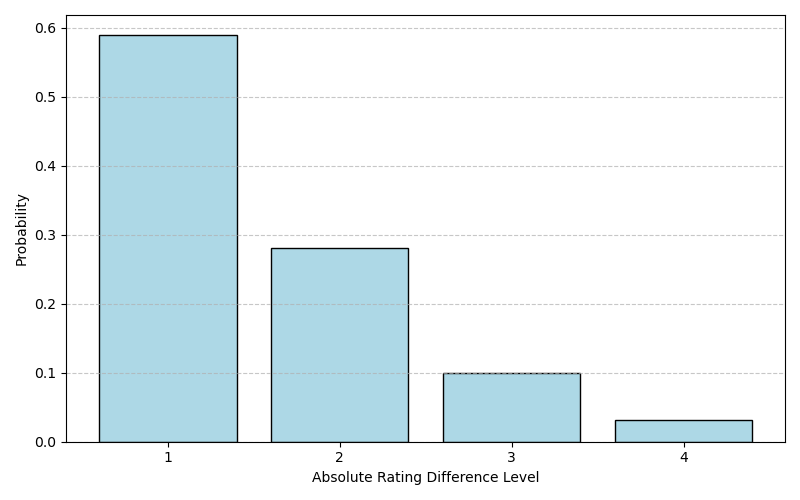}
        \caption{MovieLens 100K dataset}
    \end{subfigure}
        \begin{subfigure}[b]{0.325\textwidth}
        \centering
        \includegraphics[width=\textwidth]{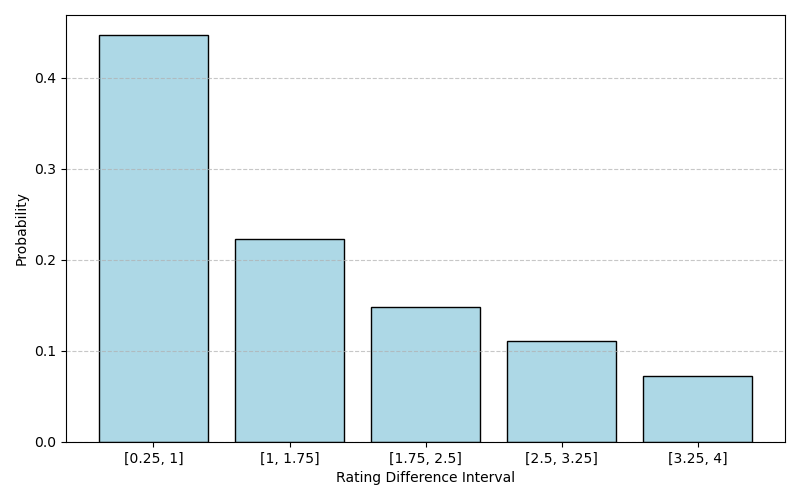}
        \caption{UltraFeedback dataset}
    \end{subfigure}
    \caption{The distributions of ordinal comparison data derived from three real datasets.}
    \label{fig:Ex1}
\end{figure}

The empirical distributions of the constructed ordinal comparison data from three datasets are presented in Figure \ref{fig:Ex1}. It is evident that more extreme comparison outcomes occur with lower probabilities. This empirical pattern suggests that a well-founded ordinal pairwise comparison model should incorporate the property that the probability of an outcome decreases as its magnitude increases. This observation serves as a key motivation for the ordinal comparison modeling framework developed in the subsequent sections.

{ 
\subsection{Ordinal Comparison Model}
In this section, we introduce a general ordinal comparison model, building upon the strength link function proposed in Section~\ref{SecSub:Link} and the observation framework discussed in Section~\ref{SubSec:Prob}.

Let $\gamma$ denote the preference difference between two items under comparison, and let $k \in \Upsilon(K)$ represent a possible ordinal outcome of the comparison. We define the propensity function $g(k\,|\,\phi,\psi_{\gamma},\gamma)$ in the generalized additive form:
\begin{align}
\label{Prop}
\textbf{Propensity Function:} \quad g(k\,|\,\phi,\psi_{\gamma},\gamma) = \phi(\operatorname{sign}(k)\gamma)+\psi_{\gamma}(k),
\end{align}
where $\psi_{\gamma}$ is an even function that modulates the influence of the ordinal outcome’s magnitude, with its specific form determined by the value of $\gamma$. The function $\psi_{\gamma}(k)$ is further introduced to capture the distributional pattern of ordinal outcomes, particularly the empirical tendency for more extreme comparison results to occur less frequently. Within the context of a comparison graph, $\psi_{\gamma}(k)$ must be an even function in $(\gamma,k)$ \citep{han2022general}. Consequently, the specification of the model in (\ref{Prop}) is identifiable, since $g(k\,|\,\phi,\psi_{\gamma},\gamma)$ decomposes into an even function of $\gamma$ and an odd function of $\gamma$, ensuring uniqueness of the model. For brevity, we will write $g(k\,|\,\phi,\psi_{\gamma},\gamma)$ simply as $g(k)$ when no confusion arises.

Let $G(\phi,\psi_{\gamma},\gamma,K)$ denote the general discrete comparison model parametrized by $\phi$, $\psi_{\gamma}$, $\gamma$, and a positive integer $K \in \mathbb{Z}^+$ specifying the range of outputs. Given a random variable $Y \sim G(\phi,\psi_{\gamma},\gamma,K)$, the probability of $Y=k$ is given as
\begin{align}
\label{Eqn:ProposedModel}
\mathbb{P}
\left(
Y=k 
\right) = &
\frac{1}{\Psi_{\phi,\psi_{\gamma}}(\gamma)}
\exp\big(
 g(k\,|\,\phi,\psi_{\gamma},\gamma)\big)
,
\end{align}
for any $k \in \Upsilon(K)$, where $\Psi_{\phi,\psi_{\gamma}}(\gamma)= \sum_{k \in \Upsilon(K)} \exp\big(g(k\,|\,\phi,\psi_{\gamma},\gamma)\big)$ is a normalizing constant. 

The model in (\ref{Eqn:ProposedModel}) is designed to capture preference differences in ordinal comparison settings, such as sports games and consumer surveys. In sports, outcomes often take symmetric, discrete values excluding zero—commonly seen in games like badminton, tennis, and football. In this framework, the parameter $\gamma$ denotes the strength difference between two teams or players, with larger values indicating greater disparity. The strength link function $\phi$ governs how $\gamma$ influences the distribution of outcomes. The value of $K$ is context-dependent. For instance, in consumer surveys, as shown in Figure \ref{fig:Example}, respondents may choose from four options: strongly agree, somewhat agree, somewhat disagree, and strongly disagree. This corresponds to the case where $K=2$.

In the following theorem, we highlight several noteworthy properties inherent in the proposed ordinal comparison model.

\begin{theorem}
\label{Thm:MeanVari}
If $Y \sim G(\phi,\psi_{\gamma},\gamma,K)$, then $Y$ possesses the following properties:
\begin{itemize}
\item[(1)] The probability of $Y$ being positive is 
$$
\mathbb{P}(Y>0) = \frac{\exp(\phi(\gamma))}{\exp(\phi(-\gamma))+\exp(\phi(\gamma))}=
\begin{cases}
    \frac{e^{\gamma}}{1+e^{\gamma}}, \text{ if }\phi(\gamma) = \frac{\gamma}{2}, \\
    \Phi(\gamma), \text{ if } \phi(\gamma) = \frac{1}{2}\log\left(\frac{\Phi(\gamma)}{1-\Phi(\gamma)}\right).
\end{cases}
$$
Particularly, if $\gamma=0$, then $\mathbb{P}(Y>0) = \mathbb{P}(Y<0)$ for any $\phi$ and $\psi_{\gamma}$.

\item[(2)] The random variable $-Y$ follows $G(\phi,\psi_{-\gamma},-\gamma,K)$.
\item[(3)] The mean and variance of $Y$ are given as
\begin{align*}
&\mathbb{E}(Y) = 
\tanh(\phi(\gamma)) \cdot \frac{\sum_{k =1 }^K k e^{\psi_{\gamma}(k)}}{\sum_{k =1 }^K e^{\psi_{\gamma}(k)}},\\
&\mathrm{Var}(Y) 
= 
\frac{\sum_{k=1}^K k^2 e^{\psi_{\gamma}(k)}}{\sum_{k=1}^K e^{\psi_{\gamma}(k)}} 
- \left(
\tanh\big(\phi(\gamma)\big) \cdot \frac{\sum_{k=1}^K k e^{\psi_{\gamma}(k)}}{\sum_{k=1}^K e^{\psi_{\gamma}(k)}}
\right)^2.
\end{align*}
The corresponding signal-to-noise ratio is given as
\begin{align*}
  \textnormal{SNR}(Y) = \frac{[\mathbb{E}(Y)]^2}{\mathrm{Var}(Y)}=
  \frac{\tanh^2(\phi(\gamma)) }{\frac{1}{\textnormal{SNR}(X_{\gamma})}+1-\tanh^2(\phi(\gamma))},
\end{align*}
where $X_{\gamma} \sim \textnormal{Geo}(\psi_{\gamma},K)$. In particular, when $K = 1$, we have $\mathrm{SNR}(X_{\gamma}) = \infty$, which implies that $\mathrm{SNR}(Y) = \sinh^2(\phi(\gamma))$.
\end{itemize}
\end{theorem}

In Theorem~\ref{Thm:MeanVari}, property (1) characterizes the probability that $Y$ is positive under the proposed model. This property is particularly important in scenarios where $Y$ represents the outcome of a comparison between two items. For example, if $Y$ denotes the score difference between teams $i$ and $j$, then $\mathbb{P}(Y > 0)$ corresponds to the probability that team $i$ defeats team $j$. Moreover, property (1) serves as a crucial bridge between the proposed model and existing models that consider only binary comparisons. Property (2) establishes the symmetry of $Y$ with respect to $\gamma$. This is especially relevant when $\gamma$ reflects the notional worth disparity between two items, indicating that reversing the sign of $\gamma$ should invert the likelihood of the comparison outcome. Property (3) provides explicit expressions for the expectation and variance of $Y$, offering insight into the signal-to-noise ratio (SNR) of the model. Notably, when $K = 1$, the SNR of $Y$ reaches its maximum, given by $\sinh^2(\phi(\gamma))$.
}

\subsection{Ordinal Comparison Graph}
\label{Sec:DCG}
{ 
In the context of comparison data, it is commonly assumed that there exists a true preference vector, denoted by $\bm{\theta}^\star = (\theta_1^\star, \ldots, \theta_n^\star)^\top$. A higher value of $\theta_i^\star$ relative to $\theta_j^\star$ (i.e., $\theta_i^\star > \theta_j^\star$) indicates that item $i$ is ranked higher than item $j$ in the ground-truth ordering. Let $y_{ij}^{(l)}$ denote the observed preference difference between items $i$ and $j$ in the $l$-th comparison. We assume that this comparison follows the distribution
$$
y_{ij}^{(l)} \sim G(\phi, \psi_{\gamma_{ij}^\star}, \gamma_{ij}^\star, K),
$$
where the distribution $G$ depends on the strength link function $\phi$, the pattern function $\psi_{\gamma_{ij}^\star}$, the preference difference $\gamma_{ij}^\star = \theta_i^\star - \theta_j^\star$, and the number of ordinal levels $K$. Specifically, the probability mass function takes the form
$$
\mathbb{P}\left(y_{ij}^{(l)} = k\right) = \frac{1}{\Psi_{\phi, \psi_{\gamma_{ij}^\star}}(\gamma_{ij}^\star)} \exp\left(g( k\,|\, \phi,\psi_{\gamma_{ij}^\star},\gamma_{ij}^\star)\right), \quad \text{for } k \in \Upsilon(K),
$$
where $y_{ij}^{(l)} = k$ indicates that item $i$ is preferred to item $j$ by $k$ ordinal levels. Here, the form of $\psi_{\gamma_{ij}^\star}$ depends on the value of $\gamma_{ij}^\star$, implying that comparisons between different items may exhibit distinct patterns across the ordinal values.
}

Similar to the BTL model, the optimality of the parameter vector $\bm{\theta}^\star$ in the proposed model is not unique, given that the values of $\gamma_{ij}^\star$ remain unchanged with any translation of $\bm{\theta}^\star$. To ensure the uniqueness of $\bm{\theta}^\star$, we require $\bm{1}_n^\top\bm{\theta}^\star = 0$ as existing literature \citep{liu2023lagrangian,fan2025ranking}.

\begin{theorem}
\label{Thm:Generalization}
Suppose that $Y_{ij} \sim G(\phi, \psi_{\gamma_{ij}^\star}, \gamma_{ij}^\star, 1)$ for some $\psi_{\gamma_{ij}^\star}$. Under this specification, the proposed model simplifies to the following binary pairwise comparison models:
\begin{itemize}
\item[(1)] (BTL model) If $\phi(x) = \frac{1}{2}\log\left(\frac{\sigma(x)}{1-\sigma(x)}\right)$ with $\sigma(x)=\frac{e^x}{1+e^x}$ being the CDF of logistic distribution, then we have $\mathbb{P}(Y_{ij}=1) = \frac{e^{\theta_i^\star}}{e^{\theta_i^\star}+e^{\theta_j^\star}}$.
\item[(2)] (Thurstone-Mosteller model) If $\phi(x) =  \frac{1}{2}\log\left(\frac{\Phi(x)}{1-\Phi(x)}\right)$ with $\Phi(x)=\int_{-\infty}^x (2\pi)^{-1/2}e^{-x^2/2}dx$ being the CDF of standard normal distribution, then we have $\mathbb{P}(Y_{ij}=1) = \Phi(\gamma_{ij}^\star)$.
\end{itemize}
\end{theorem}

In Theorem \ref{Thm:Generalization}, we present the connections of the proposed model to the BTL model and TM model when $K=1$ under specific choices of $\phi$, which are two particular cases of the class of strength link functions specified according to Lemma \ref{Lemma:CDF}. It is worth noting that when $K=1$, $\psi_{\gamma_{ij}^\star}$ becomes inactive as no ordinal structure is involved. In other words, $G(\phi,\psi_{\gamma_{ij}^\star},\gamma_{ij}^\star,1)$ and $G(\phi,0,\gamma_{ij}^\star,1)$ represent the same model, where in the latter case $\psi_{\gamma_{ij}^\star}(k)\equiv 0$ for all $k$.

\section{Is Ordinal Comparison Data Always Better?}
\label{Sec:OrdBetter}
In this section, we investigate whether ordinal comparison data or binary comparison data is more effective for inferring the true ranking of items. Recall that $\gamma_{ij}^\star = \theta_i^\star-\theta_j^\star$, we consider the setting where $y_{ij}^{(l)} \sim G(\phi,\psi_{\gamma_{ij}^\star},\gamma_{ij}^\star,K)$ represents an ordinal comparison, and its binarized counterpart $\sign(y_{ij}^{(l)}) \sim G(\phi,0,\gamma_{ij}^\star,1)$, as described in Theorem~\ref{Thm:MeanVari}. In other words, for any $\phi$, we have
\begin{align*}
    \underbrace{y_{ij}^{(l)} \sim G(\phi,\psi_{\gamma_{ij}^\star},\gamma_{ij}^\star,K)}_{\text{Ordinal Comparison}}
    \,
\xRightarrow{\text{Binarization}}
\,
    \underbrace{\sign(y_{ij}^{(l)}) \sim G(\phi,0,\gamma_{ij}^\star,1)}_{\text{Binary Comparison}}.
\end{align*}

A specific example of the above process is when $\phi(x) = \frac{1}{2} \log \frac{\sigma(x)}{1 - \sigma(x)}$, in which case $\sign(y_{12}^{(l)})$ essentially follows the BTL model. While binarization intuitively results in information loss—making binary comparison data seemingly less informative for recovering the true ranking—we will demonstrate that this intuition does not always hold. To illustrate this point, we begin with a warm-up analysis of the two-item ranking problem using the counting method \citep{busa2013top,shah2018simple}, and then extend the discussion to the setting of full ranking recovery. In practice, it is infeasible to assume that all comparisons are observed. Therefore, for ranking recovery, we adopt an Erd\H{o}s--R\'enyi graph model, in which each comparison is independently missing with probability $1-p$. Formally, we introduce a Bernoulli random variable $a_{ij}^{(l)} \sim \text{Bernoulli}(p)$, where $a_{ij}^{(l)} = 1$ indicates that $y_{ij}^{(l)}$ is observed; hence, the observed comparisons form an Erd\H{o}s--R\'enyi random graph with probability $p$. Formally, this assumption can be summarized as
\begin{align}
\label{Ass_Miss}
\text{\textbf{Random Missing Pattern Assumption: }} \,\,
a_{ij}^{(l)} = 1 \Longrightarrow y_{ij}^{(l)} \text{ is observed},
\end{align}
where $a_{ij}^{(l)}$ is independent of $y_{ij}^{(l)}$ for all $i\neq j$ and $l \in [L]$. Here, $p$ denotes the probability of observing a comparison.

\subsection{Two-item Ranking Problem}
\label{Secsub:TwoItem}
As a warm-up, we consider the two-item ranking problem. Suppose there are two items with preference parameters $\bm{\theta}^\star = (\theta_1^\star, \theta_2^\star)$. Without loss of generality, assume $\theta_1^\star > \theta_2^\star$, so that item~1 is preferred to item~2. The central goal is to recover the ranking induced by $\bm{\theta}^\star$.

One of the simplest approaches to this task is the counting algorithm, a count-based method that is known to be optimal under minimal assumptions on the pairwise comparison process. Let $\{y_{12}^{(l)}\}_{l=1}^L$ denote the collection of pairwise comparison outcomes, where observations follow the missingness assumption in~(\ref{Ass_Miss}). We also consider the binarized version $\{z_{12}^{(l)}\}_{l=1}^L$, where $z_{12}^{(l)} = \sign\!\big(y_{12}^{(l)}\big)$ indicates whether item~1 is preferred to item~2 in the $l$-th comparison.

The count-based method then derive the ordering in preference between items 1 and 2 based on the following metrics:
\begin{align*}
 \text{Count Score using Raw Data: }  & A = \frac{1}{L} \sum_{l=1}^L a_{12}^{(l)}y_{12}^{(l)}, \\
  \text{Count Score using Binarized Data: } &  B = \frac{1}{L} \sum_{l=1}^L a_{12}^{(l)} z_{12}^{(l)},
\end{align*}
where $A$ and $B$ represent the accumulated pairwise rewards of item $1$ based on the original discrete dataset and its binarized counterpart, respectively. Here, $ A > 0 $ indicates that item 1 has a higher accumulated score out of the $ L $ comparisons, implying that item 1 is preferred. Similarly, $ B > 0 $ means that more than half of the individuals choose item 1 over item 2, also indicating a preference for item 1. 

\begin{figure}[h!]
    \centering
    \begin{subfigure}[b]{0.325\textwidth}
        \centering
        \includegraphics[width=\textwidth]{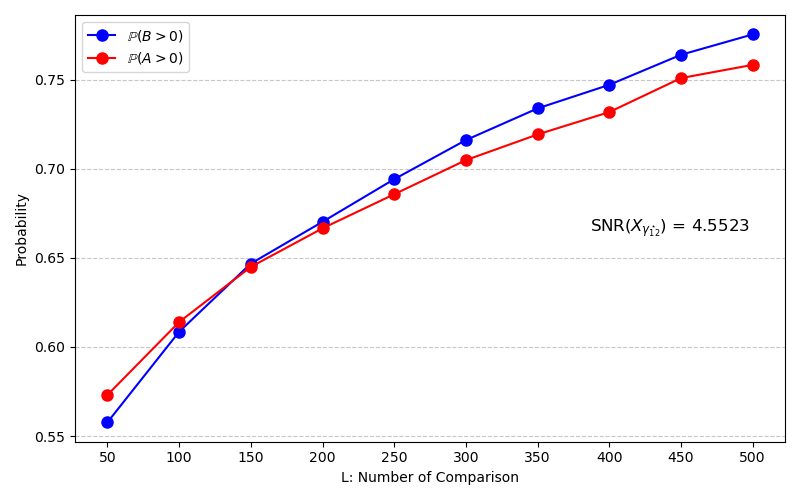}
        \caption{$(\beta,\gamma_{12}^\star)=(0.1,0.05)$}
    \end{subfigure}
        \begin{subfigure}[b]{0.325\textwidth}
        \centering
        \includegraphics[width=\textwidth]{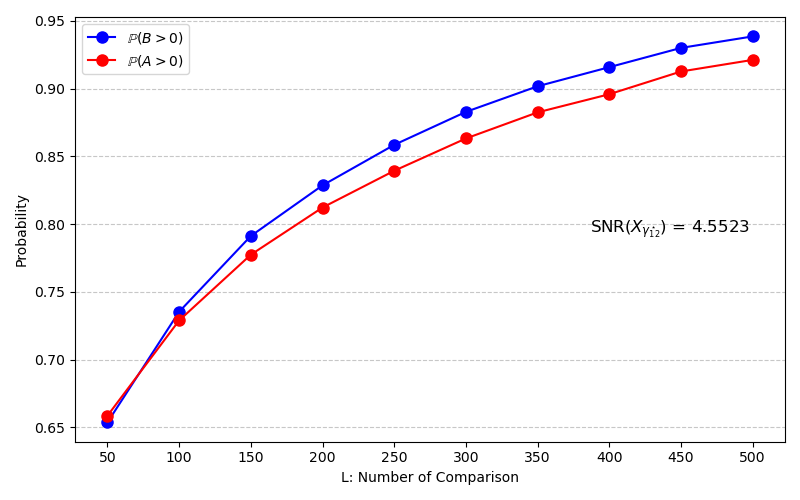}
        \caption{$(\beta,\gamma_{12}^\star)=(0.1,0.1)$}
    \end{subfigure}
        \begin{subfigure}[b]{0.325\textwidth}
        \centering
        \includegraphics[width=\textwidth]{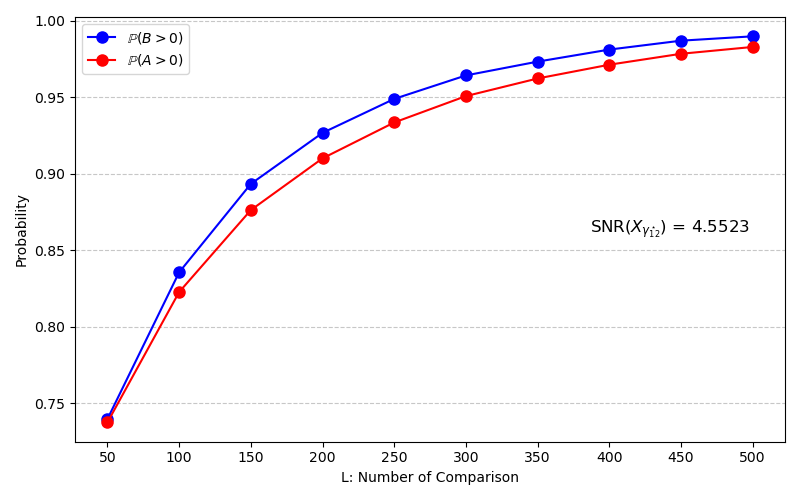}
        \caption{$(\beta,\gamma_{12}^\star)=(0.1,0.15)$}        
    \end{subfigure}
    \hfill
    \begin{subfigure}[b]{0.325\textwidth}
        \centering
        \includegraphics[width=\textwidth]{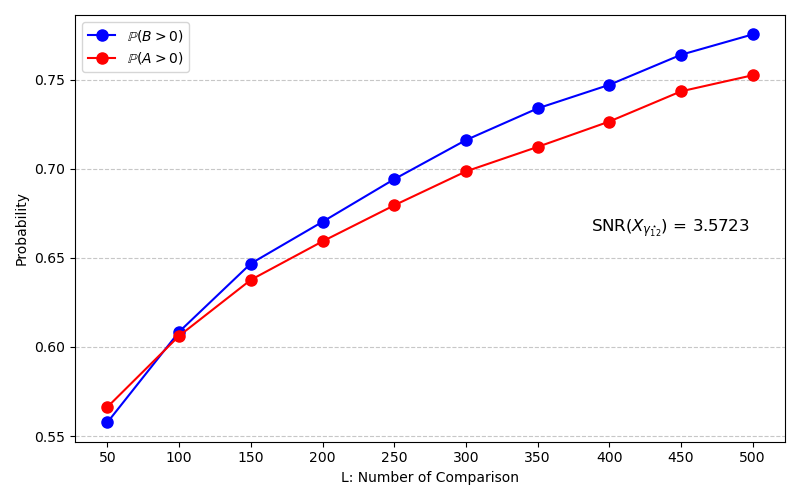}
        \caption{$(\beta,\gamma_{12}^\star)=(0.9,0.05)$}
    
    \end{subfigure}
        \begin{subfigure}[b]{0.325\textwidth}
        \centering
        \includegraphics[width=\textwidth]{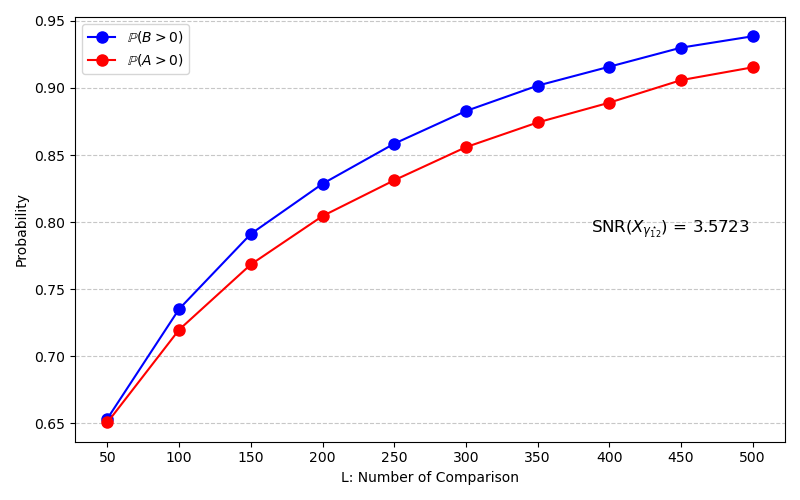}
        \caption{$(\beta,\gamma_{12}^\star)=(0.9,0.1)$}

    \end{subfigure}
        \begin{subfigure}[b]{0.325\textwidth}
        \centering
        \includegraphics[width=\textwidth]{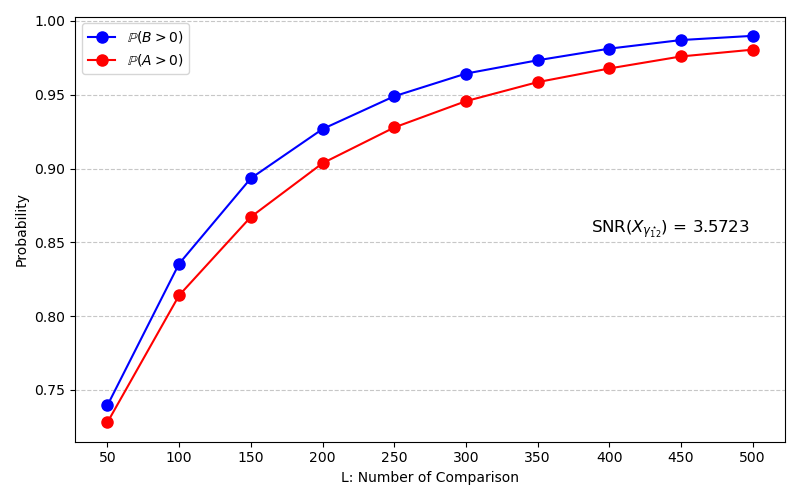}
        \caption{$(\beta,\gamma_{12}^\star)=(0.9,0.15)$}
    \end{subfigure}
    \caption{A comparison between $\mathbb{P}(A > 0)$ and $\mathbb{P}(B > 0)$ is conducted under the proposed model, where the propensity functions are specified as $\phi(x) = x$ and $\psi_{\gamma}(k) = -\beta |k|$. When $\beta = 0.1$, the SNR of $X_{\gamma_{12}^\star}$ is 4.5523, whereas for $\beta = 0.9$, the SNR of $X_{\gamma_{12}^\star}$ decreases to 3.5723. }
    \label{fig:Simu0}
\end{figure}

{ 
We are interested in which metric is more likely to provide correct preference ranking of items. This problem reduces to comparing $\mathbb{P}(A>0)$ and $\mathbb{P}(B>0)$. Intuitively, the difference between $\mathbb{P}(A>0)$ and $\mathbb{P}(B>0)$ arises from the pattern of ordinal values, that is, from the form of $\psi_{\gamma_{12}^\star}$. Specifically, we recall the distribution of $X_{\gamma_{12}^\star} \sim \mathrm{Geo}(\psi_{\gamma_{12}^\star}, K)$, which is given by 
\begin{equation*} 
\mathbb{P}\big(X_{\gamma_{12}^\star} = k\big) = \frac{\exp\big(\psi_{\gamma_{12}^\star}(k)\big)} {\sum_{j=1}^{K} \exp\big(\psi_{\gamma_{12}^\star}(j)\big)}, \quad k \in [K]. \end{equation*} 
Here, the distribution of $X_{\gamma_{12}^\star}$ is determined by the vector $(\psi_{\gamma_{12}^\star}(k))_{k \in [K]}$, which characterizes the underlying pattern of the ordinal outcomes of comparing items 1 and 2.

\begin{theorem}
\label{Thm:Compare}
Suppose that $y_{12}^{(l)} \sim G\big(\phi, \psi_{\gamma_{12}^\star}, \gamma_{12}^\star, K\big)$ and $a_{12}^{(l)} \sim \textnormal{Bernoulli}(p)$ for $l \in [L]$ with $\gamma_{12}^\star>0$, and let $ X_{\gamma_{12}^\star} \sim \mathrm{Geo}(\psi_{\gamma_{12}^\star}, K) $ for any $K \geq 2$. Then, it holds that  
\begin{align*}
\mathbb{P}(B > 0) &\xrightarrow{L \rightarrow\infty}
    \Phi\left(  \sqrt{\frac{Lp}{\csch^2(\phi(\gamma_{12}^\star))+1-p}} \right), \\
\mathbb{P}(A > 0) &\xrightarrow{L \rightarrow\infty} \Phi\left(\sqrt{\frac{Lp}{\Delta(\gamma_{12}^\star)+\csch^2(\phi(\gamma_{12}^\star))+1-p}} \right),
\end{align*}  
where $\Phi$ is the standard normal cumulative distribution function and $\Delta(\gamma_{12}^\star)$ is defined as
\[
\Delta(\gamma_{12}^\star)\triangleq
\frac{1}{\mathrm{SNR}(X_{\gamma_{12}^\star})\tanh^2(\phi(\gamma_{12}^\star))} \geq 0.
\]
with equality holding if and only if $\mathrm{SNR}(X_{\gamma_{12}^\star}) = \infty$. Here, $\mathrm{SNR}(X_{\gamma_{12}^\star}) = \infty$ indicates that $X_{\gamma_{12}^\star}$ is a degenerate distribution concentrated at a single point.
\end{theorem}

Theorem~\ref{Thm:Compare} characterizes the limiting behavior of the probabilities $\mathbb{P}(A > 0)$ and $\mathbb{P}(B > 0)$. Notably, the limit of $\mathbb{P}(B > 0)$ exceeds that of $\mathbb{P}(A > 0)$, and the gap between them depends on the signal-to-noise ratio (SNR) of $X_{\gamma_{12}^\star}$, which is governed by the ordinal value pattern $\psi_{\gamma_{12}^\star}$. This result implies that, in the asymptotic regime, binarizing ordinal comparison data leads to a faster convergence rate in recovering the true ranking. Although this may appear counterintuitive, the underlying intuition is clear: binarization discards magnitude information but significantly reduces the noise inherent in ordinal responses. In other words, while ordinal comparisons provide more detailed information, they also introduce greater uncertainty—an effect that binarization effectively mitigates. Additionally, there are two cases in which binarization yields significant improvement. 
\begin{itemize}[leftmargin=0cm, itemindent=2.1cm]
    \item[\textbf{Case 1}:] When $\text{SNR}(X_{\gamma_{12}^\star})$ is small, the gap between the limiting values of $\mathbb{P}(A > 0)$ and $\mathbb{P}(B > 0)$ increases. This suggests that binarization can be particularly beneficial when the distribution of ordinal values exhibits high variance. This conclusion is further supported by our simulation results presented in Figure \ref{fig:Simu0}.
    \item[\textbf{Case 2}:] When $\mathrm{SNR}(X_{\gamma_{12}^\star})$ is fixed, we have
\begin{equation*}
\frac{\Delta(\gamma_{12}^\star)}{\csch^2(\phi(\gamma_{12}^\star))} 
= \frac{\cosh^2(\phi(\gamma_{12}^\star))}{\mathrm{SNR}(X_{\gamma_{12}^\star})},
\end{equation*}
which increases with $|\gamma_{12}^\star|$, suggesting that binarization can be particularly effective for relatively easy ranking recovery tasks. This implication is corroborated by the subsequent simulation results presented in Figure \ref{fig:Simu0}.
\end{itemize}

Theorem~\ref{Thm:Compare} establishes that the limiting value of $\mathbb{P}(B > 0)$ is greater than that of $\mathbb{P}(A > 0)$. However, this result does not provide a direct comparison of the actual magnitudes of $\mathbb{P}(B > 0)$ and $\mathbb{P}(A > 0)$ in finite-sample settings, as approximation errors persist and it remains unclear which probability is more affected. To address this, we theoretically analyze their convergence rates and establish Theorem~\ref{Thm:BS_bound}.

\begin{theorem}
    \label{Thm:BS_bound}
Suppose that $y_{12}^{(l)} \sim G\big(\phi, \psi_{\gamma_{12}^\star}, \gamma_{12}^\star, K\big)$ and $a_{12}^{(l)} \sim \textnormal{Bernoulli}(p)$ for $l \in [L]$, where $\gamma_{12}^\star > 0$ and $K \ge 2$. Moreover, let $X_{\gamma_{12}^\star} \sim \mathrm{Geo}\big(\psi_{\gamma_{12}^\star}, K\big)$ be non-degenerate. Then,
\begin{align*}
\textnormal{Error Ratio: }
    \lim_{L \rightarrow \infty} \frac{1 - \mathbb{P}(B > 0)}{1 - \mathbb{P}(A > 0)} =   \lim_{L \rightarrow \infty} \frac{\mathbb{P}(B \leq  0)}{\mathbb{P}(A \leq 0)}= 0.
\end{align*}
This indicates that there exists a positive integer $L_0$ (depending on $\gamma_{12}^\star$, $p$ and $\psi_{\gamma_{12}^\star}$) such that, 
$$
\mathbb{P}(B > 0) > \mathbb{P}(A > 0) \,\text{ for all }\, L \geq L_0.
$$
This result indicates that the counting algorithm based on binarized comparison data outperforms its ordinal counterpart in recovering the correct ranking between the two items.
\end{theorem}

In Theorem~\ref{Thm:BS_bound}, we show that although both $\mathbb{P}(A > 0)$ and $\mathbb{P}(B > 0)$ converge to one as $L$ increases, the probability $\mathbb{P}(B \leq 0)$ becomes negligible relative to $\mathbb{P}(A \leq 0)$ for sufficiently large $L$. This result has an important practical implication: once enough comparison data is collected, the probability of misranking two items under binary comparisons is substantially smaller than under ordinal comparisons. Moreover, the threshold $L_0$ beyond which binarization becomes advantageous depends on $\text{SNR}(X_{\gamma_{12}^\star})$. Specifically, when $\text{SNR}(X_{\gamma_{12}^\star})$ is large, the benefit of binarization emerges only at larger sample sizes. In contrast, when $\text{SNR}(X_{\gamma_{12}^\star})$ is small, the advantage of using $B$ over $A$ is more pronounced, so a smaller $L$ suffices for this improvement to appear. Furthermore, for large $\gamma_{12}^\star$, the gap between $\mathbb{P}(B \leq 0)$ and $\mathbb{P}(A \leq 0)$ widens, leading to a larger $L_0$. These theoretical insights are further corroborated by the empirical results in Figure~\ref{fig:Simu0}.
}

{ 
To illustrate the validity of Theorem~\ref{Thm:BS_bound}, we conduct an experiment using a specific propensity function defined as $g(k\,|\,\phi,\psi_{\gamma},\gamma) = \operatorname{sign}(k)\gamma - \beta|k|$ with $K = 4$. To study the impact of $\gamma_{12}^\star$ and $\text{SNR}(X_{\gamma_{12}^\star})$ on $L_0$ separately, we adopt a $\gamma$-independent form of $\psi_{\gamma}(k)=-\beta |k|$. We consider $\gamma \in \{0.05, 0.1, 0.15\}$, vary $L$ from 50 to 500, set $\beta \in \{0.1, 0.9\}$ to control the value of $\text{SNR}(X_{\gamma_{12}^\star})$, and fix the missing probability $p=0.5$. The probabilities $\mathbb{P}(A > 0)$ and $\mathbb{P}(B > 0)$ are estimated using $10^6$ Monte Carlo replications under different combinations of $(\beta, \gamma_{12}^\star)$.}

Several conclusions can be drawn from Figure~\ref{fig:Simu0}. First, in each case, there exists a threshold for $L$ such that when $L$ exceeds this threshold, using binarized comparison data yields better ranking recovery performance—specifically, the blue curve (representing binary comparison data) lies above the red curve (representing ordinal comparison data). Second, as $\gamma_{12}^\star$ increases, the point at which $\mathbb{P}(B > 0)$ surpasses $\mathbb{P}(A > 0)$ occurs at a smaller sample size $L$. This observation aligns with Theorem~\ref{Thm:Compare}, which suggests that a larger value of $|\gamma_{12}^\star|$ leads to a greater asymptotic performance gap. Third, as $\text{SNR}(X_{\gamma_{12}^\star})$ decreases from 4.5523 (when $\beta = 0.1$) to 3.5723 (when $\beta = 0.9$), the blue curve shifts slightly upward, indicating that a smaller $\text{SNR}(X_{\gamma_{12}^\star})$ leads to a more pronounced advantage from binarization.

\subsection{Multiple-Item Ranking Problem}

In this section, we extend the result from Section~\ref{Secsub:TwoItem} to the full ranking recovery problem for $n$ items using the counting method. In particular, we show that binarizing the comparison data can also improve full ranking recovery performance compared to using the original ordinal comparisons. { Similar to Section~\ref{Secsub:TwoItem}, we adopt the same assumption on the missing pattern specified in (\ref{Ass_Miss}) for full ranking recovery.}

Consider $n$ items with preference parameters $\bm{\theta}^\star = (\theta_1^\star, \theta_2^\star, \ldots, \theta_n^\star)$. Without loss of generality, we assume $\theta_1^\star > \theta_2^\star > \cdots > \theta_n^\star$, indicating that item 1 is the most preferred. The goal is to recover the full ranking of $\bm{\theta}^\star$. A key distinction between the $n$-item ranking problem and the two-item ranking problem is the availability of indirect comparisons. Suppose a full comparison dataset is observed as $\{\bm{y}^{(l)}\}_{l=1}^L$ with $\bm{y}^{(l)} = (y_{ij}^{(l)})_{i,j \in [n]}$. In the two-item case, we only observe direct comparisons between items 1 and 2. In contrast, in the $n$-item setting, we also observe comparisons such as $y_{13}^{(l)}$ and $y_{23}^{(l)}$, which provide indirect evidence about the relative preference between items 1 and 2.

{ 
For $\{\bm{y}^{(l)}\}_{l=1}^L$, where $\bm{y}^{(l)} = (y_{ij}^{(l)})_{i,j \in [n]}$, we calculate the score of the $i$-th item using the original comparison data and binarized comparison data as follows:
\begin{align*}
\text{Win-Count using Raw Data:} \quad & S_i = \sum_{j \in [n]\setminus \{i\}} \mathbb{I}\left[\sum_{l=1}^L a_{ij}^{(l)} y_{ij}^{(l)}>0\right], \\
\text{Win-Count using Binarized Data:} \quad & \widetilde{S}_i = \sum_{j \in [n]\setminus \{i\}}\mathbb{I}\left[ \sum_{l=1}^L a_{ij}^{(l)}\operatorname{sign}(y_{ij}^{(l)})>0\right],
\end{align*}
where $\mathbb{I}(A)$ denotes the indicator function, taking the value 1 if the statement $A$ is true and 0 otherwise,  $S_i$ and $\widetilde{S}_i$ denote the win counts of items based on ordinal and binary data, respectively.}

Let $\bm{S}=(S_1,\ldots,S_n)$ and $\widetilde{\bm{S}}=(\widetilde{S}_1,\ldots,\widetilde{S}_n)$ denote the win-count vectors of $n$ items. To evaluate their ranking performance, we define the ranking function $\sigma(\bm{S}) = (\sigma(S_i))_{i \in [n]}$, where $\sigma(S_i)$ represents the rank of $S_i$ among the values of $\bm{S}$. Specifically, $\sigma(S_i)=k$ indicates that $S_i$ is the $k$-th largest entry of $\bm{S}$. In particular, if $\theta_1^\star > \theta_2^\star > \cdots > \theta_n^\star$, then $\sigma(\bm{\theta}^\star)=(1,2,\ldots,n)$. Theorem \ref{Thm:Consist} establishes the validity of using either $\bm{S}$ or $\widetilde{\bm{S}}$ to recover the true ranking of items.

\begin{theorem}[\textbf{Ranking Consistency}]
\label{Thm:Consist}
For both $\bm{S}$ and $\widetilde{\bm{S}}$, as $L \to \infty$, we have
\[
\sigma(S_i) \xrightarrow{\text{a.s.}} \sigma(\theta_i^\star) \quad \text{and} \quad
\sigma(\widetilde{S}_i) \xrightarrow{\text{a.s.}} \sigma(\theta_i^\star)
\]
for each $i \in [n]$.
\end{theorem}

Theorem \ref{Thm:Consist} shows that $\bm{S}$ and $\widetilde{\bm{S}}$ are both consistent with $\bm{\theta}^\star$ in terms of ranking in the asymptotic regime. Therefore, as the sample size $L$ increases, $\sigma(\bm{S})$ and $\sigma(\widetilde{\bm{S}})$ converge to the true ranking $\sigma(\bm{\theta}^\star)$ almost surely, thereby ensuring consistent ranking recovery.

In what follows, we investigate whether $\bm{S}$ or $\widetilde{\bm{S}}$ yields more accurate rankings. To this end, we assess ranking performance using the Kendall tau distance \citep{kendall1938new}, following standard practice in the literature \citep{xu2025rate, chen2022optimal}. Specifically, we measure the Kendall tau distance between the rankings induced by $\bm{S}$ and $\widetilde{\bm{S}}$ and the true ranking $\bm{\theta}^\star$ by calculating
\begin{align*}
\tau(\bm{S},\bm{\theta}^\star)=&\frac{2}{n(n-1)}
\sum_{1 \leq i <j \leq n} \mathbb{I}\left[\big(\sigma(S_i)-\sigma(S_j)\big)\big(\sigma(\theta_i^\star)-\sigma(\theta_j^\star)\big) \leq 0\right], \\
\tau(\widetilde{\bm{S}},\bm{\theta}^\star)=&\frac{2}{n(n-1)}
\sum_{1 \leq i <j \leq n} \mathbb{I}\left[\big(\sigma(\widetilde{S}_i)-\sigma(\widetilde{S}_j)\big)\big(\sigma(\theta_i^\star)-\sigma(\theta_j^\star)\big) \leq 0\right].
\end{align*}
Here, $\tau(\bm{S}, \bm{\theta}^\star)$ denotes the proportion of item pairs that are misranked by $\bm{S}$ relative to the ground truth ranking $\bm{\theta}^\star$. The case of $\tau(\bm{S}, \bm{\theta}^\star) = 0$ indicates perfect ranking agreement, meaning that $\big(\sigma(S_i) - \sigma(S_j)\big)\big(\sigma(\theta_i^\star) - \sigma(\theta_j^\star)\big) > 0$ for all $i \neq j$.

To evaluate whether $\widetilde{\bm{S}}$ produces more accurate rankings than $\bm{S}$, we analyze the convergence rates of both $\tau(\bm{S},\bm{\theta}^\star)$ and $\tau(\widetilde{\bm{S}},\bm{\theta}^\star)$, showing that $\mathbb{E}\big[\tau(\widetilde{\bm{S}}, \bm{\theta}^\star)\big]$ converges faster than $\mathbb{E}\big[\tau(\bm{S}, \bm{\theta}^\star)\big]$. A direct implication of this result is that, for sufficiently large $L$, ranking recovery based on binarized comparison data consistently outperforms that based on ordinal comparison data.

\begin{theorem}
    \label{ThmK_bound}
    Define $R(\widetilde{\bm{S}},\bm{S})\triangleq\frac{\mathbb{E}[\tau(\widetilde{\bm{S}},\bm{\theta}^\star)]}{\mathbb{E}[\tau(\bm{S},\bm{\theta}^\star)]}$. Suppose that $ y_{ij}^{(l)} \sim G\big(\phi, \psi_{\gamma_{ij}^\star}, \gamma_{ij}^\star, K\big) $ with $\gamma_{ij}^\star > 0$ for $i < j$, and let $ X_{\gamma_{ij}^\star} \sim \mathrm{Geo}(\psi_{\gamma_{ij}^\star}, K) $ for $i < j$ and $K \geq 2$. It then follows that
\begin{align}
\label{Conver}
\textnormal{Full Ranking Error Ratio: }
    \lim_{L\rightarrow \infty} R(\widetilde{\bm{S}},\bm{S}) = \lim_{L\rightarrow \infty} \frac{\mathbb{E}\big[\tau(\widetilde{\bm{S}},\bm{\theta}^\star)\big]}{\mathbb{E}\big[\tau(\bm{S},\bm{\theta}^\star)\big]}=0.
\end{align}
Furthermore, there exists a positive integer $ L_1 $ (depending on $\{\gamma_{ij}^\star:i<j\}$, $p$, and $\{\psi_{\gamma_{ij}^\star}:i<j\}$) such that 
$$
\mathbb{E}\big[\tau(\widetilde{\bm{S}},\bm{\theta}^\star)\big]
<
\mathbb{E}\big[\tau(\bm{S},\bm{\theta}^\star)\big], \text{ for all }L \geq L_1,
$$
implying that the counting algorithm based on binarized comparison data outperforms its ordinal counterpart in recovering the ranking of $n$ items.
\end{theorem}

\begin{figure}[ht!]
    \centering
    \begin{subfigure}[b]{0.325\textwidth}
        \centering
        \includegraphics[width=\textwidth]{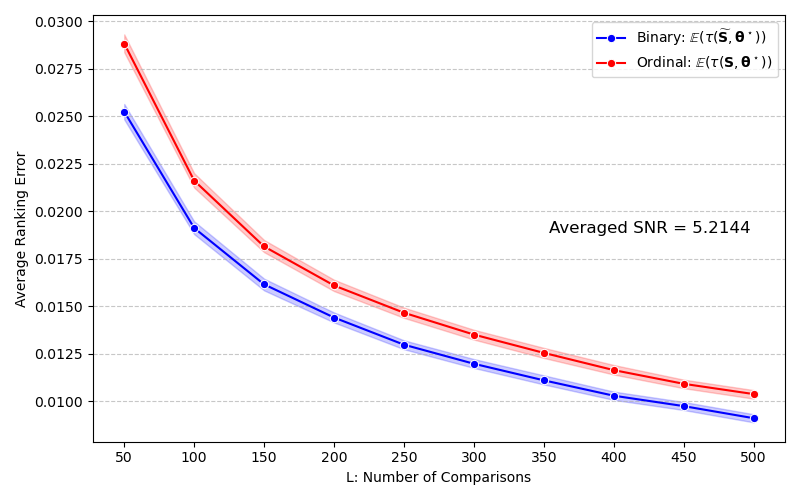}
        \caption{$(w,\beta)=(0.05,0.1)$}
    \end{subfigure}
        \begin{subfigure}[b]{0.325\textwidth}
        \centering
        \includegraphics[width=\textwidth]{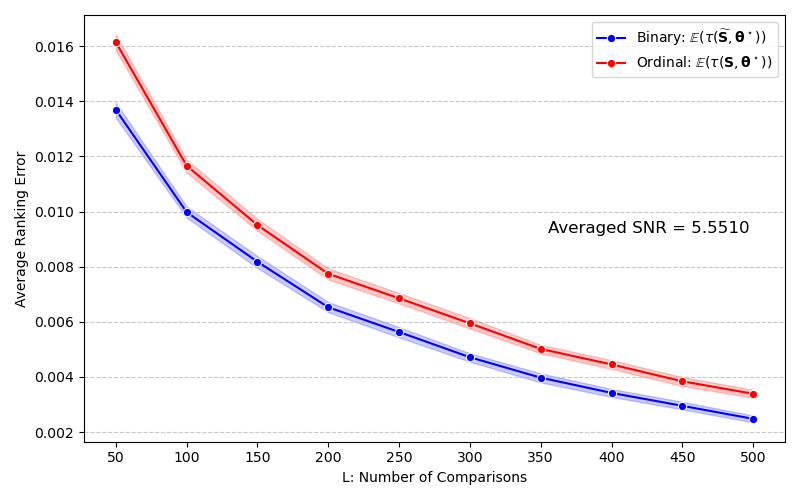}
        \caption{$(w,\beta)=(0.1,0.1)$}
    \end{subfigure}
        \begin{subfigure}[b]{0.325\textwidth}
        \centering
        \includegraphics[width=\textwidth]{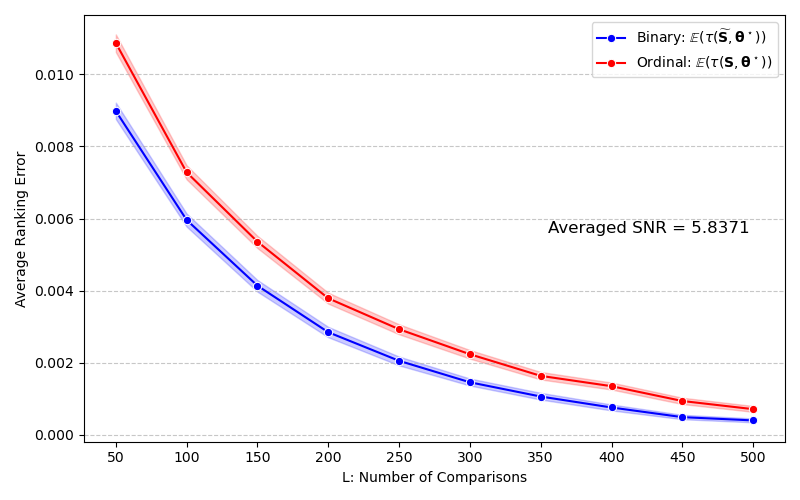}
        \caption{$(w,\beta)=(0.15,0.1)$}  
    \end{subfigure}
    \hfill
    \begin{subfigure}[b]{0.325\textwidth}
        \centering
        \includegraphics[width=\textwidth]{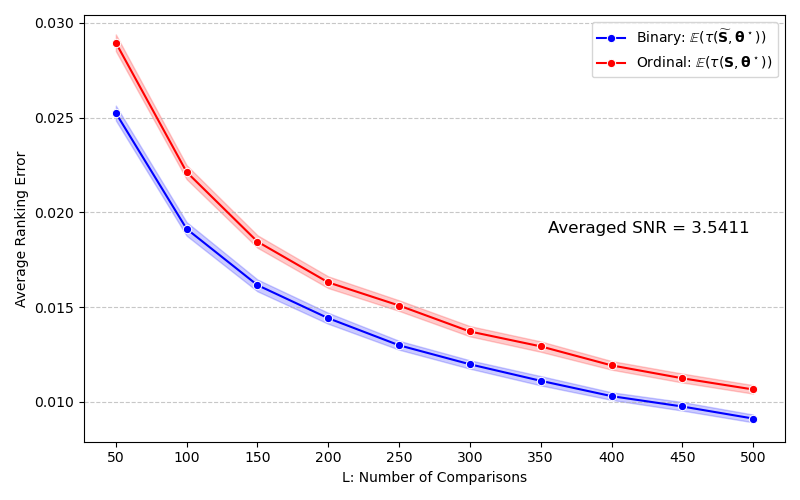}
        \caption{$(w,\beta)=(0.05,0.2)$}
    \end{subfigure}
        \begin{subfigure}[b]{0.325\textwidth}
        \centering
        \includegraphics[width=\textwidth]{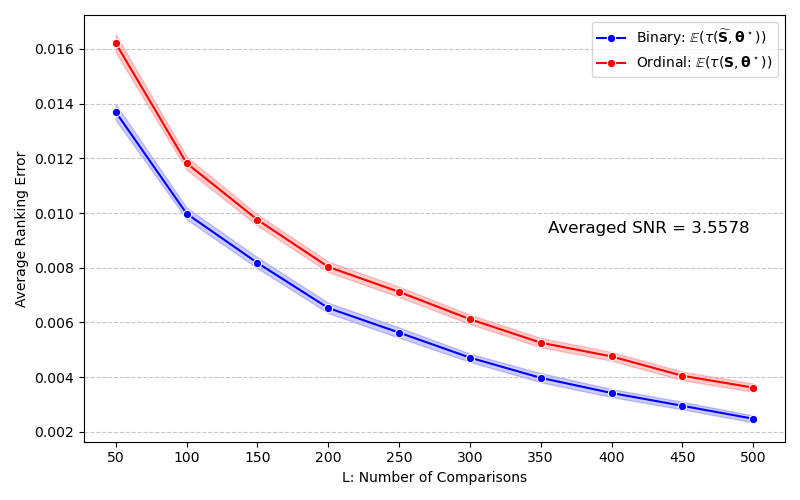}
        \caption{$(w,\beta)=(0.1,0.2)$}
    \end{subfigure}
        \begin{subfigure}[b]{0.325\textwidth}
        \centering
        \includegraphics[width=\textwidth]{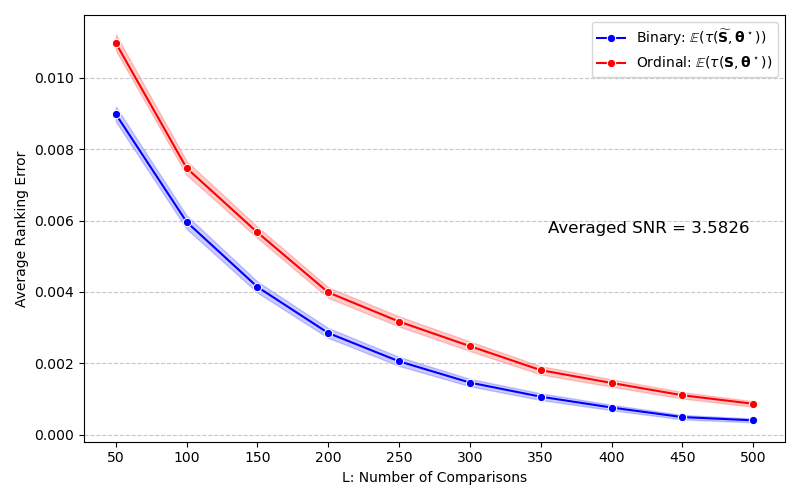}
        \caption{$(w,\beta)=(0.15,0.2)$}
    \end{subfigure}
    \caption{Comparison of $\mathbb{E}\big[\tau(\bm{S}, \bm{\theta}^\star)\big]$ and $\mathbb{E}\big[\tau(\widetilde{\bm{S}}, \bm{\theta}^\star)\big]$ under the proposed model for two choices of the pattern function: $\psi_{\gamma}(x) = -0.1|x| + 0.5\sqrt{|x\gamma|}$ (top) and $\psi_{\gamma}(x) = -0.9|x| + 0.5\sqrt{|x\gamma|}$ (bottom).}
    \label{fig:Simu1}
\end{figure}

To support the claims of Theorem \ref{ThmK_bound}, we conduct a simulation study using a specific propensity function defined as $g(k\,|\,\phi,\psi_{\gamma},\gamma) = \operatorname{sign}(k)\gamma - \beta|k|+0.5\sqrt{|k \cdot \gamma|}$, with $K = 4$ and $\beta \in \{0.1, 0.9\}$ controlling the averaged value of $\text{SNR}(X_{\gamma_{ij}^\star})$ for $i<j$. Intuitively, the result in Theorem~\ref{ThmK_bound} should depend on the averaged $\text{SNR}(X_{\gamma_{ij}^\star})$, consistent with Theorem~\ref{Thm:Compare}. We set the number of items to $n = 40$ and assume evenly spaced true preference parameters such that $\theta_{i}^\star - \theta_{i+1}^\star = w$, where $w \in \{0.05, 0.1, 0.15\}$. The sample size $L$ is varied from 50 to 500, and for each combination of $(w, \beta)$, we estimate $\mathbb{E}\big[\tau(\widetilde{\bm{S}}, \bm{\theta}^\star)\big]$ and $\mathbb{E}\big[\tau(\bm{S}, \bm{\theta}^\star)\big]$ as well as their 99\% confidence intervals based on 1,000 replications. The experimental results are reported in Figure \ref{fig:Simu1}.

As shown in Figure \ref{fig:Simu1}, $\mathbb{E}\big[\tau(\widetilde{\bm{S}}, \bm{\theta}^\star)\big]$ is smaller than $\mathbb{E}\big[\tau(\bm{S}, \bm{\theta}^\star)\big]$ in all cases. This result demonstrates that, in terms of full ranking recovery, binary comparison data also outperforms its ordinal counterparts, with the improvement largely dependent on the pattern function.

\section{Sigal-to-Noise Ratio Analysis}
\label{Sec:SNR}
Based on the analysis presented in Section~\ref{Sec:OrdBetter}, the SNR of $X_{\gamma} \sim \mathrm{Geo}(\psi_{\gamma}, K)$ emerges as a critical factor in determining the extent to which binarizing ordinal comparison data enhances ranking performance. As established in Theorems~\ref{Thm:Compare}, the asymptotic performance gap between the count-based ranking methods applied to binarized versus full ordinal data increases as $\mathrm{SNR}(X_{\gamma})$ decreases. That is, the benefit of binarization is most pronounced when $\text{SNR}(X_{\gamma})$ is minimized. This observation naturally motivates the following question:
\begin{align*}
    \text{What type of } \psi_{\gamma} \text{ minimizes } \mathrm{SNR}(X_{\gamma})?
\end{align*}
The minimal value of $\mathrm{SNR}(X_{\gamma})$ corresponds to the maximal relative gain achieved by employing binarized comparison data instead of full ordinal comparison data.

To address the above question, we consider two distinct scenarios:
(1) minimizing $\mathrm{SNR}(X_{\gamma})$ without any constraints on $\psi_{\gamma}$, and
(2) minimizing $\mathrm{SNR}(X_{\gamma})$ under the assumption that $\psi_{\gamma}$ is non-increasing in $k$. The second scenario is motivated by the empirical observation in Section~\ref{SubSec:Prob}, where more extreme ordinal comparisons are found to be less frequent in real datasets. Accordingly, we impose the constraint that $\mathbb{P}(X_{\gamma} = k) \geq \mathbb{P}(X_{\gamma} = k + 1)$ for all $k \in [K-1]$ and investigate the minimal $\mathrm{SNR}(X_{\gamma})$ under this monotonicity condition.

\begin{theorem}[\textbf{Minimal $\mathrm{SNR}(X_{\gamma})$ without Constraints}]
    \label{Thm:MinimalSNR}
Suppose $X_{\gamma} \sim \mathrm{Geo}(\psi_{\gamma}, K)$ for $K\geq 2$. Then the signal-to-noise ratio of $X_{\gamma}$ has the following lower bound
$$
\mathrm{SNR}(X_{\gamma}) \geq \frac{4K}{(K-1)^2},
$$
with the equality holding if and only if
$$
\psi_{\gamma}(k) = 
\begin{cases}
C+
\log\left( \dfrac{K}{K+1} \right), & \text{if } k = 1, \\
C+\log\left( \dfrac{1}{K+1} \right), & \text{if } k = K, \\
-\infty, & \text{otherwise},
\end{cases}
$$
for any constant $C \in \mathbb{R}$. This choice of $\psi$ corresponds to a two-point distribution of $X$ supported on $\{1, K\}$.
\end{theorem}

For the first scenario, we establish Theorem~\ref{Thm:MinimalSNR}, which characterizes the minimal SNR achievable by $X_{\gamma}$ without imposing structural constraints on $\psi_{\gamma}$. Interestingly, the minimal SNR is closely tied to the maximum category $K$ of the ordinal comparison: as $K$ increases, the minimal SNR decreases. This result suggests that in practice, when the number of categories in ordinal comparison data is large, binarizing the ordinal responses may lead to greater improvements in the asymptotic regime. 

However, the minimal SNR is attained when $\psi_{\gamma}(k) = -\infty$ for all $k \notin \{1, K\}$. Plugging this choice of $\psi_{\gamma}$ into the proposed model $g$ yields an ordinal comparison model that generates responses only from the set $\{-K, -1, 1, K\}$. This implies that users provide only extreme responses ($\pm K$) or mild responses ($\pm 1$), omitting intermediate levels. Such behavior may not align with the patterns typically observed in real-world ordinal datasets (Section \ref{SubSec:Prob}).

A more practical approach to ordinal comparison should preserve the non-increasing pattern of probabilities. To this end, we impose an additional constraint on $\psi_{\gamma}$, requiring that $\psi_{\gamma}(1) \geq \psi_{\gamma}(2) \geq \cdots \geq \psi_{\gamma}(k)$. This ensures that $\mathbb{P}(X_{\gamma} = k) \geq \mathbb{P}(X_{\gamma} = k+1)$ if $X_{\gamma} \sim \text{Geo}(\psi_{\gamma}, K)$. Under this monotonicity constraint, the minimal SNR is characterized in Theorem~\ref{Thm:MinimalSNRNo}.

\begin{theorem}[\textbf{Minimal $\mathrm{SNR}(X_{\gamma})$ with non-increasing} $\psi_{\gamma}$]
    \label{Thm:MinimalSNRNo}
Suppose $X_{\gamma} \sim \mathrm{Geo}(\psi_{\gamma}, K)$, where $\psi_{\gamma}(i) \geq  \psi_{\gamma}(j)$ for $i< j$ and $K \geq 2$. Then the signal-to-noise ratio satisfies
$$
\mathrm{SNR}(X_{\gamma}) \geq \frac{24(K+1)}{4K^2-4K+1},
$$
with the equality holding if and only if
$$
\psi_{\gamma}(k) = 
\begin{cases}
C+\log\left(\frac{(2K^2+K+2)(K-1)}{2(2K-1)}\right), & \text{if } k = 1, \\
C, & \text{if } k \in \{2,\ldots,K\},
\end{cases}
$$
for any constant $C \in \mathbb{R}$. This choice of $\psi_{\gamma}$ corresponds to the distribution of $X_{\gamma}$ uniformly supported on $\{2,\ldots, K\}$.
\end{theorem}

In Theorem~\ref{Thm:MinimalSNRNo}, we show that under the non-increasing constraint, the minimal SNR achievable is $\frac{24(K+1)}{4K^2 - 4K + 1}$. This minimal SNR is attained when $X_{\gamma}$ is uniform over ${2, \ldots, K}$ with additional mass placed at $X_{\gamma} = 1$. It is worth noting that when $K = 2$, Theorems~\ref{Thm:MinimalSNR} and~\ref{Thm:MinimalSNRNo} yield the same distribution and minimal SNR.

\section{Experiment}
\label{Sec:Exp}
In this section, we conduct extensive simulations to validate our theoretical results in Theorems \ref{Thm:Compare}–\ref{ThmK_bound}, and further demonstrate the effectiveness of binary comparisons in inferring relative item preferences using a real-world dataset.

{ 
\subsection{Simulation}
\label{SubSec:Simu}
In this part, we aim to empirically validate three key conclusions derived from our theoretical analysis. First, binarizing ordinal comparisons improves rank recovery performance when using the counting algorithm, across various configurations of $\phi$ and $\psi_{\gamma}$ (\textbf{Scenario I}). Second, we show that the advantage of binarization becomes more pronounced as the signal-to-noise ratio of the ordinal comparison pattern decreases (\textbf{Scenario II}). Third, we examine the relationship between $R(\widetilde{\bm{S}}, \bm{S})$ and $L$ to verify (\ref{Conver}) in Theorem~\ref{ThmK_bound} (\textbf{Scenario III}).

\noindent
\textbf{Scenario I.} In the first scenario, we compare the ranking errors of the counting method based on binary comparisons versus ordinal comparisons. Specifically, we consider four choices of $\phi$, including $\phi^{(1)}(x) = x$, $\phi^{(2)}(x) = \frac{1}{2}\log \frac{\Phi(x)}{1 - \Phi(x)}$, and $\phi^{(3)}(x) = \frac{1-e^{-x}}{1+e^{-x}}$, along with two choices of $\psi_{\gamma}$, namely, $\psi_{\gamma}^{(1)}(x) = -0.5|x|+0.5\sqrt{|x\gamma|}$ and $\psi_{\gamma}^{(2)}(x) = -0.5x^2+0.5\sqrt{|x\gamma|}$. We fix $K=5$ and vary the number of comparisons $L$ over the set $\{50 + 50 \times i : 0 \leq i \leq 9\}$ and consider $n \in \{20, 40\}$, with the true parameter vector $\bm{\theta}^\star$ being equally spaced with width $0.05$. Furthermore, we fix the missing probability as 0.5, that is $p=0.5$. We replicate each case 1,000 times for estimating the averaged ranking errors as well as their 99\% confidence intervals. The experimental results are reported in Figure \ref{fig:Scen1}.

\begin{figure}[h!]
    \centering
    \begin{subfigure}[b]{0.323\textwidth}
        \centering
        \includegraphics[width=\textwidth]{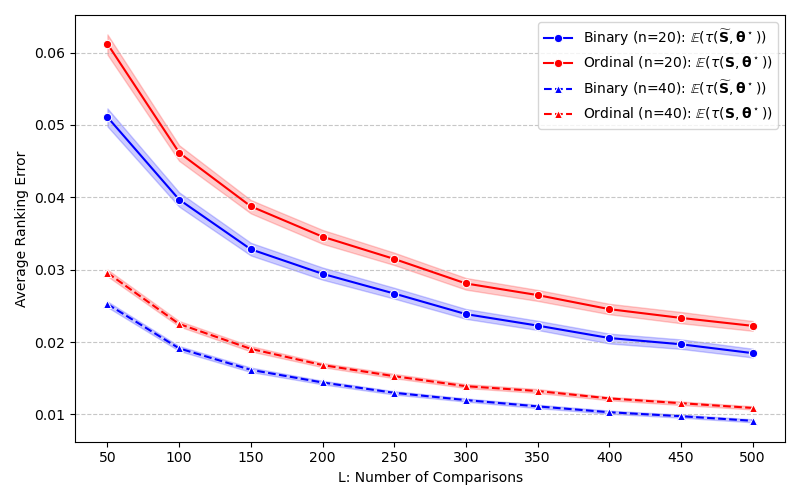}
        \caption{$\phi^{(1)}(x)$ and $\psi^{(1)}_{\gamma}(x)$}
    \end{subfigure}
        \begin{subfigure}[b]{0.323\textwidth}
        \centering
        \includegraphics[width=\textwidth]{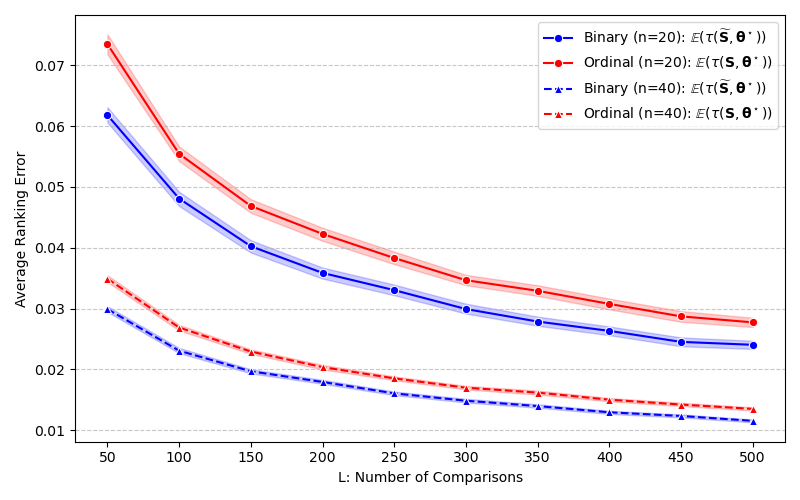}
        \caption{$\phi^{(2)}(x)$ and $\psi^{(1)}_{\gamma}(x)$}
    \end{subfigure}
        \begin{subfigure}[b]{0.323\textwidth}
        \centering
        \includegraphics[width=\textwidth]{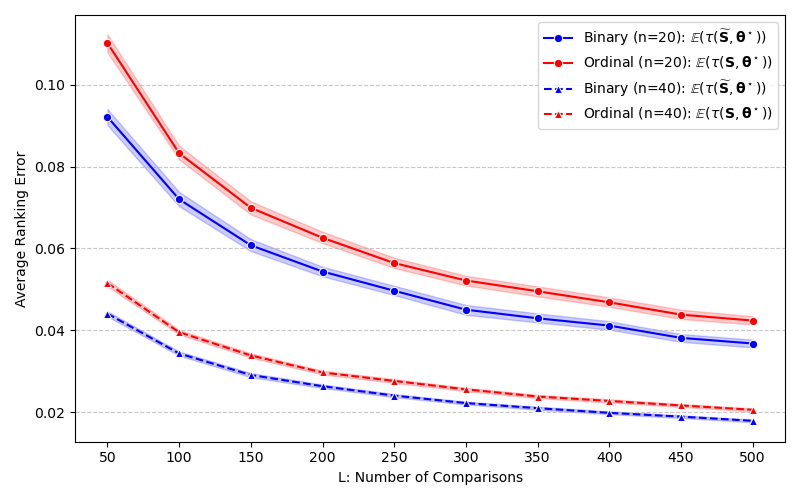}
        \caption{$\phi^{(3)}(x)$ and $\psi^{(2)}_{\gamma}(x)$}
    \end{subfigure}

        \begin{subfigure}[b]{0.323\textwidth}
        \centering
        \includegraphics[width=\textwidth]{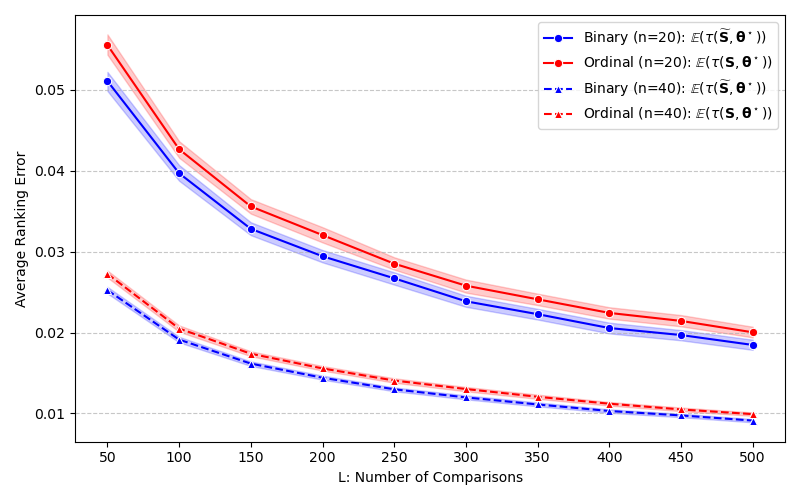}
        \caption{$\phi^{(1)}(x)$ and $\psi^{(2)}_{\gamma}(x)$}
    \end{subfigure}
        \begin{subfigure}[b]{0.323\textwidth}
        \centering
        \includegraphics[width=\textwidth]{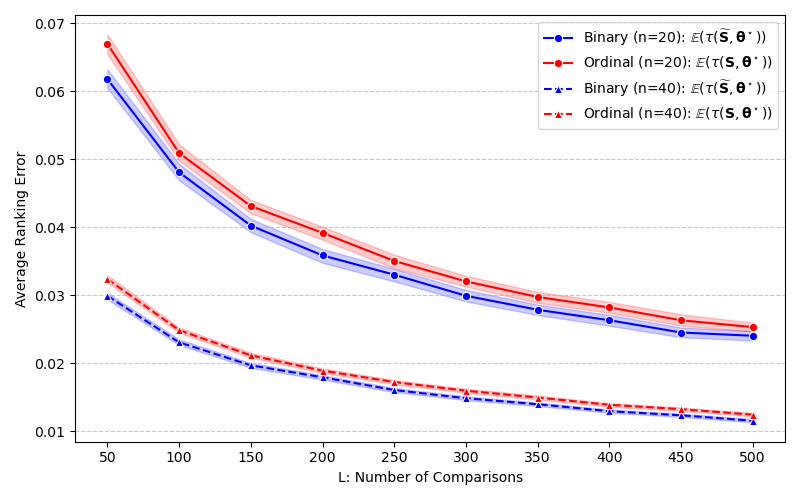}
        \caption{$\phi^{(2)}(x)$ and $\psi^{(2)}_{\gamma}(x)$}
    \end{subfigure}
            \begin{subfigure}[b]{0.323\textwidth}
        \centering
        \includegraphics[width=\textwidth]{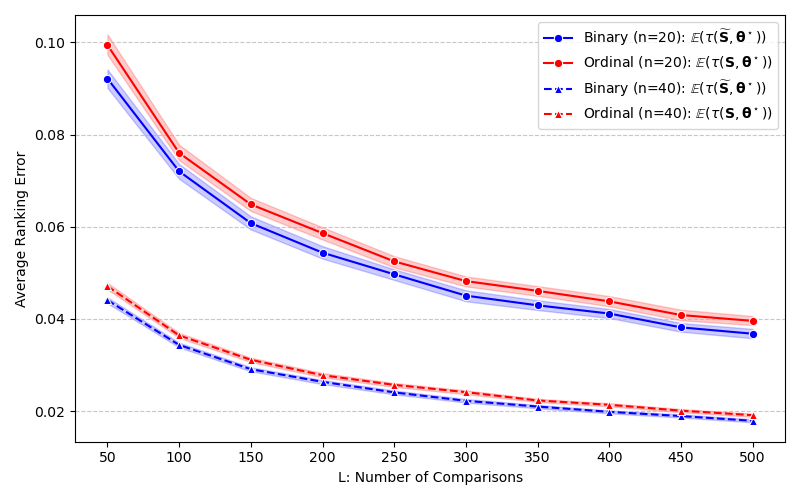}
        \caption{$\phi^{(3)}(x)$ and $\psi^{(2)}_{\gamma}(x)$}
    \end{subfigure}

    \caption{  A comparison between $\mathbb{E}\big[\tau(\bm{S},\bm{\theta}^\star)\big]$ and $\mathbb{E}\big[\tau(\widetilde{\bm{S}},\bm{\theta}^\star)\big]$ under the proposed model with varying $\phi$ and $\psi_{\gamma}$. }
    \label{fig:Scen1}
\end{figure}

As shown in Figure~\ref{fig:Scen1}, the ranking performance based on binary comparison data consistently outperforms that based on ordinal comparison data across all settings. This observation aligns with our theoretical result in Theorem~\ref{ThmK_bound}. Moreover, as either $n$ or $L$ increases, the ranking accuracy of both methods improves. Notably, when $\psi_{\gamma}$ is set to $\psi_{\gamma}^{(2)}$, the performance gap between the two methods is less pronounced than in the case of $\psi_{\gamma}^{(1)}$. This is because, under $\psi_{\gamma}^{(2)}(x)$, the ordinal comparison pattern yields a higher signal-to-noise ratio, resulting in a smaller limiting performance gap.

\noindent
\textbf{Scenario II.} In the second scenario, we investigate how the ranking performance gap, $\mathbb{E}(\tau(\widetilde{\bm{S}},\bm{\theta}^\star)) - \mathbb{E}(\tau(\bm{S},\bm{\theta}^\star))$, relates to the signal-to-noise ratio (SNR) of $X_{\gamma} \sim \mathrm{Geo}(\psi_{\gamma}, K)$. As in Scenario I, we consider three different forms of the function $\phi$. For the pattern function $\psi_{\gamma}$, we examine two $\gamma$-independent $\psi_{\gamma}$: $\psi_{\gamma}^{(3)}(x) = -\beta|x|$ and $\psi_{\gamma}^{(4)}(x) = -\beta x^2$, where $\beta$ is a nuisance parameter varying from $0.1$ to $1$. Here, the main purpose of considering a $\gamma$-independent $\psi_{\gamma}$ is to eliminate the influence of $\gamma$ when ranking multiple items, thereby allowing us to isolate and understand how the ordinal pattern impacts ranking improvement. For each $\beta$, we compute the corresponding SNR, denoted by $\mathrm{SNR}(X_3)$ and $\mathrm{SNR}(X_4)$, where $X_3 \sim \mathrm{Geo}(\psi_{\gamma}^{(3)}, K)$ and $X_4 \sim \mathrm{Geo}(\psi_{\gamma}^{(4)}, K)$. In other words, different values of $\beta$ induce different SNR levels. We fix $(n, L, K) = (10, 100, 5)$ and replicate each configuration $10^6$ times to estimate the average ranking performance gap and the associated SNR. The experimental results are presented in Figure~\ref{fig:Scen2}.

\begin{figure}[h!]
    \centering
    \begin{subfigure}[b]{0.323\textwidth}
        \centering
        \includegraphics[width=\textwidth]{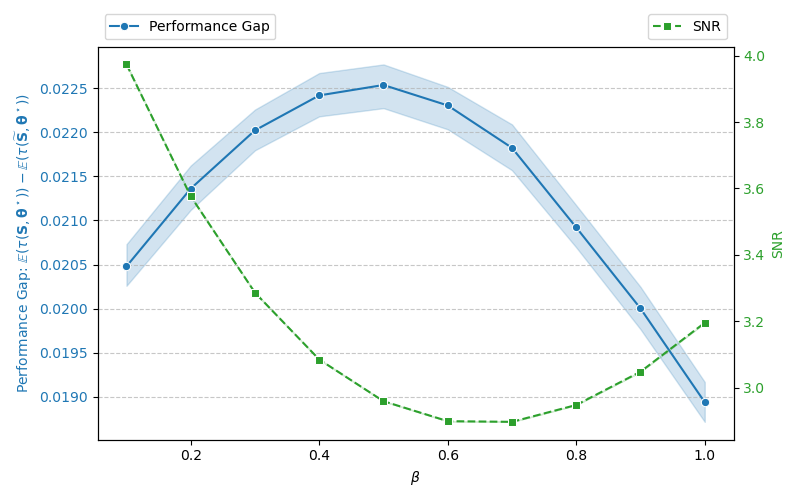}
        \caption{$\phi^{(1)}(x)$ and $\psi_{\gamma}^{(3)}(x)$}
    \end{subfigure}
        \begin{subfigure}[b]{0.323\textwidth}
        \centering
        \includegraphics[width=\textwidth]{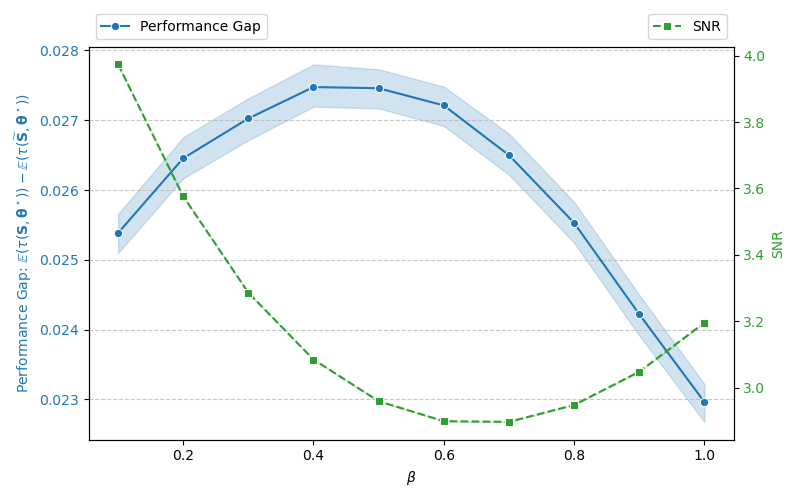}
        \caption{$\phi^{(2)}(x)$ and $\psi_{\gamma}^{(3)}(x)$}
    \end{subfigure}
        \begin{subfigure}[b]{0.323\textwidth}
        \centering
        \includegraphics[width=\textwidth]{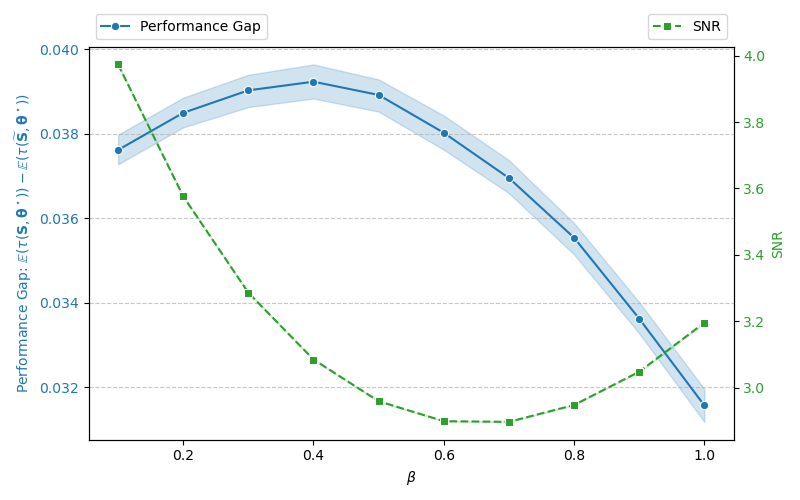}
        \caption{$\phi^{(3)}(x)$ and $\psi_{\gamma}^{(3)}(x)$}
    \end{subfigure}

        \begin{subfigure}[b]{0.323\textwidth}
        \centering
        \includegraphics[width=\textwidth]{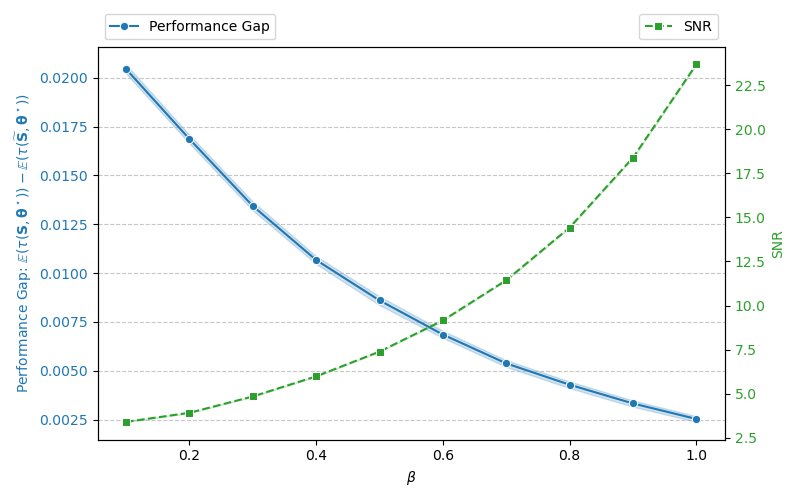}
        \caption{$\phi^{(1)}(x)$ and $\psi_{\gamma}^{(4)}(x)$}
    \end{subfigure}
        \begin{subfigure}[b]{0.323\textwidth}
        \centering
        \includegraphics[width=\textwidth]{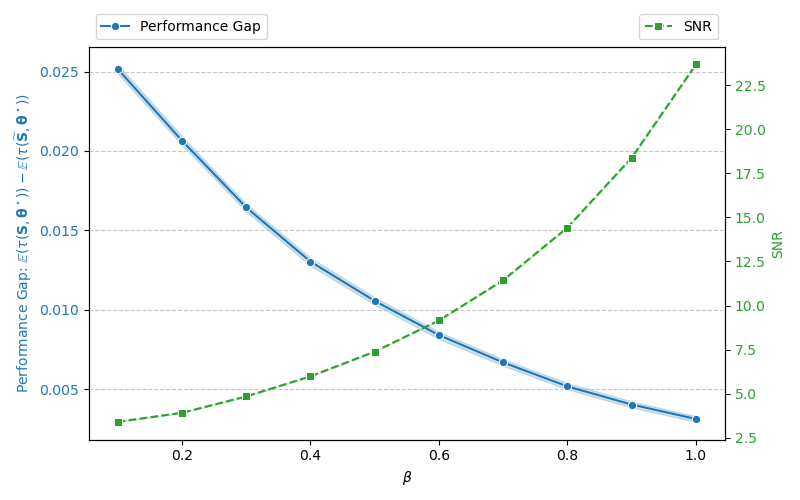}
        \caption{$\phi^{(2)}(x)$ and $\psi_{\gamma}^{(4)}(x)$}
    \end{subfigure}
            \begin{subfigure}[b]{0.323\textwidth}
        \centering
        \includegraphics[width=\textwidth]{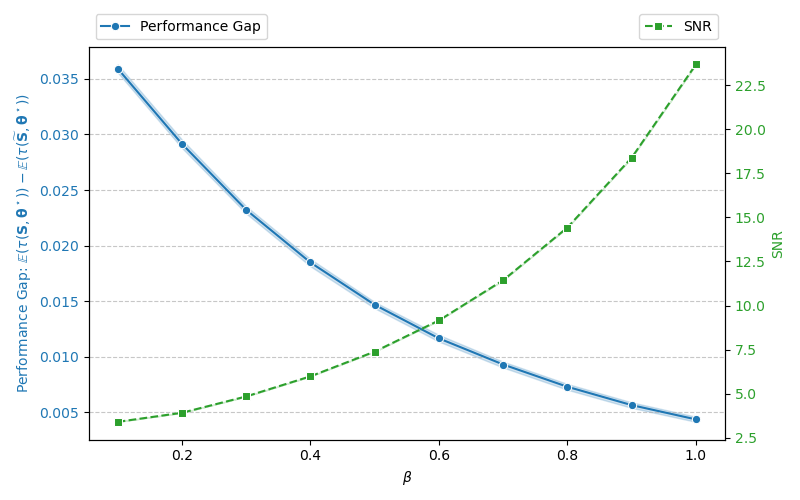}
        \caption{$\phi^{(3)}(x)$ and $\psi_{\gamma}^{(4)}(x)$}
    \end{subfigure}

    \caption{The relationship between the performance gap, $\mathbb{E}\big[\tau(\bm{S},\bm{\theta}^\star)\big] - \mathbb{E}\big[\tau(\widetilde{\bm{S}},\bm{\theta}^\star)\big]$, and the SNR under the proposed model with varying $\phi$, $\psi_{\gamma}$, and $\beta$ (x-axis).}
    \label{fig:Scen2}
\end{figure}

As shown in Figure~\ref{fig:Scen2}, the performance gap is strongly correlated with the SNR of $X_3$. When the ordinal pattern follows $\text{Geo}(\psi_{\gamma}^{(3)}, K)$, the corresponding SNR first decreases with increasing $\beta$ and then rises. Interestingly, under various specifications of $\phi$, the performance gap exhibits an inverse pattern: it first increases and then decreases, with the turning point aligning precisely with that of the SNR. A similar phenomenon is observed for $\psi_{\gamma}^{(4)}$. Specifically, When the ordinal pattern follows $\text{Geo}(\psi_{\gamma}^{(4)}, K)$, the SNR increases monotonically with $\beta$, while the performance gap follows a strictly decreasing trend. These experimental findings are consistent with our theoretical result in Theorem~\ref{ThmK_bound}, which establishes that a smaller SNR leads to a larger performance gap between ranking based on binary comparisons and that based on ordinal comparison data.

\noindent
\textbf{Scenario III.} In the third scenario, we aim to validate Theorem~\ref{ThmK_bound} by examining whether (\ref{Conver}) holds true. Specifically, we investigate whether $R(\widetilde{\bm{S}}, \bm{S})$ decreases as $L$ increases. We consider $L \in \{100+200 \times i : 0 \leq i \leq 4 \}$ and $n \in \{20,40\}$. Additionally, we explore various forms of $\phi$ and $\psi_{\gamma}$ as in Scenario I. The experimental results are presented in Figure~\ref{fig:Scen3}. 

\begin{figure}[h!]
    \centering
    \begin{subfigure}[b]{0.323\textwidth}
        \centering
        \includegraphics[width=\textwidth]{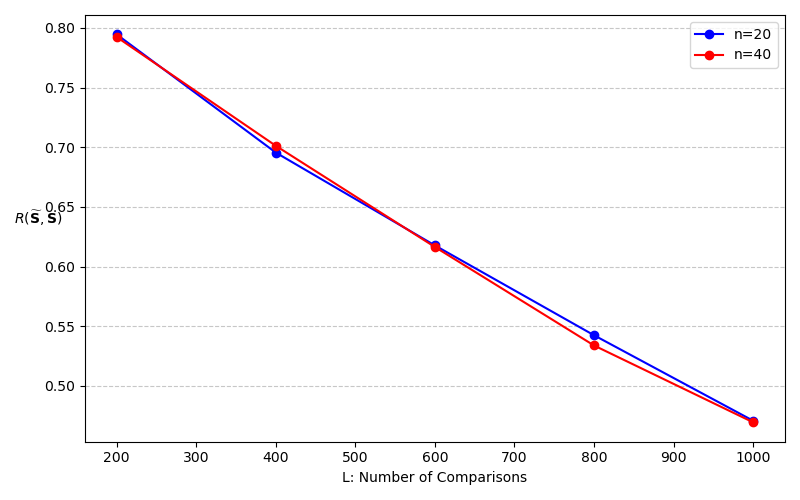}
        \caption{$\phi^{(1)}(x)$ and $\psi^{(1)}_{\gamma}(x)$}
    \end{subfigure}
        \begin{subfigure}[b]{0.323\textwidth}
        \centering
        \includegraphics[width=\textwidth]{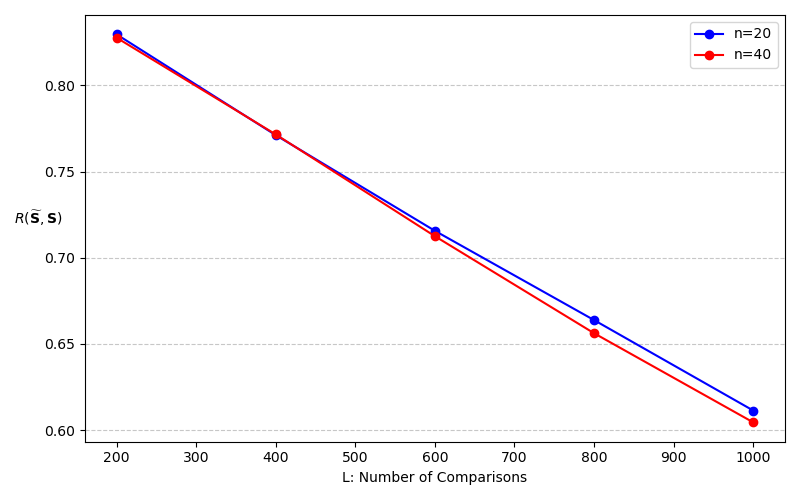}
        \caption{$\phi^{(2)}(x)$ and $\psi^{(1)}_{\gamma}(x)$}
    \end{subfigure}
        \begin{subfigure}[b]{0.323\textwidth}
        \centering
        \includegraphics[width=\textwidth]{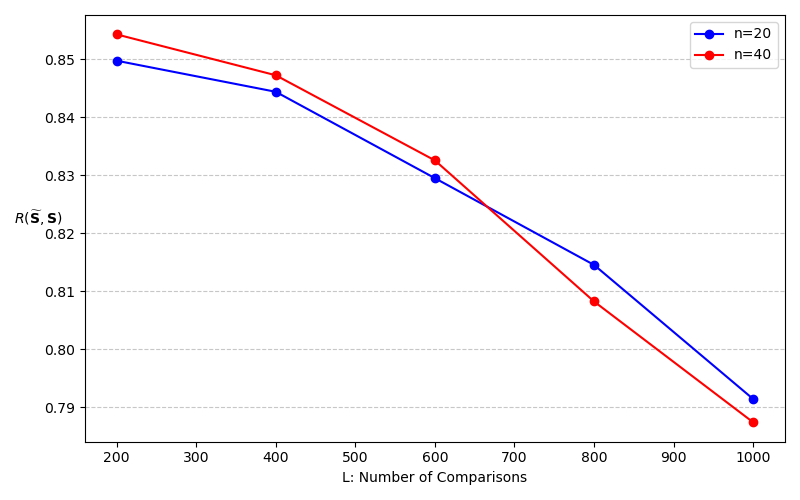}
        \caption{$\phi^{(3)}(x)$ and $\psi^{(1)}_{\gamma}(x)$}
    \end{subfigure}

        \begin{subfigure}[b]{0.323\textwidth}
        \centering
        \includegraphics[width=\textwidth]{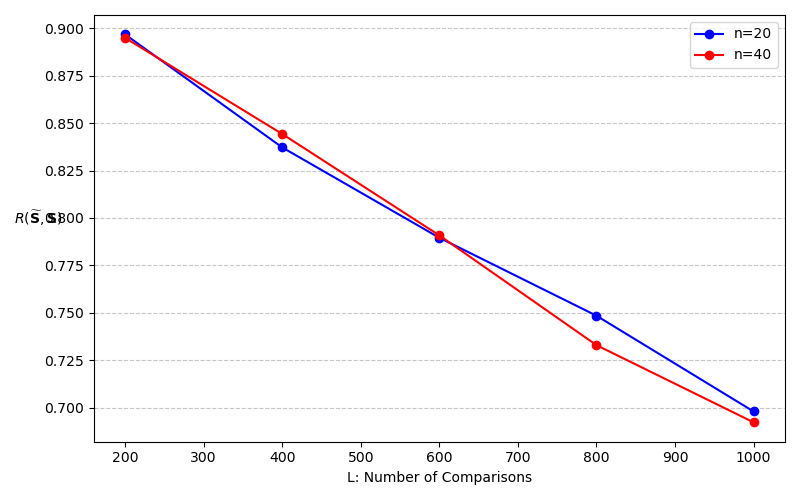}
        \caption{$\phi^{(1)}(x)$ and $\psi^{(2)}_{\gamma}(x)$}
    \end{subfigure}
        \begin{subfigure}[b]{0.323\textwidth}
        \centering
        \includegraphics[width=\textwidth]{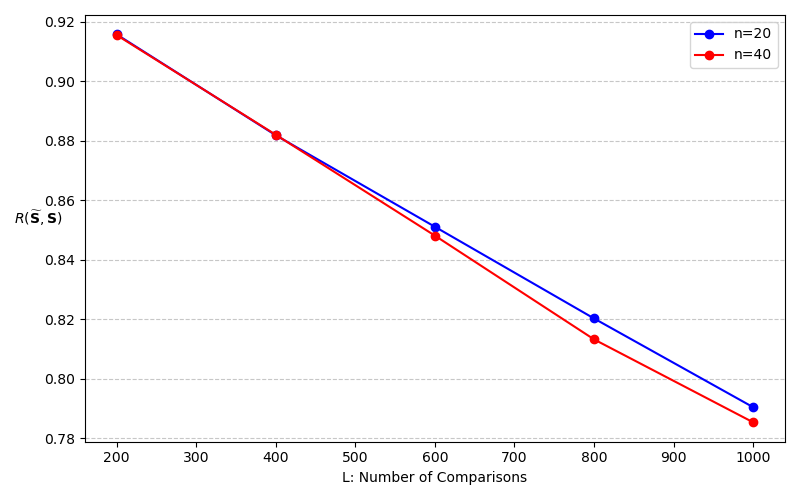}
        \caption{$\phi^{(2)}(x)$ and $\psi^{(2)}_{\gamma}(x)$}
    \end{subfigure}
            \begin{subfigure}[b]{0.323\textwidth}
        \centering
        \includegraphics[width=\textwidth]{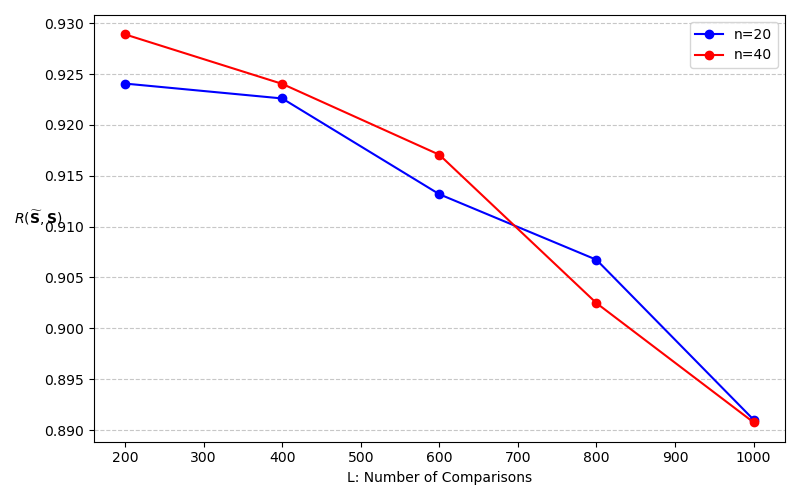}
        \caption{$\phi^{(3)}(x)$ and $\psi^{(2)}_{\gamma}(x)$}
    \end{subfigure}
    \caption{ The relationship between $R(\widetilde{\bm{S}},\bm{S})$ (y-axis) and the number of comparisons $L$ (x-axis) under the proposed model with varying $\phi$ and $\psi_{\gamma}$.}
    \label{fig:Scen3}
\end{figure}

As shown in Figure~\ref{fig:Scen3}, the results align with Theorem~\ref{ThmK_bound}, confirming that $R(\widetilde{\bm{S}}, \bm{S})$ decreases as $L$ increases. This demonstrates that rankings derived from binary comparison data converge more rapidly to the true ranking than those based on ordinal comparison data. Notably, the decreasing patterns remain nearly unchanged when $n$ increases from 20 to 40, indicating that $R(\widetilde{\bm{S}}, \bm{S})$ is largely unaffected by the number of items.
}

\subsection{Real Application}
In this section, we conducted our analysis using the MovieLens dataset \citep{harper2015movielens}, which comprises user–movie ratings collected from a large-scale movie recommendation platform and is publicly available at \url{https://www.kaggle.com/datasets/prajitdatta/movielens-100k-dataset}. Each record in the dataset includes a user identifier, a movie identifier, a discrete rating score, and a timestamp indicating when the rating was provided. To ensure statistical reliability in subsequent analyses, we focused on movies that attracted substantial user engagement. Specifically, we filtered the dataset to retain only those movies rated at least 200 times. This threshold helps exclude movies with insufficient rating data, which could introduce noise and undermine robustness. For each user in the filtered dataset, we constructed pairwise rating differences between all pairs of popular movies they had rated. Formally, for user $l$ and movies $i$ and $j$, if both movies were rated by user $l$, we define the pairwise rating difference as ordinal comparison. We retained only pairs with non-zero differences to capture meaningful preferences that distinguish between the two movies. This procedure yields a dataset of pairwise preference observations, encoding the relative strength of user preferences between pairs of popular movies.

\begin{figure}[ht]
    \centering
    \includegraphics[scale=0.29]{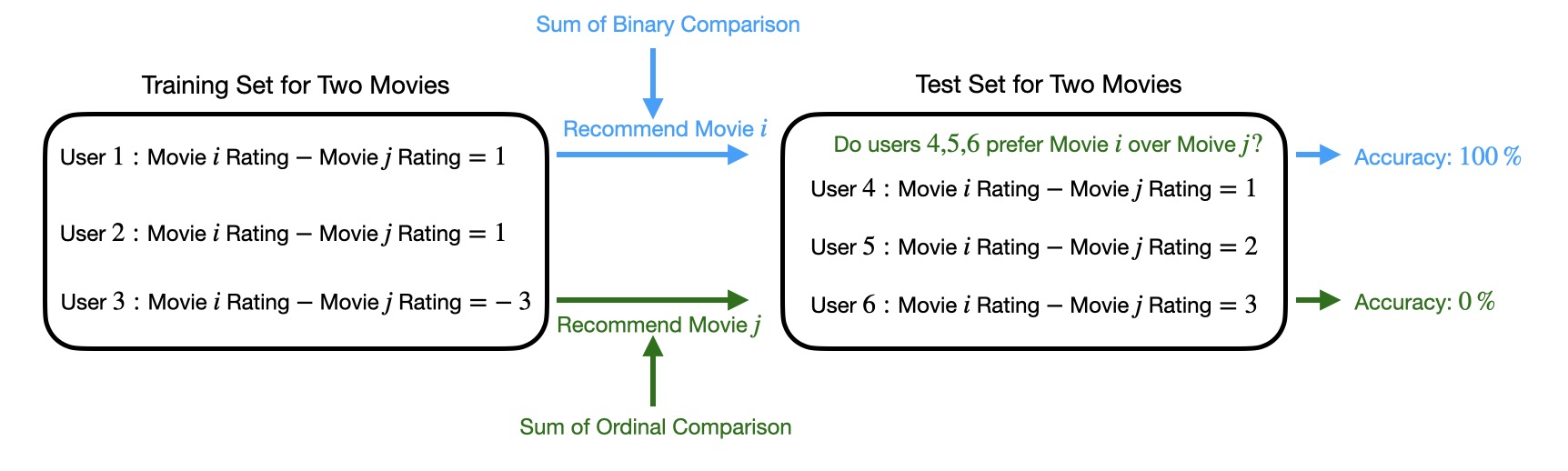}
    \caption{An illustrative example of the recommendation process based on comparison data.}
    \label{Real_App}
\end{figure}

To compare the predictive validity of learned rankings from binary and ordinal comparison data, we perform the following randomized subsampling procedure for each movie pair with sufficient comparisons. We randomly partition the observed pairwise comparisons into a training set (70\%) and a test set (30\%). From the training set, we estimate the relative preference between the two movies using two aggregation schemes:
\begin{itemize}
    \item \textbf{Ordinal Comparison Data}: compute the sum of rating differences;
    \item \textbf{Binary Comparison Data}: compute the sum of signs of rating differences, corresponding to the net win count.
\end{itemize}
The learned preference direction (i.e., the sign of the aggregated score) is then used to predict the relative preference in the test set. We evaluate prediction accuracy by computing the proportion of correctly predicted test comparisons, and repeat this procedure multiple times (100 repetitions) to obtain stable estimates of the expected prediction accuracy for each method. The general process for each pair of movies is illustrated in Figure \ref{Real_App}.

\begin{figure}[h!]
    \centering
    \begin{subfigure}[b]{0.475\textwidth}
        \centering
        \includegraphics[scale=0.45]{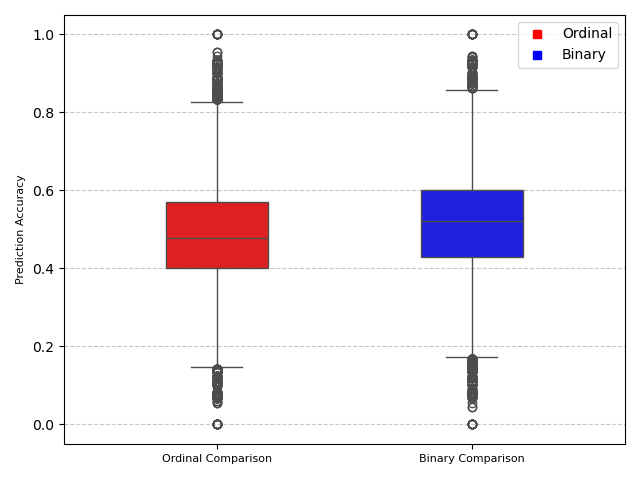}
        \caption{Boxplots of Prediction Accuracy}
    \end{subfigure}
    \hfill
    \begin{subfigure}[b]{0.475\textwidth}
        \centering
        \includegraphics[scale=0.45]{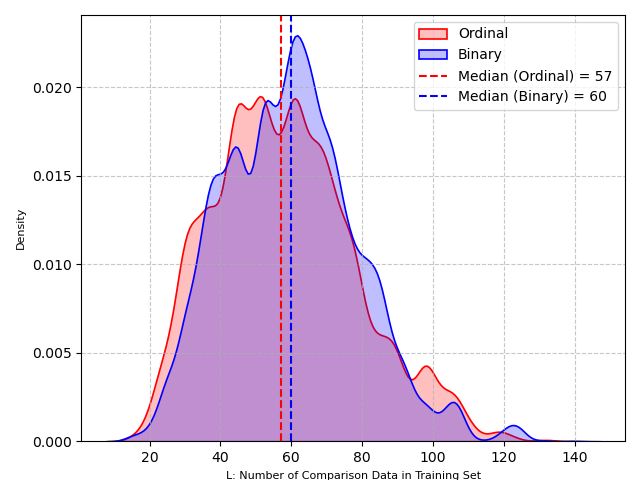}
        \caption{Kernel Density Estimates}
    \end{subfigure}
    \caption{Experimental results from the real application: (Left) Comparison of recommendation accuracies using ordinal and binary comparisons; (Right) Distributions of the number of comparisons for which each method outperforms the other.}
    \label{fig:rr}
\end{figure}

As shown in the left panel of Figure \ref{fig:rr}, using binary comparison data to infer relative preferences yields higher recommendation accuracy on the held-out users. A paired $t$-test comparing the two accuracy measures yields a $t$-statistic of 18.5978 with a $p$-value of $3.4 \times 10^{-77}$, indicating a highly significant difference. This result is consistent with our theoretical finding that binary comparison data enables more robust estimation of relative preferences.

In the right panel, we present kernel density estimates (KDE) of the distribution of the number of comparisons in cases where one method outperforms the other. Specifically, the blue curve (binary comparison) depicts the distribution of the number of comparisons in instances where the binary method yields better performance. Interestingly, the plot shows that when binary comparison data is more effective, it tends to be associated with a larger number of comparisons. This observation is also consistent with Theorem \ref{Thm:BS_bound}, which suggests that binary comparison data becomes more advantageous when sufficient comparison information is available.

\section{Summary}
\label{Sec:Sum}
This paper investigates the performance gap between ordinal and binary comparison data in ranking recovery using the counting method. To this end, we propose a general parametric framework for modeling ordinal paired comparisons without ties. When binary responses are interpreted as binarized versions of ordinal data, the framework naturally reduces to classical binary comparison models. A central finding of our study is that binarizing ordinal data can significantly improve the accuracy of ranking recovery, challenging the common intuition that ordinal comparisons carry more information than binary ones. Specifically, we show that under the counting algorithm, the ranking error associated with binary comparisons converges exponentially faster than that of ordinal data. Moreover, we demonstrate that the performance gap is determined by the pattern of ordinal levels. We identify the pattern that maximizes the benefit of binarization. Our theoretical results are further supported by extensive numerical experiments.

%\section*{Acknowledgment}
%We thank the editor, the associate editor, and the three anonymous reviewers for their valuable comments that helped improve this article.

\section*{Funding}
Jingnan Zhang's research is supported in part by National Key R$\&$D Program of China (2024YFA1012200), ``USTC Research Funds of the Double First-Class Initiative" (YD2040002020) and NSFC-12301388, and Junhui Wang's research is supported in part by HK RGC Grants GRF-11311022, GRF-14306523, GRF-14303424, and CUHK Startup Grant 4937091.

%\section*{Conflict of Interest Statement}
%The authors report there are no competing interests to declare.

\bibliographystyle{authoryear}
\putbib[Ref]
\end{bibunit}

\newpage
\baselineskip=24pt
\setcounter{page}{1}
\setcounter{equation}{0}
\setcounter{section}{0}
\setcounter{lemma}{0}
\renewcommand{\thesection}{A.\arabic{section}}
\renewcommand{\thelemma}{A\arabic{lemma}}
\renewcommand{\thetheorem}{A\arabic{theorem}}
\renewcommand{\theequation}{A\arabic{equation}}
\begin{center}
{\Large\bf Supplementary Materials} \\
\medskip
{\Large\bf ``When Less Is More: Binary Feedback Can Outperform Ordinal Comparisons in Ranking Recovery"}  \\
\bigskip
\vspace{0.2in} % comment out this line when unblind the author line
\end{center}
\bigskip

\spacingset{1.7} 

\begin{bibunit}[apalike]

In this Appendix, we provide proofs for all theoretical results presented in this paper. We begin by summarizing some necessary notations.
\begin{itemize}
    \item[1] $G(\phi,0,\gamma,1)$ is a special case of $G(\phi,\psi_{\gamma},\gamma,K)$ with $\psi_{\gamma}\equiv0$ and $K=1$. If $Y \sim G(\phi,0,\gamma,1)$, then $Y$ is a binary random variable defined as
    \begin{align*}
        \mathbb{P}(Y=k) =
        \begin{cases}
            \frac{\exp(\phi(1))}{\exp(\phi(-1))+\exp(\phi(1))},\ k=1 \\
            \frac{\exp(\phi(-1))}{\exp(\phi(-1))+\exp(\phi(1))},\ k=-1
        \end{cases}
    \end{align*}
    \item[2]For a random variable $X$, we denote by $P_X$ its probability mass function if $X$ is discrete and its density function if $X$ is continuous. We use $\mathbb{P}(\cdot)$ to represent the probability measure. For any event $A$,
$$
\mathbb{P}(X \in A) =
\begin{cases}
\displaystyle \sum_{x \in A} P_X(x), & \text{if $X$ is discrete}, \\[8pt]
\displaystyle \int_A P_X(x)\,dx, & \text{if $X$ is continuous}.
\end{cases}
$$
\end{itemize}

\section{Additional Discussions}

{\color{black}
\subsection{Extension to Ordinal Comparison Model with Ties}

Our proposed model in the main text is developed under a no-tie setting. However, it can be naturally extended to handle ties by introducing a probability $s$ for the tie event and rescaling the probabilities of all other non-tie outcomes by a factor of $1-s$. Similar extensions have been considered for the BTL model \citep{glenn1960ties, rao1967ties, davidson1970extending}. Specifically, the extended ordinal comparison model can be formulated as
\begin{align*}
\mathbb{P}\!\left(Y = k\right) = 
\begin{cases}
s, & k = 0, \\[2mm]
\displaystyle
\frac{1-s}{\Psi_{\phi,\psi_{\gamma}}(\gamma)}
\exp\!\left\{
    g(k\,|\,\phi,\psi_{\gamma},\gamma)
\right\}, & k \neq 0,
\end{cases}
\end{align*}
for any $k \in \{-K,\ldots,-1,0,1,\ldots,K\}$, 
where $\Psi_{\phi,\psi_{\gamma}}(\gamma) = \sum_{k \in \Upsilon(K)} \exp\!\left\{g(k\,|\,\phi,\psi_{\gamma},\gamma)\right\}$ is a normalizing constant.

We emphasize that all theoretical results derived for the no-tie setting remain applicable in the presence of ties. This follows from the fact that each tie can be treated as a missing comparison under the counting method, allowing us to formally connect the tie mechanism to the missingness framework. To formalize the connection between ties and missingness, we introduce an independent Bernoulli random variable $C^{(l)} \sim \mathrm{Bernoulli}(1-s)$ for each comparison $l$, indicating whether the comparison is observed (i.e., no tie occurs). We then define
\[
\widetilde{y}_{12}^{(l)} = C^{(l)} y_{12}^{(l)}, \quad l \in [L],
\]
whose distribution can be written as
\begin{align*}
\mathbb{P}\!\left(\widetilde{y}_{12}^{(l)} = k\right) = 
\begin{cases}
s, & k = 0, \\[2mm]
\displaystyle \frac{1-s}{\Psi_{\phi,\psi_{\gamma_{12}^\star}}(\gamma_{12}^\star)} \exp\!\Big\{ g(k\,|\,\phi,\psi_{\gamma_{12}^\star},\gamma_{12}^\star) \Big\}, & k \neq 0.
\end{cases}
\end{align*}
The counting method can then be applied to $\{\widetilde{y}_{12}^{(l)}\}_{l=1}^L$:
\begin{align*}
 \text{Raw Data: }  & \widetilde{A} = \frac{1}{L} \sum_{l=1}^L a_{12}^{(l)}\widetilde{y}_{12}^{(l)} = \frac{1}{L} \sum_{l=1}^L a_{12}^{(l)} C^{(l)} y_{12}^{(l)}, \\
 \text{Binarized Data: } & \widetilde{B} = \frac{1}{L} \sum_{l=1}^L a_{12}^{(l)} \widetilde{z}_{12}^{(l)} = \frac{1}{L} \sum_{l=1}^L a_{12}^{(l)} C^{(l)} z_{12}^{(l)},
\end{align*}
where $\widetilde{z}_{12}^{(l)} = \mathrm{sign}(\widetilde{y}_{12}^{(l)}) = C^{(l)} \, \mathrm{sign}(y_{12}^{(l)})$.

Since $a_{12}^{(l)}$ and $C^{(l)}$ are independent, it follows that $a_{12}^{(l)} C^{(l)} \sim \mathrm{Bernoulli}(p(1-s))$. Therefore, we can equivalently rewrite
\begin{align*}
 \widetilde{A} =  \frac{1}{L} \sum_{l=1}^L \widetilde{a}_{12}^{(l)} y_{12}^{(l)}, \quad
 \widetilde{B} =  \frac{1}{L} \sum_{l=1}^L \widetilde{a}_{12}^{(l)} z_{12}^{(l)},
\end{align*}
where $\widetilde{a}_{12}^{(l)} = a_{12}^{(l)} C^{(l)}$. Hence, for comparisons with ties, all theoretical results remain valid by replacing the original observation probability $p$ with the effective probability $p(1-s)$.
}

\subsection{Optimality of Binary Comparison Data}
In this section, we examine a general scenario in which binary comparison data can outperform other types of comparison data, whether continuous or ordinal. Specifically, we consider a random variable $W$ with symmetric support $\mathcal{W}$, meaning that $\mathbb{P}(W = -w) > 0$ whenever $\mathbb{P}(W = w) > 0$ for any $w \in \mathcal{W}$ if $W$ is discrete. In Theorem \ref{Prop:SNR}, we demonstrate that the signal-to-noise ratio of $W$ is maximized when $W$ follows a binary distribution.

\begin{theorem}
\label{Prop:SNR}
Let $W$ be a real-valued random variable with symmetric support $\mathcal{W}=\text{supp}(W)$, that is if $P_{W}(w)>0$, then we have $P_{W}(-w)>0$ with $P_W$ being the density function (continuous) or probability mass function (discrete) of $W$. Additionally, we assume that $P_{W}(w_1)/P_{W}(w_2)=P_{W}(-w_1)/P_{W}(-w_2)$ for any $w_1,w_2> 0$. Let $P = \mathbb{P}(W > 0)$. Then the signal-to-noise ratio
$$
\textnormal{SNR}(W) \triangleq \frac{(\mathbb{E}[W])^2}{\textnormal{Var}(W)}
\leq \frac{(2P-1)^2}{4P(1-P)},
$$
with equality if and only if the positive part of $W$ is concentrated at a single point.
\end{theorem}

Theorem \ref{Prop:SNR} shows that if $W$ satisfies a symmetric pattern, then its binary version achieves the maximal signal-to-noise ratio among all types of comparison data, implying that binarized $W$ is the optimal data type for ranking recovery under the counting method. 

A natural question arises: Is binary comparison data always superior to ordinal comparison data? The answer is no. To illustrate this, we provide an example. Consider a random variable $Y$ that does not satisfy the symmetric pattern assumed in Theorem~\ref{Prop:SNR}, with values and associated probabilities given as follows.
\[
\begin{array}{c|cccc}
\toprule
y & -2 & -1 & 1 & 2\\\hline
\mathbb{P}(Y=y) & 0.05 & 0.15 & 0.35 & 0.45\\
\bottomrule
\end{array}
\]
We can easily calculate the mean and the variance as $\mathbb{E}(Y)=1$ and $\text{Var}(Y)=1.5$. Therefore,
\[
\mathrm{SNR}(Y) = \frac{(\mathbb{E}[Y])^2}{\operatorname{Var}(Y)} 
= \frac{1.00^2}{1.50} 
= \frac{2}{3}.
\]
However, if we binarize $Y$ and obtain a binary comparison data, we have
\begin{align*}
\begin{array}{c|cc}
\toprule
y &  -1 & 1 \\\hline
\mathbb{P}(\sign(Y)=y) &0.2 & 0.8\\
\bottomrule
\end{array}
\end{align*}
Easily, the mean of $\sign(Y)$ is 0.6 and the variance is 0.64. Hence,
\[
\mathrm{SNR}(\sign(Y)) = \frac{(\mathbb{E}[\sign(Y)])^2}{\operatorname{Var}(\sign(Y))} 
= \frac{0.60^2}{0.64} 
= \frac{9}{16} = 0.5625<\mathrm{SNR}(Y) = \frac{2}{3}.
\]

The above derivation indicates that $Y$ conveys more information about the sign of $Y$ than $\mathrm{sign}(Y)$. To validate this finding, we consider a simple two-item comparison problem and examine $\mathbb{P}\!\left(\sum_{i=1}^L Y_i > 0\right)$ versus $\mathbb{P}\!\left(\sum_{i=1}^L \mathrm{sign}(Y_i) > 0\right)$. The results are reported in Table~\ref{Tab:AddExp}.
\begin{table}[htbp]
 
\centering
\caption{  Estimated Probabilities in $10^6$ replications for $L \in \{10,15,20,25\}$.}
\begin{tabular}{c|cccc}
\hline
$L$ &  10 & 15 & 20 & 25 \\
\hline
$\mathbb{P}(\sum_{i=1}^L Y_i > 0)$ & 0.987839 & 0.997309 & 0.999404 & 0.999856 \\
$\mathbb{P}(\sum_{i=1}^L \operatorname{sign}(Y_i) > 0)$ & 0.966959 & 0.995775 &  0.997420 &  0.999627 \\
\hline
\end{tabular}
\label{Tab:AddExp}
\end{table}
As shown in Table~\ref{Tab:AddExp}, the sum of ordinal values is more effective in recovering the ground-truth sign of $Y$. This is mainly because $Y$ does not satisfy the symmetric pattern assumed in Theorem~\ref{Prop:SNR}.

{\color{black}
\subsection{Sensitivity to Ordinal Levels}
In this experiment, we investigate the ranking performance of the counting method based on binary versus ordinal pairwise comparisons. We fix the number of items at $n = 50$ and consider repeated comparisons per pair $K$ ranging from $2$ to $10$. The latent scores $\bm{\theta}^\star$ are equally spaced with an increment of $0.05$. For each pairwise comparison, the outcomes are generated using different mechanisms: three choices of $\phi$, namely (1) $\phi^{(1)}(x) = x$, (2) $\phi^{(2)}(x) = \frac{1}{2}\log \frac{\Phi(x)}{1-\Phi(x)}$, and (3) $\phi^{(3)}(x) = \frac{1-e^{-x}}{1+e^{-x}}$, while we fix $\psi_\gamma^{(2)}(x) = -0.5x^2 + 0.5\sqrt{|x\gamma|}$. The number of comparisons per pair is fixed at $L = 100$, and the missing probability is set to $0.5$. For each combination of $(K,\phi)$, we replicate the experiment $1,000$ times. In each run, the pairwise comparisons are aggregated into scores using the counting method for both ordinal and binary outcomes. Ranking accuracy is quantified by the normalized Kendall tau distance between the aggregated scores and the true latent scores. The results, including the average ranking errors and 99\% confidence intervals, are visualized as functions of $K$ in Figure~\ref{fig:EX3}.

\begin{figure}[ht!]
    \centering
    \begin{subfigure}[b]{0.325\textwidth}
        \centering
        \includegraphics[width=\textwidth]{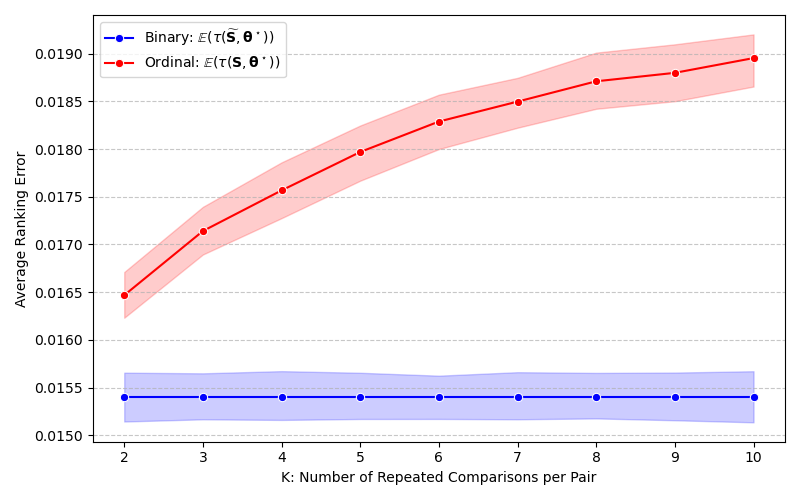}
        \caption{$\phi = \phi^{(1)}$}
    \end{subfigure}
        \begin{subfigure}[b]{0.325\textwidth}
        \centering
        \includegraphics[width=\textwidth]{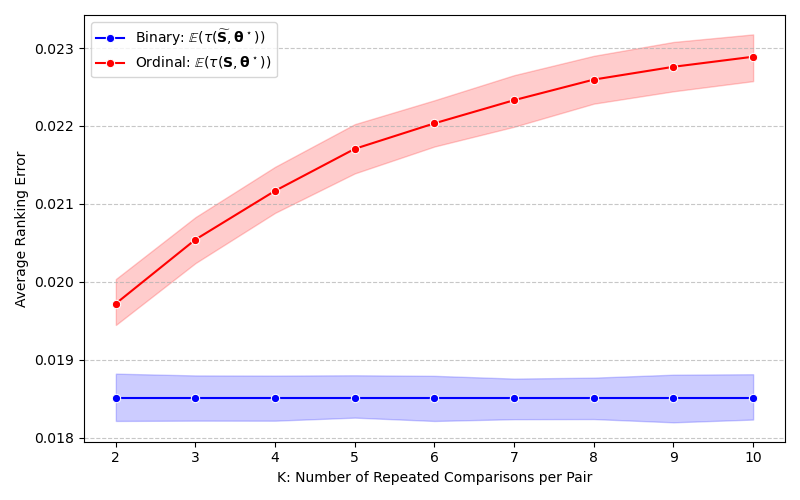}
        \caption{$\phi = \phi^{(2)}$}
    \end{subfigure}
        \begin{subfigure}[b]{0.325\textwidth}
        \centering
        \includegraphics[width=\textwidth]{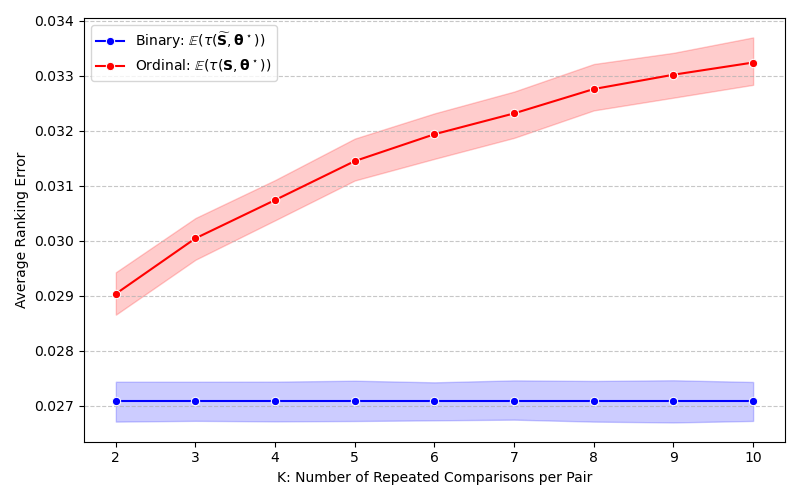}
        \caption{$\phi = \phi^{(3)}$}  
    \end{subfigure}
    \caption{A comparison between $\mathbb{E}\big[\tau(\bm{S},\bm{\theta}^\star)\big]$ and $\mathbb{E}\big[\tau(\widetilde{\bm{S}},\bm{\theta}^\star)\big]$ with varying $K$.}
    \label{fig:EX3}
\end{figure}

The experimental results show a consistent pattern across all settings. As the number of ordinal levels $K$ increases, the ranking accuracy of the ordinal method gradually decreases. In contrast, the binary method remains relatively stable. This behavior can be attributed to the increasing uncertainty introduced by larger $K$, which amplifies the variability in ordinal outcomes and consequently degrades the performance of the counting method based on ordinal comparisons.}

\section{Maximum Likelihood Estimation of Comparison Models}

\color{black}
In this section, we investigate the maximum likelihood estimation for both the ordinal and binary comparison models. We assume that each comparison outcome $y_{ij}^{(l)}$ follows the distribution
\[
y_{ij}^{(l)} \sim G(\phi, \psi_{\gamma_{ij}^\star}, \gamma_{ij}^\star, K), \quad l \in [L].
\]
For simplicity, we assume that all pairwise comparisons are observed. Recall that $\gamma_{ij} = \theta_i - \theta_j$ and $\Upsilon(K)$ denote the set of possible outcomes for each comparison. The log-likelihood of the observed data $\{y_{ij}^{(l)}: i,j = 1,\dots,n,\, l=1,\dots,L\}$ is
\begin{align*}
\mathcal{L}(\bm \theta) 
=  & \sum_{1 \le i < j \le n} \sum_{l=1}^{L} \Bigg[
\phi\big(\operatorname{sign}(y_{ij}^{(l)})(\theta_i-\theta_j) \big) +\psi_{\gamma_{ij}}(y_{ij}^{(l)}) \\
&- \log \sum_{k \in \Upsilon(K)} \exp \Big( \phi(\operatorname{sign}(k) (\theta_i-\theta_j)) + \psi_{\gamma_{ij}}(k) \Big)
\Bigg].
\end{align*}

It is worth noting that when $\phi(x)=x/2$ and $K=1$, maximizing $\mathcal{L}(\bm{\theta})$ reduces to finding the MLE under the BTL model, which has been extensively studied in the literature \citep{chen2019spectral,gao2023uncertainty}. The following theorem (Theorem \ref{Thm:Appendix}) establishes that the MLE under the general ordinal comparison model is essentially invariant to the choice of $K$ in two specific scenarios. In particular, when $\phi(x)=x/2$, the estimator $\widehat{\bm{\theta}}$ obtained from the ordinal comparison model \textbf{coincides with} that from the BTL model. Analogous results can also be derived for the Thurstone–Mosteller (TM) model.

\begin{theorem}
    \label{Thm:Appendix}
    Let $\widehat{\bm{\theta}}=\argmax\limits_{\bm{\theta}^\top \bm{1}_n=0}\mathcal{L}(\bm \theta)$ denote the maximum likelihood estimator for the ordinal comparison model, where $y_{ij}^{(l)} \sim G(\phi, \psi_{\gamma_{ij}^\star}, \gamma_{ij}^\star, K)$ for $i<j$ and $l \in [L]$.
    \begin{itemize}
        \item[(1)] If $\psi_{\gamma_{ij}}$ is well specified as $\psi_{\gamma_{ij}^\star}$ for $ i\neq j$, then $\widehat{\bm{\theta}}$ is invariant to both the value of $K$ and the form of $\psi_{\gamma_{ij}^\star}$.
        \item[(2)] If $\frac{\partial \psi_{\gamma}}{\partial \gamma}$ does not depend on $k$, then $\widehat{\bm{\theta}}$ is invariant to both the value of $K$ and the form of $\psi_{\gamma_{ij}^\star}$.
    \end{itemize}
\end{theorem}

In Theorem \ref{Thm:Appendix}, $\bm\theta^\top\bm{1}_n=0$ is used to solve the identifiability issue of $\bm\theta^\star$. A direct implication of Theorem \ref{Thm:Appendix} is that using ordinal comparison data or its binarized counterpart for MLE yields the same estimator, and thus both approaches have identical estimation efficiency.

When $\psi_{\gamma_{ij}^\star}$'s are known, the estimator $\widehat{\bm{\theta}}$ does not depend on the number of ordinal levels $K$ or the specific functional form of $\psi_{\gamma_{ij}^\star}$. This case implies that if the ordinal patterns are correctly specified, the MLE is robust to the choice of $K$. In the second case, we explicitly consider $\psi_{\gamma_{ij}^\star}$ is unknown and $\psi_{\gamma}$ is treated as a function of $\gamma$ during the optimization. Interestingly, the MLE remains invariant to $K$ as long as $\frac{\partial \psi_{\gamma}}{\partial \gamma}$ does not depend on the outcome level $k$. This scenario includes functions of the form $\psi_{\gamma}(k) = f(\gamma) + g(k)$, where $f$ depends only on $\gamma$ and $g$ depends only on the ordinal level $k$. Practically, this means that $\psi_{\gamma}$ is specified so that $\partial \psi_{\gamma} / \partial \gamma$ does not depend on $k$, and consequently the MLE is driven primarily by the corresponding binary comparison model.

\color{black}

\begin{figure}[ht!]
    \centering
    \begin{subfigure}[b]{0.323\textwidth}
        \centering
        \includegraphics[width=\textwidth]{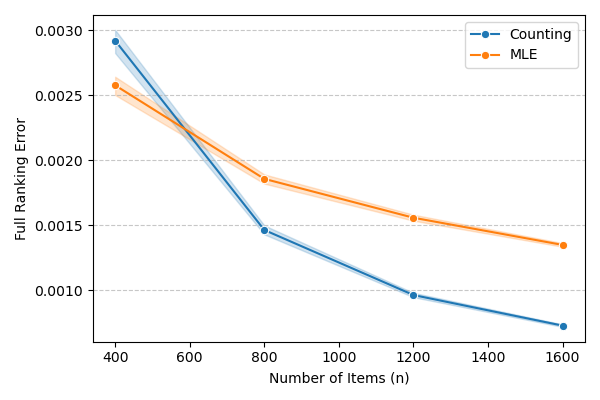}
        \caption{Ranking Error: $L=10$}
    \end{subfigure}
        \begin{subfigure}[b]{0.323\textwidth}
        \centering
        \includegraphics[width=\textwidth]{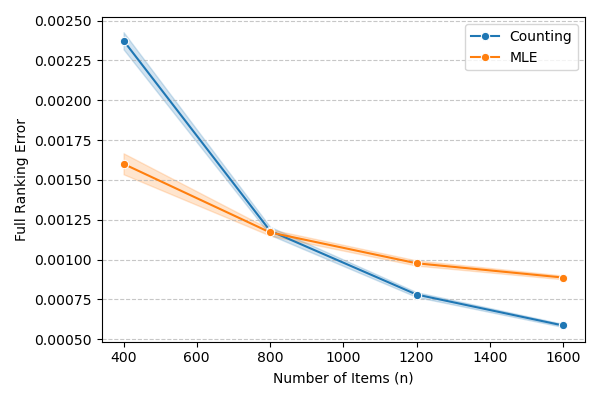}
        \caption{Ranking Error: $L=20$}
    \end{subfigure}
        \begin{subfigure}[b]{0.323\textwidth}
        \centering
        \includegraphics[width=\textwidth]{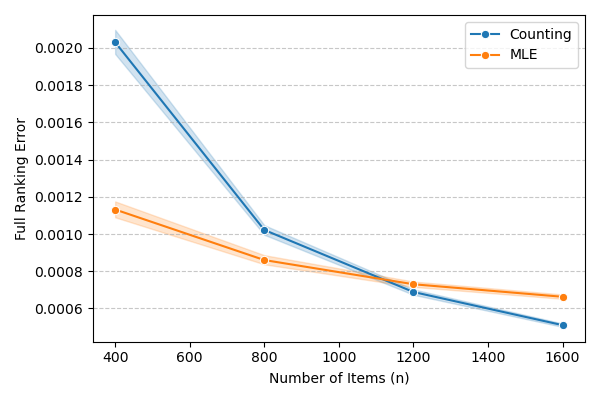}
        \caption{Ranking Error: $L=30$}
    \end{subfigure}

        \begin{subfigure}[b]{0.323\textwidth}
        \centering
        \includegraphics[width=\textwidth]{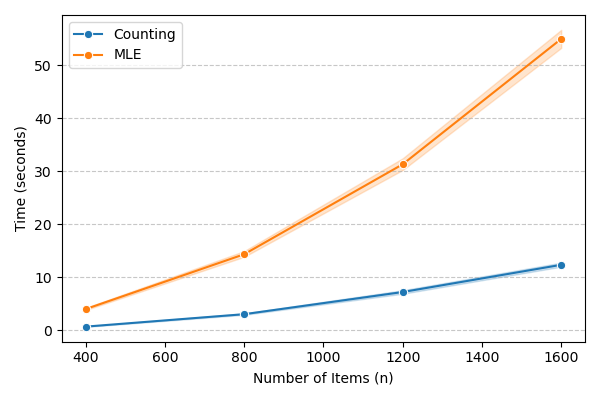}
        \caption{Time: $L=10$}
    \end{subfigure}
        \begin{subfigure}[b]{0.323\textwidth}
        \centering
        \includegraphics[width=\textwidth]{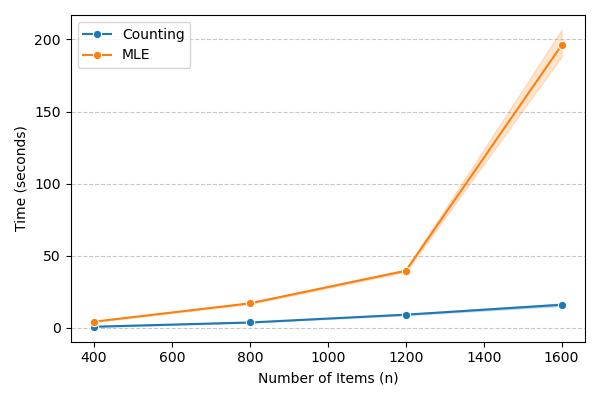}
        \caption{Time: $L=20$}
    \end{subfigure}
            \begin{subfigure}[b]{0.323\textwidth}
        \centering
        \includegraphics[width=\textwidth]{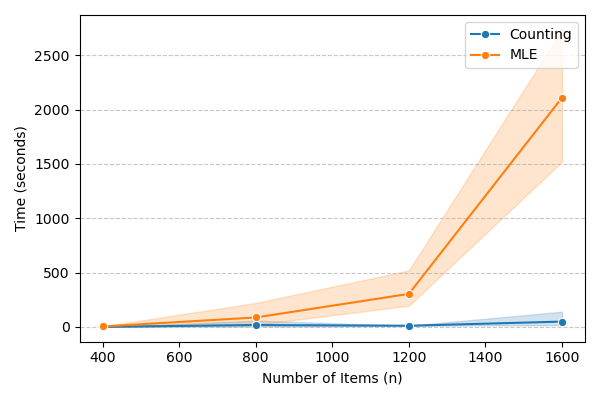}
        \caption{Time: $L=30$}
    \end{subfigure}

    \caption{A comparison between the counting method and the MLE with respect to full ranking error and computational time.}
    \label{fig:App2}
\end{figure}

The remaining question is which method (counting method or MLE) is more effective in recovering the true ranking of items from binary comparison data. Since the MLE does not admit a closed-form solution, we primarily compare the two methods empirically, following the approach of \citet{shah2018simple}. Specifically, we consider the setting of large scale items and few users providing comparison data. We set $n \in \{400,800,1200,1600\}$ and $L \in \{10,20,30\}$, which is commonly encountered in the domain of recommender systems \citep{negahban2012iterative}. We replicate each case 30 times and report the averaged full ranking error and computational times in Figure \ref{fig:App2}.

As shown in Figure \ref{fig:App2}, the counting method significantly outperforms the MLE when $n \gg L$. As the number of items increases, the full ranking error of the counting method exhibits a faster convergence rate, whereas the MLE is less sensitive to this growth. In contrast, the MLE outperforms the counting method as the number of users increases when the number of items $n$ is small. These results suggest that each method has its own comfort zone for full ranking recovery. In stark contrast, in terms of computational time, the MLE method incurs a heavy computational cost, especially for large $n$. This observation aligns with \citet{shah2018simple}, who noted that the MLE is less efficient for large-scale comparison graphs. Consequently, this computational burden limits the applicability of the MLE in large-scale settings.

\clearpage
\section{Proof of Theorems}

\subsection{Proof of Theorem \ref{Thm:MeanVari}}
\textbf{Proof of Property (1)}: From the definition of $G(\phi,\psi_{\gamma},\gamma,K)$, we have
%We define the probability mass function as
\begin{align*}
\mathbb{P}(Y = k) = \frac{1}{\Psi_{\phi,\psi_{\gamma}}(\gamma)} \exp\left( g(k\,|\,\phi,\psi_{\gamma},\gamma) \right), \quad k \in \Upsilon(K) \triangleq \{-K, \ldots, -1, 1, \ldots, K\},
\end{align*}
where $g(k\,|\,\phi,\psi_{\gamma},\gamma) = \phi(\operatorname{sign}(k)\gamma) +\psi_{\gamma}(|k|)$.

To compute $\mathbb{P}(Y > 0)$, we sum over all positive values of $k$:
\begin{align*}
\mathbb{P}(Y > 0) &= \sum_{k=1}^{K} \mathbb{P}(Y = k) = \sum_{k=1}^{K} \frac{1}{\Psi_{\phi,\psi_{\gamma}}(\gamma)} \exp\left( \phi(\gamma) +\psi_{\gamma}(k) \right) \\
&= \frac{1}{\Psi_{\phi,\psi_{\gamma}}(\gamma)} \exp(\phi(\gamma)) \sum_{k=1}^{K} \exp(\psi_{\gamma}(k)).
\end{align*}
The normalizing constant is
\begin{align*}
\Psi_{\phi,\psi_{\gamma}}(\gamma) &= \sum_{k=-K}^{-1} \exp\left( \phi(-\gamma) +\psi_{\gamma}(-k) \right) + \sum_{k=1}^{K} \exp\left( \phi(\gamma) +\psi_{\gamma}(k)\right) \\
&= \exp(\phi(-\gamma))\sum_{k=1}^{K} \exp(\psi_{\gamma}(k)) + \exp(\phi(\gamma)) \sum_{k=1}^{K} \exp(\psi_{\gamma}(k)) \\
&= \left[ \exp(\phi(-\gamma)) + \exp(\phi(\gamma)) \right] \sum_{k=1}^{K} \exp(\psi_{\gamma}(k)).
\end{align*}
Therefore, we conclude that
\begin{align*}
\mathbb{P}(Y > 0) = \frac{\exp(\phi(\gamma))}{\exp(\phi(\gamma)) + \exp(\phi(-\gamma))} = 
\frac{\exp(2\phi(\gamma))}{\exp(2\phi(\gamma) + 1)}
\end{align*}
According to the definition of $G(\phi,\psi_{\gamma},\gamma,K)$, $\sign(Y)$ follows the following distribution
\begin{align*}
    \mathbb{P}(\sign(Y)=1) = \frac{\exp(2\phi(\gamma))}{\exp(2\phi(\gamma))+1}.
\end{align*}
Therefore, $\sign(Y)$ can be viewed as following $G(\phi,0,\gamma,1)$. This then implies the desired result. 

\noindent
\textbf{Proof of Property (2)}: Recall that $\psi_{\gamma}(k)$ is assumed to be an even function in both $\gamma$ and $k$. Next, by the definition of $G(\phi,\psi_{\gamma},\gamma,K)$, we have
\begin{align*}
&\mathbb{P}(-Y=k)=\mathbb{P}(Y=-k)\\
= &
 \left[\Psi_{\phi,\psi_{\gamma}}(\gamma)\right]^{-1}\exp(g(-k\,|\,\phi,\psi_{\gamma},\gamma))\\
 = & \left[\Psi_{\phi,\psi_{\gamma}}(\gamma)\right]^{-1}\exp(g(k\,|\,\phi,\psi_{-\gamma},-\gamma)) \\
 = & \left[\Psi_{\phi,\psi_{-\gamma}}(-\gamma)\right]^{-1}\exp(g(k\,|\,\phi,\psi_{-\gamma},-\gamma)),
\end{align*}
where the last equality follows from the property that $g(-k\,|\,\phi,\psi_{\gamma},\gamma)=g(k\,|\,\phi,\psi_{-\gamma},-\gamma)$.

\noindent
\textbf{Proof of Property (3)}: We first derive the expectation of $Y$.
\begin{align*}
\mathbb{E}(Y) &= \sum_{k \in \Upsilon(K)} k \cdot \mathbb{P}(Y = k) = \frac{1}{\Psi_{\phi,\psi_{\gamma}}(\gamma)} \sum_{k \in \Upsilon(K)} k \cdot \exp\left( g(k\,|\,\phi,\psi_{\gamma},\gamma) \right) \\
&= \frac{1}{\Psi_{\phi,\psi_{\gamma}}(\gamma)} \left[ \sum_{k=1}^K k \cdot \exp\left( \phi(\gamma) +\psi_{\gamma}( k) \right) + \sum_{k=1}^K (-k) \cdot \exp\left( \phi(-\gamma) + \psi_{\gamma}(k) \right) \right] \\
&= \frac{1}{\Psi_{\phi,\psi_{\gamma}}(\gamma)} \sum_{k=1}^{K} k \left[ \exp\left( \phi(\gamma) + \psi_{\gamma}(k) \right) - \exp\left( \phi(-\gamma) + \psi_{\gamma}(k) \right) \right].
\end{align*}
Note that $\psi_{\gamma}(k)$ is an even function. We factor out $\exp(\phi(\gamma))$ and $\exp(\phi(-\gamma))$ and obtain
\begin{align*}
\mathbb{E}(Y) = \frac{1}{\Psi_{\phi,\psi_{\gamma}}(\gamma)} \left[ \exp(\phi(\gamma)) - \exp(\phi(-\gamma)) \right] \sum_{k=1}^{K} k \exp(\psi_{\gamma}(k)).
\end{align*}
Using the expression for the normalizing constant
\begin{align*}
\Psi_{\phi,\psi_{\gamma}}(\gamma) = \left[ \exp(\phi(\gamma)) + \exp(\phi(-\gamma)) \right] \sum_{k=1}^{K} \exp(\psi_{\gamma}(k)),
\end{align*}
we conclude that
\begin{align*}
\mathbb{E}(Y) = \frac{ \left[ \exp(\phi(\gamma)) - \exp(\phi(-\gamma)) \right] \cdot \sum_{k=1}^{K} k \exp(\psi_{\gamma}(k)) }{ \left[ \exp(\phi(\gamma)) + \exp(\phi(-\gamma)) \right] \cdot \sum_{k=1}^{K} \exp(\psi_{\gamma}(k)) }= \tanh(\phi(\gamma)) \cdot \frac{\sum_{k=1}^{K} k \exp(\psi_{\gamma}(k))}{\sum_{k=1}^{K} \exp(\psi_{\gamma}(k))}.
\end{align*}

Next, we turn to calculate $\text{Var}(Y)$. We first compute $\mathbb{E}(Y^2)$.
\begin{align*}
\mathbb{E}(Y^2) &= \sum_{k \in \Upsilon(K)} k^2 \cdot \mathbb{P}(Y = k) = \frac{1}{\Psi_{\phi,\psi_{\gamma}}(\gamma)} \sum_{k \in \Upsilon(K)} k^2 \cdot \exp\left( g(k\,|\,\phi,\psi_{\gamma},\gamma) \right) \\
&= \frac{1}{\Psi_{\phi,\psi_{\gamma}}(\gamma)} \left[ \sum_{k=1}^K k^2 \exp\left( \phi(\gamma) +\psi_{\gamma}(k) \right) + \sum_{k=1}^K k^2 \exp\left( \phi(-\gamma) +\psi_{\gamma}(k) \right) \right] \\
&= \frac{ \left[ \exp(\phi(\gamma)) + \exp(\phi(-\gamma)) \right] \cdot \sum_{k=1}^K k^2 \exp(\psi_{\gamma}(k)) }{ \left[ \exp(\phi(\gamma)) + \exp(\phi(-\gamma)) \right] \cdot \sum_{k=1}^K \exp(\psi_{\gamma}(k)) } \\%\cdot \sum_{k=1}^K \exp(\psi_{\gamma}(k)) \\
&= \frac{ \sum_{k=1}^K k^2 \exp(\psi_{\gamma}(k)) }{ \sum_{k=1}^K \exp(\psi_{\gamma}(k)) }.
\end{align*}
The expression for $\mathbb{E}(Y^2)$ is symmetric in $\gamma$, because $k^2$ is even and cancels the asymmetry of $\phi(\gamma)$ vs $\phi(-\gamma)$. Recall that
\[
\mathbb{E}(Y) = \frac{ \left[ \exp(\phi(\gamma)) - \exp(\phi(-\gamma)) \right] \cdot \sum_{k=1}^K k \exp(\psi_{\gamma}(k)) }{ \left[ \exp(\phi(\gamma)) + \exp(\phi(-\gamma)) \right] \cdot \sum_{k=1}^K \exp(\psi_{\gamma}(k)) }.
\]
Then, the variance is given as
\begin{align*}
\operatorname{Var}(Y)
= & \frac{ \sum_{k=1}^K k^2 \exp(\psi_{\gamma}(k)) }{ \sum_{k=1}^K \exp(\psi_{\gamma}(k)) }
- \left( \frac{ \left[ \exp(\phi(\gamma)) - \exp(\phi(-\gamma)) \right] \cdot \sum_{k=1}^K k \exp(\psi_{\gamma}(k)) }{ \left[ \exp(\phi(\gamma)) + \exp(\phi(-\gamma)) \right] \cdot \sum_{k=1}^K \exp(\psi_{\gamma}(k)) } \right)^2 \\
= &
\frac{ \sum_{k=1}^K k^2 \exp(\psi_{\gamma}(k)) }{ \sum_{k=1}^K \exp(\psi_{\gamma}(k)) }-
\left(
\tanh(\phi(\gamma)) \cdot \frac{\sum_{k=1}^{K} k \exp(\psi_{\gamma}(k))}{\sum_{k=1}^{K} \exp(\psi_{\gamma}(k))}
\right)^2.
\end{align*}
\textbf{Derivation of SNR$(Y)$.} We define
\[
\mu \triangleq \frac{\sum_{k =1 }^K k e^{\psi_{\gamma}(k)}}{\sum_{k =1 }^K e^{\psi_{\gamma}(k)}}, \quad 
\sigma^2 \triangleq \frac{\sum_{k =1 }^K k^2 e^{\psi_{\gamma}(k)}}{\sum_{k =1 }^K e^{\psi_{\gamma}(k)}} - \mu^2.
\]
Then the expectation and variance of $Y$ are given by
\[
\mathbb{E}(Y) = \tanh(\phi(\gamma)) \cdot \mu, \quad 
\mathrm{Var}(Y) = \sigma^2 + \mu^2 (1 - \tanh^2(\phi(\gamma))).
\]
Hence, the signal-to-noise ratio (SNR) of $Y$ is
\[
\mathrm{SNR}(Y) 
= \frac{\mathbb{E}(Y)^2}{\mathrm{Var}(Y)} 
= \frac{\mu^2 \tanh^2(\phi(\gamma))}{\sigma^2 + \mu^2 (1 - \tanh^2(\phi(\gamma)))}.
\]
Using the identity $\mathrm{SNR}(X_{\gamma}) \triangleq \mu^2 / \sigma^2$, we can rewrite the above as
\[
\mathrm{SNR}(Y) 
= \frac{\mathrm{SNR}(X_{\gamma}) \cdot \tanh^2(\phi(\gamma))}{1 + \mathrm{SNR}(X_{\gamma})(1 - \tanh^2(\phi(\gamma)))}.
\]
Equivalently,
\[
\mathrm{SNR}(Y) 
= \frac{\tanh^2(\phi(\gamma))}{\frac{1}{\mathrm{SNR}(X_{\gamma})} + 1 - \tanh^2(\phi(\gamma))}.
\]
When $K  = 1$, then $\mathbb{E}(Y)$ and $\text{Var}(Y)$ become
\begin{align*}
    \mathbb{E}(Y) = \tanh(\phi(\gamma)) \,\mbox{ and }\,
    \text{Var}(Y) = 1 - \tanh^2(\phi(\gamma)).
\end{align*}
The signal-to-noise ratio is then given as
\begin{align*}
    \frac{[\mathbb{E}(Y)]^2}{\text{Var}(Y)} = \frac{\tanh^2(\phi(\gamma))}{1 - \tanh^2(\phi(\gamma))} = \sinh^2(\phi(\gamma)).
\end{align*}
Since $\text{SNR}(X_\gamma)$ is always non-negative, $\sinh^2(\phi(\gamma))$ is the maximal value. This completes the whole proof. \hfill ${\color{red}\blacksquare}$ \\

\noindent
\textbf{Proof of Theorem \ref{Thm:Generalization}.} The proof of Theorem \ref{Thm:Generalization} is mainly based on the result of Lemma \ref{Lemma:CDF}. First, we establish the relationship between the proposed model and the Bradley-Terry-Luce (BTL) model. Suppose that $Y_{ij} \sim G(\phi,\psi_{\gamma_{ij}^\star},\gamma_{ij}^\star,1)$ and $\phi(x)=\frac{1}{2}\log\left(\frac{\sigma(x)}{1-\sigma(x)}\right)$. Therefore, $g(k\,|\,\phi,\psi_{\gamma_{ij}^\star},\gamma_{ij}^\star)$ is given as
$$
g(k\,|\,\phi,\psi_{\gamma_{ij}^\star},\gamma_{ij}^\star) = 
\frac{1}{2}\log\left(\frac{\sigma\left(\sign(k)(\theta_i^\star-\theta_j^\star)\right)}{1-\sigma\left(\sign(k)(\theta_i^\star-\theta_j^\star)\right)}\right)+\psi_{\gamma_{ij}^\star}(k) =  \frac{\sign(k)(\theta_i^\star-\theta_j^\star)}{2}+\psi_{\gamma_{ij}^\star}(k),
$$
for $k \in \{-1,1\}$. Then we can derive the following formula:
\begin{align*}
    \mathbb{P}(Y_{ij} = 1) = \frac{e^{\frac{\theta_i^\star-\theta_j^\star}{2}+\psi_{\gamma_{ij}^\star}(1)}}{e^{\frac{\theta_i^\star-\theta_j^\star}{2}+\psi_{\gamma_{ij}^\star}(1)}+e^{-\frac{\theta_i^\star-\theta_j^\star}{2}+\psi_{\gamma_{ij}^\star}(1)}}=
    \frac{e^{\theta_i^\star-\theta_j^\star}}{1+e^{\theta_i^\star-\theta_j^\star}}.
\end{align*}

Next, we establish the relationship between the proposed model and the Thurstone-Mosteller model. Given that $\phi(x) =\frac{1}{2}\log\left(\frac{\Phi(x)}{1-\Phi(x)}\right)$ with $\Phi(x)=\int_{-\infty}^x (2\pi)^{-1/2}e^{-x^2/2}dx$, we have
\begin{align*}
    \mathbb{P}(Y_{ij} = 1) =& \frac{\sqrt{\frac{\Phi(\theta_i^\star-\theta_j^\star)}{1-\Phi(\theta_i^\star-\theta_j^\star)}}\cdot e^{\psi_{\gamma_{ij}^\star}(1)}}{\sqrt{\frac{\Phi(\theta_i^\star-\theta_j^\star)}{1-\Phi(\theta_i^\star-\theta_j^\star)}}\cdot e^{\psi_{\gamma_{ij}^\star}(1)}+\sqrt{\frac{1-\Phi(\theta_i^\star-\theta_j^\star)}{\Phi(\theta_i^\star-\theta_j^\star)}}\cdot e^{\psi_{\gamma_{ij}^\star}(1)}} \\
    = &
     \frac{\Phi(\theta_i^\star-\theta_j^\star)}{\Phi(\theta_i^\star-\theta_j^\star)+1-\Phi(\theta_i^\star-\theta_j^\star)}=\Phi(\theta_i^\star-\theta_j^\star).
\end{align*}
This completes the proof. \hfill ${\color{red}\blacksquare}$ \\

\noindent
\textbf{Proof of Theorem \ref{Thm:Compare}.} We aim to compare the probabilities $\mathbb{P}(A > 0)$ and $\mathbb{P}(B > 0)$ in the limit as $L \to \infty$. For each $l \in [L]$, the comparison outcome $y_{12}^{(l)}$ is drawn i.i.d. from the discrete distribution $G(\phi, \psi_{\gamma_{12}^\star}, \gamma_{12}^\star,K)$, where $\gamma_{12}^\star > 0$ by assumption. By assumption, $y_{12}^{(l)} \in \{-K, \dots, -1, 1, \dots, K\}$ and never takes the value zero.

\vspace{0.5em}
\noindent\textbf{1. Asymptotic behavior of $\mathbb{P}(B > 0)$:} Let $z_{12}^{(l)} = \sign(y_{12}^{(l)}) \in \{-1, 1\}$ denote the binarized outcome. Define the sample average
\[
B = \frac{1}{L} \sum_{l=1}^L a_{12}^{(l)} z_{12}^{(l)}, \quad \mu_B = \mathbb{E}[a_{12}^{(l)}\cdot z_{12}^{(l)}] =p\left[ \mathbb{P}(y_{12}^{(l)} > 0) - \mathbb{P}(y^{(l)}_{12} < 0)\right].
\]
Because $y \ne 0$, this simplifies to
\begin{align*}
    \mu_B =& p \cdot \frac{\sum_{k=1}^{K} e^{\phi(\gamma_{12}^\star) +\psi_{\gamma_{12}^\star}(k)} - \sum_{k=1}^{K} e^{-\phi(\gamma_{12}^\star)+\psi_{\gamma_{12}^\star}(k)}}{2 \sum_{k=1}^{K} e^{\psi_{\gamma_{12}^\star}(k)} \cosh\left(\phi (\gamma_{12}^\star)\right)}
 \\
 = & p \cdot \frac{\sum_{k=1}^{K} e^{\psi_{\gamma_{12}^\star}(k)} \sinh\left(\phi(\gamma_{12}^\star)\right)}{\sum_{k=1}^{K} e^{\psi_{\gamma_{12}^\star}(k)} \cosh\left( \phi (\gamma_{12}^\star)\right)}
= p \cdot \tanh\left(\phi(\gamma_{12}^\star)\right),
\end{align*}
where we utilize the facts that $\phi(\cdot)$ is an odd function and $\psi_{\gamma_{12}^\star}(\cdot)$ is an even function. 

Note that $\mathbb{E}[(z_{12}^{(l)})^2]=1$. Applying the central limit theorem (CLT), as $L \to \infty$,
\[
\sqrt{L}(B - \mu_B) \overset{d}{\longrightarrow} \mathcal{N}(0, \sigma_B^2), 
\]
where $\sigma_B^2 = \mathrm{Var}(a_{12}^{(l)}z_{12}^{(l)}) = p - \mu_B^2=p-p^2\tanh^2\left(\phi(\gamma_{12}^\star)\right)$. Hence,
\[
\mathbb{P}(B > 0)  \xrightarrow{L \rightarrow\infty} \Phi\left( \frac{\sqrt{L} \mu_B}{\sqrt{p - \mu_B^2}} \right) = 
\Phi\left(  \frac{\sqrt{Lp}\tanh(\phi(\gamma_{12}^\star))}{\sqrt{1-\tanh^2\left(\phi(\gamma_{12}^\star)\right)+(1-p)\tanh^2\left(\phi(\gamma_{12}^\star)\right)}} \right).
\]
Note that $\csch^2(x)=\frac{1-\tanh^2(x)}{\tanh^2(x)}$ if $x>0$. Therefore, we have
\begin{align*}
    \mathbb{P}(B > 0) \xrightarrow{L \rightarrow\infty}
    \Phi\left(  \sqrt{\frac{Lp}{\csch^2(\phi(\gamma_{12}^\star))+(1-p)}} \right).
\end{align*}

\vspace{0.5em}
\noindent\textbf{2. Asymptotic behavior of $\mathbb{P}(A > 0)$:} Let $A = \frac{1}{L} \sum_{l=1}^L a_{12}^{(l)}y_{12}^{(l)}$ be the average of the raw comparison outcomes. By Theorem \ref{Thm:MeanVari}, we have
\begin{align*}
    \mu_A &=  \mathbb{E}(a_{12}^{(l)}y_{12}^{(l)})= p\cdot
\tanh(\phi(\gamma_{12}^\star)) \cdot \frac{\sum_{k =1 }^K k e^{\psi_{\gamma_{12}^\star}(k)}}{\sum_{k =1 }^K e^{\psi_{\gamma_{12}^\star}(k)}},  \\
\sigma_A^2 &
= \text{Var}(a_{12}^{(l)}y_{12}^{(l)})= p\cdot
\frac{\sum_{k=1}^K k^2 e^{\psi_{\gamma_{12}^\star}(k)}}{\sum_{k=1}^K e^{\psi_{\gamma_{12}^\star}(k)}} 
- \left(p\cdot
\tanh\big(\phi(\gamma_{12}^\star)\big) \cdot \frac{\sum_{k=1}^K k e^{\psi_{\gamma_{12}^\star}(k)}}{\sum_{k=1}^K e^{\psi_{\gamma_{12}^\star}(k)}}
\right)^2 \\
&= p \left[\frac{\sum_{k=1}^K k^2 e^{\psi_{\gamma_{12}^\star}(k)}}{\sum_{k=1}^K e^{\psi_{\gamma_{12}^\star}(k)}} 
- \left(
\tanh\big(\phi(\gamma_{12}^\star)\big) \cdot \frac{\sum_{k=1}^K k e^{\psi_{\gamma_{12}^\star}(k)}}{\sum_{k=1}^K e^{\psi_{\gamma_{12}^\star}(k)}}
\right)^2\right]\\
&+p(1-p)\left(
\tanh\big(\phi(\gamma_{12}^\star)\big) \cdot \frac{\sum_{k=1}^K k e^{\psi_{\gamma_{12}^\star}(k)}}{\sum_{k=1}^K e^{\psi_{\gamma_{12}^\star}(k)}}
\right)^2.
\end{align*}
By the CLT, we have
\[
\sqrt{L}(A - \mu_A) \overset{d}{\longrightarrow} \mathcal{N}(0, \sigma_A^2).
\]
Therefore,
\[
\mathbb{P}(A > 0)  \xrightarrow{L \rightarrow\infty} \Phi\left( \frac{\sqrt{L} \mu_A}{\sigma_A} \right).
\]
Thus,
\[
\mathbb{P}(A > 0) \xrightarrow{L \rightarrow\infty} \Phi\left(\sqrt{\frac{Lp}{\frac{1}{\mathrm{SNR}(X_{\gamma_{12}^\star})\tanh^2(\phi(\gamma_{12}^\star))}+\csch^2(\phi(\gamma_{12}^\star))+1-p}} \right).
\]
Given that $\mathrm{SNR}(X_{\gamma_{12}^\star}) > 0$ with $X_{\gamma_{12}^\star}$ being non-degenerate and $\gamma_{12}^\star>0$, it holds that
\begin{align*}
\frac{Lp}{\frac{1}{\mathrm{SNR}(X_{\gamma_{12}^\star})\tanh^2(\phi(\gamma_{12}^\star))}+\csch^2(\phi(\gamma_{12}^\star))+1-p} < \frac{Lp}{\csch^2(\phi(\gamma_{12}^\star))+1-p}.
\end{align*}
This completes the proof. \hfill ${\color{red}\blacksquare}$ \\

\noindent
\textbf{Proof of Theorem \ref{Thm:BS_bound}.} In this proof, we aim to establish that there exists $L_0\in\mathbb N$ such that for all $L\ge L_0$
$$
\mathbb{P}(B > 0) > \mathbb{P}(A > 0),
$$
under the assumption that $y_{12}^{(l)} \sim G(\phi, \psi_{\gamma_{12}^\star}, \gamma_{12}^\star, K)$ with $\gamma_{12}^\star > 0$, and that $X_{\gamma_{12}^\star} \sim \mathrm{Geo}(\psi_{\gamma_{12}^\star}, K)$ is non-degenerate. Proving this is equivalent to showing that there exists $L_0$ such that for all $L\ge L_0$
$$
\mathbb{P}(B \leq 0) < \mathbb{P}(A \leq 0)
\quad \Leftrightarrow \quad
\mathbb{P}(-B \geq 0) < \mathbb{P}(-A \geq 0).
$$

By the definitions of $A$ and $B$, the probabilities can be expressed as
\begin{align*}
    \mathbb{P}(-B \geq 0) = &\ \mathbb{P}\left(-\sum_{l=1}^L a_{12}^{(l)}\sign(y_{12}^{(l)}) \geq 0\right), \\
    \mathbb{P}(-A \geq 0) = &\ \mathbb{P}\left(-\sum_{l=1}^L a_{12}^{(l)}y_{12}^{(l)} \geq 0\right).
\end{align*}
Since $\gamma_{12}^\star > 0$, it follows that
\begin{align*}
\mathbb{E}[-a_{12}^{(l)}\sign(y_{12}^{(l)})] &= -p\cdot \tanh(\phi(\gamma_{12}^\star)) < 0, \\
\mathbb{E}[-a_{12}^{(l)}y_{12}^{(l)}] &= -p \cdot \tanh(\phi(\gamma_{12}^\star)) \cdot \frac{\sum_{k=1}^K k e^{\psi_{\gamma_{12}^\star}(k)}}{\sum_{k=1}^K e^{\psi_{\gamma_{12}^\star}(k)}} < 0.
\end{align*}
Applying Lemma~\ref{Lemma:RateFunction} and Theorem~\ref{Thm:Cramer}, we have
$$
\lim_{L \to \infty} \frac{1}{L} \log \mathbb{P}(-B \geq 0) = -I_1(0), \quad
\lim_{L \to \infty} \frac{1}{L} \log \mathbb{P}(-A \geq 0) = -I_2(0),
$$
where $I_1(0)$ and $I_2(0)$ denote the rate functions at zero for $-a_{12}^{(l)}\sign(y_{12}^{(l)})$ and $-a_{12}^{(l)}y_{12}^{(l)}$, respectively:
\begin{align*}
I_1(0) & = \log\left(\frac{\cosh(\phi(\gamma_{12}^\star))}{p+(1-p)\cosh(\phi(\gamma_{12}^\star))}\right)  \\
I_2(0) &= \log\left(\frac{\cosh(\phi(\gamma_{12}^\star))}{p+(1-p)\cosh(\phi(\gamma_{12}^\star))}\right) - Q(p,\gamma_{12}^\star),
\end{align*}
where $Q(p,\gamma_{12}^\star)$ is defined as
\begin{align*}
Q(p,\gamma_{12}^\star)=\inf_{\lambda \in \mathbb{R}} 
    \log \left(\frac{ \frac{p\sum_{k = 1}^K e^{\psi_{\gamma_{12}^\star}(k)} \cosh\left( \phi(\gamma_{12}^\star) + \lambda k \right)}{\sum_{k = 1}^K e^{\psi_{\gamma_{12}^\star}(k)}}+(1-p) \cosh\left( \phi(\gamma_{12}^\star) \right)}{p+(1-p)\cosh(\phi(\gamma_{12}^\star))}\right)
\end{align*}

Since $K \geq 2$ and $X_{\gamma_{12}^\star} \sim \mathrm{Geo}(\psi_{\gamma_{12}^\star}, K)$ is non-degenerate and $\cosh(x)$ is a convex function, we have the strict inequality
$$
 \frac{\sum_{k=1}^K e^{\psi_{\gamma_{12}^\star}(k)} \cosh(\phi(\gamma_{12}^\star) + \lambda k)}{\sum_{k=1}^K e^{\psi_{\gamma_{12}^\star}(k)}}  
>  \cosh\left(\phi(\gamma_{12}^\star)+\lambda \cdot \frac{\sum_{k=1}^K k \cdot e^{\psi_{\gamma_{12}^\star}(k)}}{\sum_{k=1}^K e^{\psi_{\gamma_{12}^\star}(k)}} \right) = \cosh \left(\phi(\gamma_{12}^\star)+\lambda \mathbb{E}(X_{\gamma_{12}^\star})\right),
$$
which implies that
$$
Q(p,\gamma_{12}^\star) > \inf_{\lambda \in \mathbb{R}} 
    \log \left(\frac{p\cdot \cosh \left(\phi(\gamma_{12}^\star)+\lambda \mathbb{E}(X_{\gamma_{12}^\star})\right)+(1-p) \cosh\left( \phi(\gamma_{12}^\star) \right)}{p+(1-p)\cosh(\phi(\gamma_{12}^\star))}\right)=0.
$$
Therefore, we have $I_1(0)>I_2(0)$. For $\varepsilon > 0$ and a positive integer $L_0$ depending on $\varepsilon$ such that for all $L \geq L_0$,
$$
\exp\left(-L(I_1(0)+\varepsilon)\right) \leq \mathbb{P}(-B \geq 0) \leq \exp\left(-L(I_1(0)-\varepsilon)\right),
$$
$$
\exp\left(-L(I_2(0)+\varepsilon)\right) \leq \mathbb{P}(-A \geq 0) \leq \exp\left(-L(I_2(0)-\varepsilon)\right).
$$
Choosing $\varepsilon$ such that $\varepsilon < \frac{I_1(0) - I_2(0)}{2}$, we further obtain
$$
\mathbb{P}(-A \geq 0) \geq \exp\left(-L(I_2(0)+\varepsilon)\right)
> \exp\left(-L(I_1(0)-\varepsilon)\right) \geq \mathbb{P}(-B \geq 0).
$$
Furthermore, we have
\begin{align*}
    \lim_{L\rightarrow \infty}\frac{ \mathbb{P}(-B \geq 0)}{ \mathbb{P}(-A \geq 0)} \leq 
     \lim_{L\rightarrow \infty} \frac{e^{-L(I_1(0)-\varepsilon)}}{e^{-L(I_2(0)+\varepsilon)}}
     = \lim_{L\rightarrow \infty}e^{-L(I_1(0)-I_2(0)-2\varepsilon)}=0.
\end{align*}
Note that the choice of $\varepsilon$ depends on $I_1(0) - I_2(0)$ and consequently influences the value of $L_0$. Hence, $L_0$ is determined by $I_1(0) - I_2(0)$, which equals $Q(p,\gamma_{12}^\star)$. Clearly, $Q(p,\gamma_{12}^\star)$ is a function of $p$, $\gamma_{12}^\star$, and the pattern function $\psi_{\gamma_{12}^\star}$. This completes the proof.\hfill ${\color{red}\blacksquare}$ \\

\noindent
\textbf{Proof of Theorem \ref{Thm:Consist}.} Fix a pair $(i,j)$ with $i\neq j$. Consider the sequence $\{X_{ij}^{(l)}\}_{l\ge1}$ defined by
\[
X_{ij}^{(l)} \triangleq  a_{ij}^{(l)} y_{ij}^{(l)}.
\]
By the assumption that the $X_{ij}^{(l)}$ are independent across $l$ with mean $\mu_{ij}\neq0$ and finite variance. Hence the strong law of large numbers (SLLN) yields
\[
\frac{1}{L}\sum_{l=1}^L X_{ij}^{(l)} \xrightarrow{\text{a.s.}} \mu_{ij}
\qquad (L\to\infty).
\]
Multiplying both sides by $L$ gives
\[
\sum_{l=1}^L X_{ij}^{(l)} \xrightarrow{\text{a.s.}} 
\begin{cases}
+\infty, & \text{if }\mu_{ij}>0,\\
-\infty, & \text{if }\mu_{ij}<0,
\end{cases}
\]
in the sense that for sufficiently large $L$ the sign of the finite sum equals $\operatorname{sign}(\mu_{ij})$ almost surely. Consequently, the indicator
\[
\mathbb{I}\Big\{\sum_{l=1}^L a_{ij}^{(l)}y_{ij}^{(l)}>0\Big\}
\]
converges almost surely to the constant $\mathbb{I}\{\mu_{ij}>0\}$ as $L\to\infty$.

The same argument applies to the binarized terms. Let
\[
\widetilde X_{ij}^{(l)} \triangleq a_{ij}^{(l)}\operatorname{sign}(y_{ij}^{(l)}),
\]
with mean $\widetilde\mu_{ij}\neq0$. Then by SLLN
\[
\frac{1}{L}\sum_{l=1}^L \widetilde X_{ij}^{(l)} \xrightarrow{\text{a.s.}} \widetilde\mu_{ij},
\]
and therefore
\[
\mathbb{I}\Big\{\sum_{l=1}^L a_{ij}^{(l)}\operatorname{sign}(y_{ij}^{(l)})>0\Big\}
\xrightarrow{\text{a.s.}} \mathbb{I}\{\widetilde\mu_{ij}>0\}.
\]

Now fix an item $i$. The win-counts are finite sums over $j\in[n]\setminus\{i\}$:
\[
S_i=\sum_{j\neq i}\mathbb{I}\Big\{\sum_{l=1}^L a_{ij}^{(l)}y_{ij}^{(l)}>0\Big\},\qquad
\widetilde S_i=\sum_{j\neq i}\mathbb{I}\Big\{\sum_{l=1}^L a_{ij}^{(l)}\operatorname{sign}(y_{ij}^{(l)})>0\Big\}.
\]
Since each summand converges almost surely to a constant, and the sum is finite (over $n-1$ terms), we may interchange limit and finite sum to obtain almost sure limits:
\begin{align*}
   &S_i \xrightarrow{\text{a.s.}}  \sum_{j\neq i}\mathbb{I}\{\mu_{ij}>0\} = |\{j \in [n]\setminus\{i\}:\theta_i^\star>\theta_j^\star\}|, \\
&
\widetilde S_i \xrightarrow{\text{a.s.}} \sum_{j\neq i}\mathbb{I}\{\widetilde\mu_{ij}>0\}=|\{j \in [n]\setminus\{i\}:\theta_i^\star>\theta_j^\star\}|, 
\end{align*}
where $|A|$ represents the cardinality of a set $A$, and the equality follows from $\sign(\mu_{ij})=\sign(\theta_i^\star-\theta_j^\star)$ (and similarly for $\widetilde\mu_{ij}$). Hence the rank of $S_i$ (resp.\ $\widetilde S_i$) converges almost surely to the rank of $\theta_i^\star$:
\[
\sigma(S_i)\xrightarrow{\text{a.s.}}\sigma(\theta_i^\star),
\qquad
\sigma(\widetilde S_i)\xrightarrow{\text{a.s.}}\sigma(\theta_i^\star).
\]
This completes the proof.  \hfill ${\color{red}\blacksquare}$ \\

\noindent
\textbf{Proof of Theorem \ref{ThmK_bound}.} For simplicity, and without loss of generality, we assume throughout the proof that $\theta_1^\star>\theta_2^\star>\cdots>\theta_n^\star$, indicating that $\sigma(\theta_i^\star)=i$ for $i \in [n]$. Therefore, for $i<j$, we have
\begin{align*}
    [\sigma(S_i)-\sigma(S_j)] \cdot 
    [\sigma(\theta_i^\star)-\sigma(\theta_j^\star)] \leq 0
    \Longleftrightarrow [\sigma(S_i)-\sigma(S_j)] \geq 0
    \Longleftrightarrow S_i \leq S_j.
\end{align*}
Further, $\tau(\bm{S},\bm{\theta}^\star)$ and $\tau(\widetilde{\bm{S}},\bm{\theta}^\star)$ can be expressed as
\begin{align*}
&\tau(\bm{S},\bm{\theta}^\star) = 
\frac{2}{n(n-1)}
\sum_{1 \leq i <j \leq n}\mathbb{I}\left(S_i \leq S_j\right), \\
   & \tau(\widetilde{\bm{S}},\bm{\theta}^\star) = 
    \frac{2}{n(n-1)}
\sum_{1 \leq i <j \leq n} \mathbb{I}\left(\widetilde{S}_i \leq \widetilde{S}_j\right).
\end{align*}
Next, we analyze $\mathbb{I}\left(S_i \leq S_j\right)$ and $\mathbb{I}\left(\widetilde{S}_i \leq \widetilde{S}_j\right)$ for any $i<j$ with $\theta_i^\star>\theta_j^\star$. Here, $i$ ranges over $\{1, \ldots, n-1\}$, and for each $i$, $j$ ranges over $\{i+1, \ldots, n\}$. Note that the probability $ \mathbb{P}\big(\sum_{l=1}^L a_{ij}^{(l)} y_{ij}^{(l)} = 0\big) $ is typically negligible compared with $ \mathbb{P}\big(\sum_{l=1}^L a_{ij}^{(l)} y_{ij}^{(l)} > 0\big) $. Therefore, in the following proof, we consider the case conditional on $ \sum_{l=1}^L a_{ij}^{(l)} y_{ij}^{(l)} \neq 0 $.

\noindent
\textbf{Step 1. Decomposition of $\mathbb{I}\left(S_i \leq S_j\right)$ and $\mathbb{I}\left(\widetilde{S}_i \leq \widetilde{S}_j\right)$.} First, for notational convenience, we denote for every $i<j$ that
\begin{align*}
    Z_{ij} =\mathbb{I}\left[\sum_{l=1}^L a_{ij}^{(l)} y_{ij}^{(l)}>0\right] \in \{0,1\} \, \mbox{ and } \,
    \widetilde{Z}_{ij}= \mathbb{I}\left[\sum_{l=1}^L a_{ij}^{(l)} \sign(y_{ij}^{(l)})>0\right] \in \{0,1\}.
\end{align*}
Conditional on $a_{ij}^{(l)} y_{ij}^{(l)} \neq 0$, we have the relation that $Z_{ij}=1-Z_{ji}$. Therefore,
\begin{align*}
    S_i - S_j &= \sum_{k \in [n]\setminus \{i\}} Z_{ik}-
    \sum_{k \in [n]\setminus \{j\}} Z_{jk} \\
    &= 2Z_{ij}-1+\sum_{k \in [n]\setminus \{i,j\}} Z_{ik}-\sum_{k \in [n]\setminus \{i,j\}} Z_{jk} \\
    &\triangleq 2Z_{ij}-1 + S_{i,-j}-S_{j,-i}.
\end{align*}
Here, $S_{i,-j}$ and $S_{j,-i}$ denote the number of wins achieved by items $i$ and $j$, respectively, against all items other than $i$ and $j$. Furthermore, we have
\begin{align*}
    \mathbb{I}(S_i \leq S_j) = & \mathbb{I}(2Z_{ij}-1+ S_{i,-j} \leq S_{j,-i})  \\
    = & \mathbb{I}(1 + S_{i,-j} \leq S_{j,-i})
    \cdot \mathbb{I}(Z_{ij}=1)+\mathbb{I}(-1+S_{i,-j} \leq S_{j,-i})
    \cdot \mathbb{I}(Z_{ij}=0).
\end{align*}
Next, we further decompose $\mathbb{I}(1 + S_{i,-j} \leq S_{j,-i})$ and $\mathbb{I}(-1+S_{i,-j} \leq S_{j,-i})$ as follows
\begin{align*}
   \mathbb{I}(1 + S_{i,-j} \leq S_{j,-i}) &= \sum_{a=0}^{n-3}
   \sum_{b=a+1}^{n-2} \mathbb{I}(S_{i,-j}=a)\cdot\mathbb{I}(S_{j,-i}=b), \\
   \mathbb{I}(-1 + S_{i,-j} \leq S_{j,-i}) &= 
   \mathbb{I}(S_{i,-j}=0)+
   \sum_{a=1}^{n-2}
   \sum_{b=a-1}^{n-2} \mathbb{I}(S_{i,-j}=a)\cdot\mathbb{I}(S_{j,-i}=b).
\end{align*}
Further, for any $a,b \in [n-2]$, we consider the following decompositions:
\begin{align*}
\mathbb{I}(S_{i,-j} = a) 
&= \sum_{\substack{A \subseteq \{1,\dots,n\}\setminus\{i,j\} \\ |A| = a}} \underbrace{\left(
\prod_{m \in A} Z_{im} \prod_{m \notin A} (1 - Z_{im})\right)}_{\triangleq P_{i,-j}(A,a)}, \\
\mathbb{I}(S_{j,-i} = b) 
&= \sum_{\substack{B \subseteq \{1,\dots,n\}\setminus\{i,j\} \\ |B| = b}} \underbrace{\left(
\prod_{m \in B} Z_{jm} \prod_{m \notin B} (1 - Z_{jm})\right)}_{\triangleq P_{j,-i}(B,b)}.
\end{align*}
Here, $m \in [n]\setminus \{i,j\}$, $A$ denotes the set of items (excluding $j$) that lose to item $i$, and $B$ denotes the set of items (excluding $i$) that lose to item $j$.

To sum up, $\mathbb{I}(S_i \leq S_j)$ can be written as
\begin{align*}
    &\mathbb{I}(S_i \leq S_j) = 
    \mathbb{I}(Z_{ij}=1)\left\{ \sum_{a=0}^{n-3}
   \sum_{b=a+1}^{n-2} \left[\sum_{\substack{A \subseteq \{1,\dots,n\}\setminus\{i,j\} \\ |A| = a}} P_{i,-j}(A,a) \cdot\sum_{\substack{B \subseteq \{1,\dots,n\}\setminus\{i,j\} \\ |B| = b}}\cdot P_{j,-i}(B,b)\right] \right\}\\
+&\mathbb{I}(Z_{ij}=0) \left\{\prod_{m \in [n]\setminus\{i,j\}}(1-Z_{im})
+\sum_{a=1}^{n-2}
   \sum_{b=a-1}^{n-2}\left[ \sum_{\substack{A \subseteq \{1,\dots,n\}\setminus\{i,j\} \\ |A| = a}}P_{i,-j}(A,a) \cdot \sum_{\substack{B \subseteq \{1,\dots,n\}\setminus\{i,j\} \\ |B| = b}} P_{j,-i}(B,b)\right]\right\}.
\end{align*}
Similarly, using the same treatment, we have
\begin{align*}
     & \mathbb{I}(\widetilde{S}_i \leq \widetilde{S}_j) = 
    \mathbb{I}(\widetilde{Z}_{ij}=1) \left\{\sum_{a=0}^{n-3}
   \sum_{b=a+1}^{n-2} \left[\sum_{\substack{A \subseteq \{1,\dots,n\}\setminus\{i,j\} \\ |A| = a}} \widetilde{P}_{i,-j}(A,a) \cdot \sum_{\substack{B \subseteq \{1,\dots,n\}\setminus\{i,j\} \\ |B| = b}} \widetilde{P}_{j,-i}(B,b)\right] \right\}\\
+&\mathbb{I}(\widetilde{Z}_{ij}=0)\left\{\prod_{m \in [n]\setminus\{i,j\}}(1-\widetilde{Z}_{im})+ \sum_{a=1}^{n-2}
   \sum_{b=a-1}^{n-2}\left[\sum_{\substack{A \subseteq \{1,\dots,n\}\setminus\{i,j\} \\ |A| = a}} \widetilde{P}_{i,-j}(A,a) \cdot \sum_{\substack{B \subseteq \{1,\dots,n\}\setminus\{i,j\} \\ |B| = b}} \widetilde{P}_{j,-i}(B,b)\right]\right\},  
\end{align*}
where $\widetilde{P}_{i,-j}(A,a)$ and $\widetilde{P}_{j,-i}(B,b)$ are defined as
\begin{align*}
\widetilde{P}_{i,-j}(A,a) =& 
\prod_{m \in A} \widetilde{Z}_{im} \prod_{m \notin A} (1 - \widetilde{Z}_{im}), \\
    \widetilde{P}_{j,-i}(B,b) = &\prod_{m \in B} \widetilde{Z}_{jm} \prod_{m \notin B} (1 - \widetilde{Z}_{jm}),
\end{align*}
where $A,B \subset [n]\setminus \{i,j\}$ with $|A|=a$ and $|B|=b$. Here, we denote that
\begin{align*}
    m \notin A \Leftrightarrow m \in \{1,2,\ldots,n\} \setminus \left(\{i,j\} \cup A\right), \\
    m \notin B \Leftrightarrow m \in \{1,2,\ldots,n\} \setminus \left(\{i,j\} \cup B\right).
\end{align*}

Since $\{Z_{ij} : i<j\}$ consists of independent random variables, and likewise $\{\widetilde{Z}_{ij} : i<j\}$ are independent, we have
\begin{align*}
    &\mathbb{E}
    \left(\mathbb{I}(S_i \leq S_j)\right)=\mathbb{P}(S_i \leq S_j)  \\
   = &\ 
   \mathbb{P}(Z_{ij}=1) \sum_{a=0}^{n-3}
   \sum_{b=a+1}^{n-2} \left\{ \sum_{\substack{A \subseteq \{1,\dots,n\}\setminus\{i,j\} \\ |A| = a}} \mathbb{E}[P_{i,-j}(A,a)] \cdot \sum_{\substack{B \subseteq \{1,\dots,n\}\setminus\{i,j\} \\ |B| = b}} \mathbb{E}[P_{j,-i}(B,b)]\right\} \\
   &\ + \mathbb{P}(Z_{ij}=0) \sum_{a=1}^{n-2}
   \sum_{b=a-1}^{n-2} \left\{\sum_{\substack{A \subseteq \{1,\dots,n\}\setminus\{i,j\} \\ |A| = a}} \mathbb{E}[P_{i,-j}(A,a)] \cdot \sum_{\substack{B \subseteq \{1,\dots,n\}\setminus\{i,j\} \\ |B| = b}} \mathbb{E}[P_{j,-i}(B,b)]\right\}\\
   &\ +\prod_{m \in [n]\setminus \{i\}} \mathbb{P}(Z_{im} =0).
\end{align*}
Similarly, we have
\begin{align*}
       &\mathbb{E}
    \left(\mathbb{I}(\widetilde{S}_i \leq \widetilde{S}_j)\right)=\mathbb{P}(\widetilde{S}_i \leq \widetilde{S}_j)  \\
    =&    \mathbb{P}(\widetilde{Z}_{ij}=1) \sum_{a=0}^{n-3}
   \sum_{b=a+1}^{n-2} \left\{\sum_{\substack{A \subseteq \{1,\dots,n\}\setminus\{i,j\} \\ |A| = a}} \mathbb{E}[\widetilde{P}_{i,-j}(A,a)] \cdot \sum_{\substack{B \subseteq \{1,\dots,n\}\setminus\{i,j\} \\ |B| = b}}\mathbb{E}[\widetilde{P}_{j,-i}(B,b)]\right\} \\
   +&\mathbb{P}(\widetilde{Z}_{ij}=0) \sum_{a=1}^{n-2}
   \sum_{b=a-1}^{n-2}\left\{\sum_{\substack{A \subseteq \{1,\dots,n\}\setminus\{i,j\} \\ |A| = a}} \mathbb{E}[\widetilde{P}_{i,-j}(A,a)] \cdot \sum_{\substack{B \subseteq \{1,\dots,n\}\setminus\{i,j\} \\ |B| = b}} \mathbb{E}[\widetilde{P}_{j,-i}(B,b)]\right\} \\
   +&\prod_{m \in [n]\setminus \{i\}} \mathbb{P}(\widetilde{Z}_{im} =0).
\end{align*}

\noindent
\textbf{Step 2. Convergence of $\mathbb{E}
    \left(\mathbb{I}(S_i \leq S_j)\right)$ and $\mathbb{E}
    \left(\mathbb{I}(\widetilde{S}_i \leq \widetilde{S}_j)\right)$ to Zero.} In this part, we show that given that $\theta_i^\star>\theta_j^\star$, both $\mathbb{E}
    \left(\mathbb{I}(S_i \leq S_j)\right)$ and $\mathbb{E}
    \left(\mathbb{I}(\widetilde{S}_i \leq \widetilde{S}_j)\right)$ converge to zero. We mainly focus on $\mathbb{E}
    \left(\mathbb{I}(S_i \leq S_j)\right)$, and then the result for $\mathbb{E}
    \left(\mathbb{I}(\widetilde{S}_i \leq \widetilde{S}_j)\right)$ can be similarly derived.

First, noting that $\theta_i^\star>\theta_j^\star$, we obtain $\mathbb{E}(a_{ij}^{(l)} y_{ij}^{(l)})>0$. Then, by Lemma \ref{Lemma:SumVarRate}, we have
\begin{align*}
    \mathbb{P}(Z_{ij}=0) = \mathbb{P}\left(\sum_{l=1}^L a_{ij}^{(l)} y_{ij}^{(l)} \leq 0\right) = \mathbb{P}\left(-\sum_{l=1}^L a_{ij}^{(l)} y_{ij}^{(l)} \geq 0\right) \xrightarrow{L\to \infty} 0.
\end{align*}
Therefore, we have 
$$
\prod_{m \in [n]\setminus \{i\}} \mathbb{P}(Z_{im} =0) \xrightarrow{L\to \infty}0
$$
and 
\begin{align*}
    \mathbb{P}(Z_{ij}=0) \underbrace{\sum_{a=1}^{n-2}
   \sum_{b=a-1}^{n-2} \left\{ \sum_{\substack{A \subseteq \{1,\dots,n\}\setminus\{i,j\} \\ |A| = a}} \mathbb{E}[P_{i,-j}(A,a)] \cdot \sum_{\substack{B \subseteq \{1,\dots,n\}\setminus\{i,j\} \\ |B| = b}} \mathbb{E}[P_{j,-i}(B,b)]\right\}}_{<\infty} \xrightarrow{L\to \infty} 0.
\end{align*}
Next, we turn to show that given $Z_{ij}=1$
\begin{align}
\label{Temp_Con}
    \sum_{a=0}^{n-3}
   \sum_{b=a+1}^{n-2}\left\{ \sum_{\substack{A \subseteq \{1,\dots,n\}\setminus\{i,j\} \\ |A| = a}} \mathbb{E}[P_{i,-j}(A,a)] \cdot \sum_{\substack{B \subseteq \{1,\dots,n\}\setminus\{i,j\} \\ |B| = b}} \mathbb{E}[P_{j,-i}(B,b)]\right\}\xrightarrow{L\rightarrow \infty}0.
\end{align}

To prove (\ref{Temp_Con}), we define 
$$
A^\star = \{i+1, \ldots, n\} \setminus \{j\},
$$ 
which represents the set of all indices corresponding to items less preferred than item $i$ except item $j$. Clearly, $|A^\star| = n - i - 1$. If $i = 1$, then item $i$ is the most preferred item. In this case, $S_i \leq S_j$ implies that there exists some $m \neq i,j$ such that $Z_{im} = 0$. Hence, condition~(\ref{Temp_Con}) holds immediately from Lemma \ref{Lemma:SumVarRate}. For $2 \leq i \leq n-1$, we consider the following cases:
\begin{align*}
      &  \sum_{a=0}^{n-3}
   \sum_{b=a+1}^{n-2}\left\{ \sum_{\substack{A \subseteq \{1,\dots,n\}\setminus\{i,j\} \\ |A| = a}} \mathbb{E}[P_{i,-j}(A,a)] \cdot \sum_{\substack{B \subseteq \{1,\dots,n\}\setminus\{i,j\} \\ |B| = b}} \mathbb{E}[P_{j,-i}(B,b)]\right\} \\
  \textbf{Case 1:}\quad =& 
   \sum_{b=n-i}^{n-2}\left\{  \mathbb{E}[P_{i,-j}(A^\star,n-i-1)] \cdot \sum_{\substack{B \subseteq \{1,\dots,n\}\setminus\{i,j\} \\ |B| = b}} \mathbb{E}[P_{j,-i}(B,b)]\right\} \\
     \textbf{Case 2:}\quad   +&
   \sum_{b=n-i}^{n-2}\left\{  \sum_{\substack{A \subseteq \{1,\dots,n\}\setminus\{i,j\} \\ |A| = n-i-1,A\neq A^\star}} \mathbb{E}[P_{i,-j}(A,n-i-1)] \cdot \sum_{\substack{B \subseteq \{1,\dots,n\}\setminus\{i,j\} \\ |B| = b}} \mathbb{E}[P_{j,-i}(B,b)]\right\} \\
        \textbf{Case 3:}\quad   +&\sum_{ \substack{0 \leq a \leq n-3\\
        a \neq n-i-1}}
   \sum_{b=a+1}^{n-2}\left\{  \sum_{\substack{A \subseteq \{1,\dots,n\}\setminus\{i,j\} \\ |A| = a}} \mathbb{E}[P_{i,-j}(A,a)] \cdot \sum_{\substack{B \subseteq \{1,\dots,n\}\setminus\{i,j\} \\ |B| = b}} \mathbb{E}[P_{j,-i}(B,b)]\right\},
\end{align*}

\textbf{Case 1:} Since every $ m \in A^\star $, we have $ \sigma(\theta_i^\star) < \sigma(\theta_m^\star) $ that is $\theta_i^\star>\theta_m^\star$. Then, we have
\begin{align*}
    \mathbb{E}[P_{i,-j}(A^\star,n-i-1)] = &
    \prod_{m \in A^\star} \mathbb{E}(Z_{im}) \prod_{m \notin A^\star} (1 - \mathbb{E}(Z_{im})) \\
  = &
    \prod_{m \in A^\star} \mathbb{P}\left(\sum_{l=1}^L a_{im}^{(l)} y_{im}^{(l)}>0\right) \cdot \prod_{m \notin A^\star} \left[1 - \mathbb{P}\left(\sum_{l=1}^L a_{im}^{(l)} y_{im}^{(l)}>0\right)\right]
    \xrightarrow{L\rightarrow \infty} 1,
\end{align*}
where $\mathbb{E}(y_{im}^{(l)})>0$ for any $m \in A^\star$, and $\mathbb{E}(y_{im}^{(l)})<0$ for any $m \notin A^\star$.

Since $b \geq a+1=n-i$ and $\sigma(\theta_i^\star)<\sigma(\theta_j^\star)$, for any $B$ with $|B|=b$, there exists $m_0 \in B$ such that $\mathbb{E}(a_{jm_0}^{(l)}y_{jm_0}^{(l)})<0$ with $m_0<j$. Therefore, follows from Lemma \ref{Lemma:SumVarRate}, we can find $m_0 \in B$ such that
\begin{align*}
    \mathbb{P}(Z_{jm_0}) = \mathbb{P}\left(\sum_{l=1}^L a_{jm_0}^{(l)}y_{jm_0}^{(l)}>0\right)\xrightarrow{L\rightarrow \infty} 0.
\end{align*}
This then indicates that for any $B$ with $b \geq a+1$
\begin{align*}
    \mathbb{E}[P_{j,-i}(B,b)] = &\ \prod_{m \in B} \mathbb{P}\left(\sum_{l=1}^L a_{jm}^{(l)} y_{jm}^{(l)}>0\right) \cdot \prod_{m \notin B} \left[1 - \mathbb{P}\left(\sum_{l=1}^L a_{jm}^{(l)} y_{jm}^{(l)}>0\right)\right] \\
    \leq & \ \mathbb{P}\left(\sum_{l=1}^L a_{jm_0}^{(l)}y_{jm_0}^{(l)}>0\right)
    \xrightarrow{L\rightarrow \infty} 0.
\end{align*}
To sum up, for \textbf{Case 1}, we have
\begin{align}
    \label{Case1}
    \mathbb{P}(Z_{ij}=1) 
   \sum_{b=n-i}^{n-2}\left\{  \mathbb{E}[P_{i,-j}(A^\star,n-i-1)] \cdot \sum_{\substack{B \subseteq \{1,\dots,n\}\setminus\{i,j\} \\ |B| = b}} \mathbb{E}[P_{j,-i}(B,b)]\right\} \xrightarrow{L\rightarrow \infty} 0.
\end{align}

\textbf{Case 2 and Case 3:} When $A \neq A^\star$ with $|A| = n-i-1$, or when $|A| \neq n-i-1$, there must exist either some $m_1 \in A$ such that $\mathbb{E}(Z_{i m_1}) = 0$ while $\theta_i^\star < \theta_{m_1}^\star$, or some $m_1 \notin A$ such that $\mathbb{E}(Z_{i m_1}) = 1$ while $\theta_i^\star > \theta_{m_1}^\star$. Therefore, we have
\begin{align}
\label{Case2}
    \sum_{b=n-i}^{n-2}\left\{  \sum_{\substack{A \subseteq \{1,\dots,n\}\setminus\{i,j\} \\ |A| = n-i-1,A\neq A^\star}} \mathbb{E}[P_{i,-j}(A,n-i-1)] \cdot \sum_{\substack{B \subseteq \{1,\dots,n\}\setminus\{i,j\} \\ |B| = b}} \mathbb{E}[P_{j,-i}(B,b)]\right\}\xrightarrow{L\rightarrow \infty} 0, \\
    \label{Case3}
    \sum_{ \substack{0 \leq a \leq n-3\\
        a \neq n-i-1}}
   \sum_{b=a+1}^{n-2}\left\{  \sum_{\substack{A \subseteq \{1,\dots,n\}\setminus\{i,j\} \\ |A| = a}} \mathbb{E}[P_{i,-j}(A,a)] \cdot \sum_{\substack{B \subseteq \{1,\dots,n\}\setminus\{i,j\} \\ |B| = b}} \mathbb{E}[P_{j,-i}(B,b)]\right\}\xrightarrow{L\rightarrow \infty} 0.
\end{align}
since the sum only contains finite terms.

Combining (\ref{Case1})-(\ref{Case3}) yields the result in (\ref{Temp_Con}). This then implies that 
\begin{align*}
    \mathbb{P}(S_i \leq S_j) \xrightarrow{L\rightarrow \infty} 0\, \text{ for any }\, i<j.
\end{align*}
Since $\mathbb{P}(S_i \leq S_j)$ and $\mathbb{P}(\widetilde{S}_i \leq \widetilde{S}_j)$ are similar in nature. Using the same treatment, we also have
\begin{align*}
     \mathbb{P}(\widetilde{S}_i \leq \widetilde{S}_j) \xrightarrow{L\rightarrow \infty} 0\, \text{ for any }\, i<j.
\end{align*}

\noindent
\textbf{Step 3. Comparing $\mathbb{P}(\widetilde{S}_i \leq \widetilde{S}_j)$ and $\mathbb{P}(S_i \leq S_j)$.} In this step, we intend to show that for any $i<j$ with $\theta_i^\star>\theta_j^\star$, we have
\begin{align*}
    \frac{\mathbb{P}(\widetilde{S}_i \leq \widetilde{S}_j)}{\mathbb{P}(S_i \leq S_j)} \xrightarrow{L\rightarrow \infty} 0.
\end{align*}
Note that $\mathbb{P}(S_i \leq S_j)$ can be written as 
\begin{align*}
  &\ \mathbb{P}(S_i \leq S_j)\\
  = &\ \mathbb{P}(Z_{ij}=1) \sum_{\substack{A,B \subseteq \{1,\dots,n\}\setminus\{i,j\} \\ |A|\leq |B|-1}} \left\{\mathbb{E}[P_{i,-j}(A,a)] \cdot\mathbb{E}[P_{j,-i}(B,b)]\right\} \\
  + &\ 
  \mathbb{P}(Z_{ij}=0) \sum_{\substack{A,B \subseteq \{1,\dots,n\}\setminus\{i,j\} \\ |A|\leq |B|+1}} \left\{\mathbb{E}[P_{i,-j}(A,a)] \cdot\mathbb{E}[P_{j,-i}(B,b)]\right\} 
  \\
  = &\ \mathbb{P}(Z_{ij}=1)\sum_{\substack{A,B \subseteq \{1,\dots,n\}\setminus\{i,j\} \\ |A|\leq |B|-1}}  \prod_{m \in A} \mathbb{P}(Z_{im}=1) \prod_{m \notin A} (1 - \mathbb{P}(Z_{im}=1)) \prod_{m \in B} \mathbb{P}(Z_{jm}=1) \prod_{m \notin B} (1 - \mathbb{P}(Z_{jm}=1)) \\
  +&\ \mathbb{P}(Z_{ij}=0)\sum_{\substack{A,B \subseteq \{1,\dots,n\}\setminus\{i,j\} \\ |A|\leq |B|+1}}  \prod_{m \in A} \mathbb{P}(Z_{im}=1) \prod_{m \notin A} (1 - \mathbb{P}(Z_{im}=1)) \prod_{m \in B} \mathbb{P}(Z_{jm}=1) \prod_{m \notin B} (1 - \mathbb{P}(Z_{jm}=1))\\
  \triangleq &\ \mathbb{P}(Z_{ij}=1) \sum_{\substack{A,B \subseteq \{1,\dots,n\}\setminus\{i,j\} \\ |A|\leq |B|-1}} Q_{ij}(A,B)
  +\mathbb{P}(Z_{ij}=0) \sum_{\substack{A,B \subseteq \{1,\dots,n\}\setminus\{i,j\} \\ |A|\leq |B|-1}} Q_{ij}(A,B),
\end{align*}
where $Q_{ij}(A,B)$ is defined as
\begin{align*}
    Q_{ij}(A,B) = \prod_{m \in A} \mathbb{P}(Z_{im}=1) \prod_{m \notin A} (1 - \mathbb{P}(Z_{im}=1)) \prod_{m \in B} \mathbb{P}(Z_{jm}=1) \prod_{m \notin B} (1 - \mathbb{P}(Z_{jm}=1)).
\end{align*}
Similarly, we have
\begin{align*}
&\ \mathbb{P}(\widetilde{S}_i \leq \widetilde{S}_j) \\
= &\ \mathbb{P}(\widetilde{Z}_{ij}=1)\sum_{\substack{A,B \subseteq \{1,\dots,n\}\setminus\{i,j\} \\ |A|\leq |B|-1}}  \prod_{m \in A} \mathbb{P}(\widetilde{Z}_{im}=1) \prod_{m \notin A} (1 - \mathbb{P}(\widetilde{Z}_{im}=1)) \prod_{m \in B} \mathbb{P}(\widetilde{Z}_{jm}=1) \prod_{m \notin B} (1 - \mathbb{P}(\widetilde{Z}_{jm}=1)) \\
  +&\ \mathbb{P}(\widetilde{Z}_{ij}=0)\sum_{\substack{A,B \subseteq \{1,\dots,n\}\setminus\{i,j\} \\ |A|\leq |B|+1}}  \prod_{m \in A} \mathbb{P}(\widetilde{Z}_{im}=1) \prod_{m \notin A} (1 - \mathbb{P}(\widetilde{Z}_{im}=1)) \prod_{m \in B} \mathbb{P}(\widetilde{Z}_{jm}=1) \prod_{m \notin B} (1 - \mathbb{P}(\widetilde{Z}_{jm}=1)) \\
  \triangleq &\ \mathbb{P}(\widetilde{Z}_{ij}=1)\sum_{\substack{A,B \subseteq \{1,\dots,n\}\setminus\{i,j\} \\ |A|\leq |B|-1}} \widetilde{Q}_{ij}(A,B)+\mathbb{P}(\widetilde{Z}_{ij}=0)\sum_{\substack{A,B \subseteq \{1,\dots,n\}\setminus\{i,j\} \\ |A|\leq |B|+1}}\widetilde{Q}_{ij}(A,B).
\end{align*}
% As established in \textbf{Step 2}, both $\mathbb{P}(\widetilde{S}_i \leq \widetilde{S}_j)$ and $\mathbb{P}(S_i \leq S_j)$ converge to zero, implying that each term in the above summation also converges to zero. In other words, we have
% \begin{align*}
%  \text{For any $A,B$ with }|A|\leq |B|-1, \quad    \mathbb{P}(\widetilde{Z}_{ij}=1)\widetilde{Q}_{ij}(A,B)\xrightarrow{L\to \infty} 0, \\
%    \text{For any $A,B$ with }|A|\leq |B|+1, \quad  \mathbb{P}(\widetilde{Z}_{ij}=0) \widetilde{Q}_{ij}(A,B)\xrightarrow{L\to \infty} 0, \\
%    \text{For any $A,B$ with }|A|\leq |B|-1, \quad  \mathbb{P}(Z_{ij}=1)  Q_{ij}(A,B)\xrightarrow{L\to \infty} 0,\\
%   \text{For any $A,B$ with }|A|\leq |B|+1, \quad \mathbb{P}(Z_{ij}=0)  Q_{ij}(A,B)\xrightarrow{L\to \infty} 0.
% \end{align*}
By Lemma \ref{Lemma:SumVarRate}, for any $m\ne i$ we have
\begin{align*}
    \min\left\{\frac{\mathbb{P}(\widetilde{Z}_{im}=1)}{\mathbb{P}(Z_{im}=1)}, \frac{1-\mathbb{P}(\widetilde{Z}_{im}=1)}{1-\mathbb{P}(Z_{im}=1)}\right\}\xrightarrow{L\rightarrow\infty}0 ,
\end{align*}
and 
\begin{align*}
\max\left\{\frac{\mathbb{P}(\widetilde{Z}_{im}=1)}{\mathbb{P}(Z_{im}=1)}, \frac{1-\mathbb{P}(\widetilde{Z}_{im}=1)}{1-\mathbb{P}(Z_{im}=1)}\right\}\xrightarrow{L\rightarrow\infty} 1.
\end{align*}
Therefore, we have
\begin{align*}
 \text{If $A,B$ with }|A|\leq |B|-1, \quad 
    \frac{ \mathbb{P}(\widetilde{Z}_{ij}=1)\widetilde{Q}_{ij}(A,B)}{\mathbb{P}(Z_{ij}=1)  Q_{ij}(A,B)}\xrightarrow{L\to \infty} 0,\\
   \text{If $A,B$ with }|A|\leq |B|+1, \quad       \frac{ \mathbb{P}(\widetilde{Z}_{ij}=0)\widetilde{Q}_{ij}(A,B)}{\mathbb{P}(Z_{ij}=0)  Q_{ij}(A,B)}\xrightarrow{L\to \infty} 0.
\end{align*}
Finally, using Lemma \ref{Lemma:Convergence}, we have
\begin{align*}
    \frac{\mathbb{P}(\widetilde{S}_i \leq \widetilde{S}_j)}{\mathbb{P}(S_i \leq S_j)} \xrightarrow{L\to \infty} 0.
\end{align*}
This then implies that 
\begin{align*}
      \lim_{L\rightarrow \infty} \frac{\mathbb{E}\big[\tau(\widetilde{\bm{S}},\bm{\theta}^\star)\big]}{\mathbb{E}\big[\tau(\bm{S},\bm{\theta}^\star)\big]}=
      \frac{\sum_{1 \leq i <j \leq n}\mathbb{P}\left(\widetilde{S}_i \leq \widetilde{S}_j\right)}{\sum_{1 \leq i <j \leq n}\mathbb{P}\left(S_i \leq S_j\right)}\xrightarrow{L\to \infty} 0,
\end{align*}
where the convergence implied by Lemma \ref{Lemma:Convergence} again. Therefore, there exists a positive integer $L_1$ such that
\begin{align*}
\mathbb{E}\big[\tau(\widetilde{\bm{S}},\bm{\theta}^\star)\big]
<\mathbb{E}\big[\tau(\bm{S},\bm{\theta}^\star)\big] \text{ for all } L \geq L_1.
\end{align*}
This completes the proof. \hfill ${\color{red}\blacksquare}$\\

\color{black}

\noindent
\textbf{Proof of Theorem \ref{Thm:MinimalSNR}.} For the function $\psi_{\gamma}$, we denote the corresponding probability mass vector by
\begin{align*}
\bm{p}(\psi_{\gamma}) = (p_1(\psi_{\gamma}), \ldots, p_K(\psi_{\gamma})) =
\left(
\frac{e^{\psi_{\gamma}(1)}}{\sum_{k=1}^K e^{\psi_{\gamma}(k)}},
\ldots,
\frac{e^{\psi_{\gamma}(K)}}{\sum_{k=1}^K e^{\psi_{\gamma}(k)}}
\right),
\end{align*}
where each $p_k(\psi)$ represents the probability mass assigned to category $k$ under $\psi$.

Note that $X_{\gamma}$ takes values in the finite set $\{1, \ldots, K\}$ for any choice of $\psi_{\gamma}$, and hence its expectation $\mathbb{E}[X_{\gamma}]$ lies in the interval $[1, K]$.

\noindent
\textbf{Step 1. Maximum Variance with Mean Constraint.} For a fixed value $s \in [1, K]$, we define the associated class of probability mass vectors $\bm{p} = (p_1, \ldots, p_K)$ as
\[
\mathcal{H}(s) = \left\{ \bm{p} : \sum_{k=1}^K p_k = 1 \text{ and } \sum_{k=1}^K k \cdot p_k = s \right\}.
\]
It is important to note that vectors in $\mathcal{H}(s)$ are general probability mass functions, without being restricted to those induced by $\psi$. In contrast, $\bm{p}(\psi)$ refers to the subclass of distributions that satisfy the structural constraint imposed by the softmax transformation of $\psi$.

We first consider the problem of determining the maximum variance of a \textbf{general} random variable $X_0$ whose distribution satisfies $\mathbb{P}(X_0 = k) = p_k$ for $k \in [K]$, where the probability vector $\bm{p}=(p_1,\ldots,p_K)$ belongs to the set $\mathcal{H}(s)$. We aim to maximize the variance of $X_0$ supported on $\{1, 2, \dots, K\}$, under the constraint that its mean is fixed at $s \in [1, K]$. The variance of $X_0$ is given by
$$
\operatorname{Var}(X_0) = \mathbb{E}[X_0^2] - s^2 = \sum_{k=1}^K k^2 p_k - s^2.
$$
We thus aim to maximize the second moment $\sum_{k=1}^K k^2 p_k$ under the constraints
$$
\sum_{k=1}^K p_k = 1, \quad \sum_{k=1}^K k p_k = s, \quad p_k \geq 0.
$$
This is a linear programming problem, hence the optimum is attained at an extreme point. We introduce Lagrange multipliers $\lambda_1$ and $\lambda_2$, and define the Lagrangian as
$$
\mathcal{L}(\bm{p}, \lambda_1, \lambda_2) = \sum_{k=1}^K k^2 p_k - \lambda_1 \left( \sum_{k=1}^K k p_k - s \right) - \lambda_2 \left( \sum_{k=1}^K p_k - 1 \right).
$$
Taking the partial derivative with respect to $p_k$ yields that
$$
\frac{\partial \mathcal{L}}{\partial p_k} = k^2 - \lambda_1 k - \lambda_2 = 0, \text{ for all }k \in [K]
$$
This quadratic equation implies that $p_k > 0$ only if $k$ satisfies
$$
k^2 - \lambda_1 k - \lambda_2 = 0.
$$
Since this equation has at most two real roots, the optimal distribution must be supported on at most two points. Hence, the maximum is attained by a \textbf{two-point distribution}. This is due to the complementary slackness that the equality $k^2-\lambda_1 k-\lambda_2=0$ holds for all $k$ with $p_k>0$. As this is a quadratic equation in $k$, the support size of the optimal solution is at most two. Therefore, for any $s>0$, the maximum variance $\sum_{k=1}^K k^2 p_k$ is achieved at a two-point distribution.

\noindent
\textbf{Step 2. Closed Form of Variance for Two-point Distribution.} Let the two support points be $a < b \in \{1, 2, \dots, K\}$, and suppose $\mathbb{P}(X_0 = a) = p$, $\mathbb{P}(X_0 = b) = 1 - p$. The mean constraint gives:
$$
p a + (1 - p) b = s \quad \Rightarrow \quad p = \frac{b - s}{b - a}.
$$
Then,
$$
\mathbb{E}[X_0^2] = p a^2 + (1 - p) b^2 = \frac{b - s}{b - a} a^2 + \frac{s - a}{b - a} b^2.
$$
This expression is maximized when $a = 1$ and $b = K$, giving the maximal second moment:
$$
\mathbb{E}[X_0^2] = \frac{K - s}{K - 1} \cdot 1^2 + \frac{s - 1}{K - 1} \cdot K^2.
$$
Hence, the maximum variance under the mean constraint is:
\begin{align}
    \label{Var_Formula}
    \operatorname{Var}(X_0) = \left( \frac{K - s}{K - 1} \cdot 1 + \frac{s - 1}{K - 1} \cdot K^2 \right) - s^2.
\end{align}

\noindent
\textbf{Step 3. Minimum SNR.} From Step 1 and Step 2, we conclude that \textbf{the maximum variance given a fixed mean is achieved by a two-point distribution supported on} $\{1, K\}$. Therefore, the \textbf{minimum signal-to-noise ratio (SNR)} is also attained by a distribution supported on $\{1, K\}$. Let $K \ge 2$ be fixed, and suppose $X_0$ takes values $1$ and $K$ with probabilities $\mathbb{P}(X_0 = 1) = p$ and $\mathbb{P}(X_0 = K) = 1 - p$, respectively. Then the ratio of the second moment and the first moment is
$$
\frac{\mathbb{E}(X_0^2)}{[\mathbb{E}(X_0)]^2} = f(p) = \frac{p + K^2(1 - p)}{(p + K(1 - p))^2}, \quad p \in (0,1).
$$
Let $q = 1 - p$, so $p = 1 - q$ and $q \in (0,1)$. Then we obtain a function $g(q)$ as
$$
g(q) = \frac{1 + (K^2 - 1)q}{(1 + (K - 1)q)^2}.
$$
Set $a = K^2 - 1$ and $b = K - 1$, so
$$
g(q) = \frac{aq + 1}{(bq + 1)^2}.
$$
To find the maximizer, take the derivative:
$$
g'(q) = \frac{a(bq + 1)^2 - 2b(aq + 1)(bq + 1)}{(bq + 1)^4}.
$$
Set the numerator equal to zero, we have
\begin{align*}
&a(bq + 1)^2 - 2b(aq + 1)(bq + 1) = 0 \Longrightarrow
a(bq + 1) - 2b(aq + 1) = 0 \\
 \Longrightarrow & abq + a - 2abq - 2b = 0 \Longrightarrow
-abq + a - 2b = 0 \Longrightarrow
abq = a - 2b \Longrightarrow q^\star = \frac{a - 2b}{ab}.
\end{align*}
Thus,
$$
p^\star = 1 - q^\star = 1 - \frac{a - 2b}{ab} = \frac{ab - a + 2b}{ab}.
$$
Recall $a = K^2 - 1$, $b = K - 1$, so
\begin{align*}
ab &= (K^2 - 1)(K - 1), \\
ab - a + 2b &= (K^2 - 1)(K - 2) + 2(K - 1),
\end{align*}
and hence
\begin{align*}
p^\star &= \frac{(K^2 - 1)(K - 2) + 2(K - 1)}{(K^2 - 1)(K - 1)} = \frac{K^3 - 2K^2 + K}{K^3 - K^2 - K + 1} = \frac{K}{K + 1}.
\end{align*}
Substitute $p = \frac{K}{K + 1}$ into the original function:
\begin{align*}
f\left( \frac{K}{K + 1} \right) &= \frac{\frac{K}{K + 1} + K^2 \cdot \frac{1}{K + 1}}{\left( \frac{K}{K + 1} + K \cdot \frac{1}{K + 1} \right)^2} = \frac{K + K^2}{(K + K)^2} \cdot (K + 1) = \frac{(K+1)^2}{4K}.
\end{align*}
Therefore, the function
$$
f(p) = \frac{p + K^2(1 - p)}{(p + K(1 - p))^2}
$$
achieves its maximum at
$$
p^\star = \frac{K}{K + 1}, \quad \text{with maximum value} \quad f(p^\star) = \frac{(K+1)^2}{4K}.
$$
Hence, the minimum SNR is 
$$
\text{SNR}_{min} = \frac{1}{\frac{(K+1)^2}{4K}-1} = \frac{4K}{(K-1)^2},
$$
ttained uniquely when $X_0$ is supported on $\{1, K\}$ with probability $\frac{K}{K + 1}$ on $1$ and $\frac{1}{K + 1}$ on $K$.

\noindent
\textbf{Step 4. Construction of $\psi_{\gamma}$.} Define the function $\psi$ as
\begin{align*}
    \psi_{\gamma}(k) = 
    \begin{cases}
    C+\log\left( \dfrac{K}{K+1} \right), & \text{if } k = 1, \\
    C+\log\left( \dfrac{1}{K+1} \right), & \text{if } k = K, \\
    -\infty, & \text{otherwise},
    \end{cases}
\end{align*}
for some constant $C$. Then the distribution induced by $\psi_{\gamma}$ satisfies
$$
\mathbb{P}(X_{\gamma} = 1) = \frac{e^{\psi_{\gamma}(1)}}{e^{\psi_{\gamma}(1)} + e^{\psi_{\gamma}(K)}} = \frac{K}{K+1}, \quad
\mathbb{P}(X_{\gamma} = K) = \frac{e^{\psi_{\gamma}(K)}}{e^{\psi_{\gamma}(1)} + e^{\psi_{\gamma}(K)}} = \frac{1}{K+1}.
$$
This completes the proof. \hfill ${\color{red}\blacksquare}$ \\

\noindent
\textbf{Proof of Theorem \ref{Thm:MinimalSNRNo}.} Note that when $K=2$, Theorem \ref{Thm:MinimalSNR} indicates that the minimal SNR is attained when 
$\mathbb{P}(X_{\gamma}=1)=\frac{K}{K+1}$ and $\mathbb{P}(X_{\gamma}=K)=\frac{1}{K+1}$, 
which follows the decreasing pattern. Therefore, under the decreasing constraint, the minimal SNR is also achieved by this distribution.

Next, we turn to show the result when $K \geq 3$. To begin with, for a probability mass vector $\bm{p} = (p_1, \ldots, p_K)$, let $X(\bm{p})$ denote the discrete random variable such that $\mathbb{P}(X(\bm{p}) = k) = p_k$. It is worth noting that under the constraint $p_1 \geq p_2 \geq \cdots \geq p_K$, the expectation of $X(\bm{p})$ cannot exceed $\frac{K+1}{2}$, with equality if and only if the distribution is uniform. We further define $\bm{p}(\psi_{\gamma}) = (p_1(\psi_{\gamma}), \ldots, p_K(\psi_{\gamma}))$ as the probability mass vector induced by a real-valued function $\psi_{\gamma}$, where $p_k(\psi_{\gamma}) = \frac{e^{\psi_{\gamma}(k)}}{\sum_{j=1}^K e^{\psi_{\gamma}(j)}} \quad \text{for } k = 1, \ldots, K$.

Let $\mathcal{H}(s)$ denote the set of all non-increasing probability mass vectors with fixed expectation $s$, defined as
$$
\mathcal{H}(s) = \left\{ \bm{p} : \sum_{k=1}^K p_k = 1, \; \sum_{k=1}^K k \cdot p_k = s, \; \text{and } p_1 \geq p_2 \geq \cdots \geq p_K \right\}.
$$
For any $s \in [1, \frac{K+1}{2}]$, we define $\bm{p}_s^\star$ as the minimizer of the signal-to-noise ratio (SNR) among all distributions in $\mathcal{H}(s)$:
$$
\bm{p}_s^\star = \argmin_{\bm{p} \in \mathcal{H}(s)} \text{SNR}(X(\bm{p})).
$$
Next, we aim to show that $\bm{p}_s^\star$ must satisfy $p^\star_{s,2} = p^\star_{s,3} = \cdots = p^\star_{s,K}$, indicating that $X(\bm{p}_s^\star)$ places equal probability mass on the values $\{2, \ldots, K\}$.

\textbf{Proof by Contradiction.} We assume that $\bm{p}_s^\star$ does not satisfy $p^\star_{s,2} = p^\star_{s,3} = \cdots = p^\star_{s,K}$. This implies that there exists $k \in \{2,\ldots,K\}$ such that $p_{s,k}^\star>p_{s,k+1}^\star$. Then, we define a new mass vector $\bm{p}^\varepsilon = (p_1^\varepsilon, \ldots, p_K^\varepsilon)$ by transferring mass $\varepsilon$ from $p_{s,k}^\star$ to $p_{s,1}^\star$ and $p_{s,k+1}^\star$ as follows:
\begin{align*}
\text{Construction of $\bm{p}^\varepsilon$: }
\begin{cases}
p_k^\varepsilon = p_{s,k}^\star - \varepsilon, \\
p_1^\varepsilon = p_{s,1}^\star + \dfrac{\varepsilon}{k}, \\
p_{k+1}^\varepsilon = p_{s,k+1}^\star + \left(1 - \dfrac{1}{k} \right) \varepsilon, \\
p_i^\varepsilon = p_{s,i}^\star \quad \text{for } i \notin \{1, k, k+1\}.
\end{cases}    
\end{align*}

\paragraph{Step 1: Total Probability is Preserved}
\[
\sum_{i=1}^K p_i^\varepsilon = \sum_{i=1}^K p_{s,i}^\star - \varepsilon + \frac{\varepsilon}{k} + \left(1 - \frac{1}{k} \right)\varepsilon = \sum_{i=1}^K p_i = 1.
\]
Here, at the new probability mass vector, we can choose $\varepsilon$ sufficiently small so that the following inequality is preserved.
\begin{align*}
    p_1^\varepsilon \geq p_2^\varepsilon \geq \cdots \geq p_K^\varepsilon.
\end{align*}
\paragraph{Step 2: Expectation is Preserved}

Note that $\mathbb{E}[X(\bm{p}_s^\star)]=s$. Then:
\begin{align*}
    \mathbb{E}[X(\bm{p}^\varepsilon)] &= s - \varepsilon \cdot k + \dfrac{\varepsilon}{k} \cdot 1 + \left(1 - \dfrac{1}{k} \right)\varepsilon \cdot (k+1) \\
&= s + \varepsilon \left( \dfrac{1}{k} - k + \left(1 - \dfrac{1}{k} \right)(k+1) \right)=s.
\end{align*}
This shows that $\bm{p}^\varepsilon \in \mathcal{H}(s)$.

\paragraph{Step 3: Variance Increases}

We consider the change in second moment:
\begin{align*}
\delta_{\varepsilon} &= \mathbb{E}\left\{[X(\bm{p}^\varepsilon)]^2\right\} - \mathbb{E}\left\{[X(\bm{p}_s^\star)]^2\right\} \\
&= -\varepsilon \cdot k^2 + \frac{\varepsilon}{k} \cdot 1^2 + \left(1 - \frac{1}{k} \right)\varepsilon \cdot (k+1)^2 \\
&= \varepsilon \left[ -k^2 + \frac{1}{k} + \left(1 - \frac{1}{k} \right)(k+1)^2 \right] \\
&=  \frac{\varepsilon}{k} \left[ -k^3 + 1 + (k - 1)(k^2 + 2k + 1) \right] \\
& = \frac{\varepsilon}{k} \left(
k^2-k
\right) = \varepsilon(k-1)>0,
\end{align*}
for any $k \geq 2$. One can verify 
\begin{align*}
    \text{SNR}(X(\bm{p}^\varepsilon)) - \text{SNR}(X(\bm{p}_{s}^\star)) = \frac{s^2}{\mathbb{E}\left\{[X(\bm{p}^\varepsilon)]^2\right\}-s^2} - \frac{s^2}{\mathbb{E}\left\{[X(\bm{p}_s^\star)]^2\right\}-s^2}<0
\end{align*}
Because the perturbation $\varepsilon$ only changes three coordinates slightly, strict monotonicity is preserved for sufficiently small $\varepsilon$. This contradicts with the assumption that $\bm{p}_{s}^\star$ minimizes $\text{SNR}(X(\bm{p}))$ within $\mathcal{H}(s)$. Therefore, we conclude that for any $s \in [1, \frac{K+1}{2}]$, the minimal SNR achievable over $\mathcal{H}(s)$ must satisfy
$$
p_{s,2}^\star = p_{s,3}^\star = \cdots = p_{s,K}^\star.
$$
Therefore, the optimal $\bm{p}$ that minimizes $\text{SNR}(X(\bm{p}))$ must also satisfy this condition.

Next, we intend to investigate which $\bm{p}=(p_1,\ldots,p_k)$ with $p_2=\ldots = p_K$ gives the minimum $\text{SNR}(X(\bm{p}))$. For a $\bm{p}$, let $X(\bm{p})$ be a discrete random variable supported on $\{1, 2, \ldots, K\}$, with probability mass function
$$
\mathbb{P}(X(\bm{p}) = 1) = p_1, \quad \mathbb{P}(X(\bm{p}) = i) = p \quad \text{for } i = 2, \ldots, K,
$$
where $p_1 \ge p>0$ and $p_1 + (K - 1)p = 1$. The goal is to minimize the SNR with respect to $\bm{p}$ defined by
$$
\mathrm{SNR}(X(\bm{p})) = \frac{(\mathbb{E}[X(\bm{p})])^2}{\operatorname{Var}(X(\bm{p}))}.
$$
Using the constraint $p_1 = 1 - (K - 1)p$, we write
\begin{align*}
\mathbb{E}[X(\bm{p})] &= 1 \cdot (1 - (K - 1)p) + \sum_{i=2}^K i \cdot p 
= 1 - (K - 1)p + p \sum_{i=2}^K i \\
&= 1 - (K - 1)p + p\left( \frac{K(K+1)}{2} - 1 \right)
 \\
& = p \frac{K(K-1)}{2}+1 \triangleq \mu(p).
\end{align*}
Similarly, for the second moment:
\begin{align*}
\mathbb{E}\left\{[X(\bm{p})]^2\right\} &= 1^2 \cdot (1 - (K - 1)p) + \sum_{i=2}^K i^2 \cdot p \\
&= 1 - (K - 1)p + p \left( \frac{K(K+1)(2K+1)}{6} - 1 \right) \\
& = p \left( \frac{K(K+1)(2K+1)}{6} - K \right)+1
= M_2(p).
\end{align*}
Thus, the variance is $\operatorname{Var}(X(\bm{p})) = \sigma^2(p) = M_2(p) - \mu(p)^2$. We define the function
$$
f(p) \triangleq \frac{\mu(p)^2}{\sigma^2(p)} = \frac{(a_1 p + 1)^2}{b_1 p + 1 - (a_1 p + 1)^2},
$$
where
$$
a_1 = \frac{K(K - 1)}{2}, \quad
b_1 = \frac{K(K + 1)(2K + 1)}{6} - K.
$$
To simplify notation, define
$$
u = a_1 p + 1,
$$
which implies
$$
p = \frac{u - 1}{a_1}, \quad u \in \left[1,\, a_1 \cdot \frac{1}{K - 1} + 1\right].
$$
Rewriting $f$ as a function of $u$, we have
$$
f(u) = \frac{u^2}{b_1 \cdot \frac{u - 1}{a_1} + 1 - u^2}
= \frac{u^2}{\frac{b_1}{a_1}(u - 1) + 1 - u^2}.
$$
Set the constant
$$
c = \frac{b_1}{a_1} = \frac{\frac{K(K + 1)(2K + 1)}{6} - K}{\frac{K(K - 1)}{2}}
= \frac{(K + 1)(2K + 1) - 6}{3(K - 1)}.
$$
Thus, $f(u)$ can be written as
$$
f(u) = \frac{u^2}{c(u - 1) + 1 - u^2} = \frac{u^2}{D(u)},
$$
where $D(u) = c(u - 1) + 1 - u^2$. Differentiating $f(u)$ with respect to $u$, we have
\begin{align*}
    f'(u) =
    \frac{2uD(u)-u^2(c-2u)}{D(u)^2}
    =\frac{u (c u - 2 c + 2)}{D(u)^2}.
\end{align*}
Setting $f'(u) = 0$ to find critical points yields
$$
u = 0 \quad \text{or} \quad c u - 2 c + 2 = 0.
$$
Since $u = a_1 p + 1 \geq 1$, we discard $u = 0$. Solving for $u$ gives
$$
u = \frac{2 c - 2}{c} = 2 - \frac{2}{c}.
$$
Returning to variable $p$,
$$
p^\star = \frac{u - 1}{a_1} = \frac{2 - \frac{2}{c} - 1}{a_1} = \frac{1 - \frac{2}{c}}{a_1}.
$$
Substituting the expressions for $a_1$ and $c$,
$$
a_1 = \frac{K(K - 1)}{2}, \quad
c = \frac{2 K^2 + 3 K - 5}{3 (K - 1)}.
$$
Calculate the numerator,
$$
1 - \frac{2}{c} = \frac{2 K^2 + 3 K - 5 - 6 (K - 1)}{2 K^2 + 3 K - 5} = \frac{2 K^2 - 3 K + 1}{2 K^2 + 3 K - 5}.
$$
Therefore,
$$
p^\star = \frac{1 - \frac{2}{c}}{a_1} = \frac{2 (2 K^2 - 3 K + 1)}{K (K - 1) (2 K^2 + 3 K - 5)} =\frac{2(2K-1)}{K(K-1)(2K+5)}.
$$
This value $p^\star$ lies within the allowed domain and corresponds to the minimum of $f(p)$. With this, 
\begin{align*}
    p_1 = 1- p^\star (K-1) = 1-\frac{4K-2}{2K^2+5K} = \frac{2K^2+K+2}{2K^2+5K}.
\end{align*}
To make $X(\bm{p}(\psi_{\gamma}))$ achieve this minimum SNR, we can choose
\begin{align*}
    \psi_{\gamma}(k) = \begin{cases}
        C+\log\left(\frac{(2K^2+K+2)(K-1)}{2(2K-1)}\right), &\text{ if } k=1, \\
        C, & \text{ if } 2 \leq k \leq K,
    \end{cases}
\end{align*}
for any $C\in \mathbb{R}$. Plugging this $\psi_{\gamma}$ to $\text{SNR}(\bm{p}(\psi_{\gamma}))$ yields that
\begin{align*}
\text{SNR}_{min} = \frac{24(K+1)}{4K^2-4K+1}.
\end{align*}
Particularly, when $K=2$,
\begin{align*}
    \frac{24(K+1)}{4K^2-4K+1} = \frac{72}{9} = \frac{4K}{(K-1)^2} = 8.
\end{align*}
This shows that the lower bounds in Theorems \ref{Thm:MinimalSNR} and \ref{Thm:MinimalSNRNo} match. This completes the proof. \hfill ${\color{red}\blacksquare}$ \\

{ 
\noindent
\textbf{Proof of Theorem \ref{Prop:SNR}.} The proof of Theorem \ref{Prop:SNR} is structured into the following steps. First, by the definition of $\textnormal{SNR}(W)$, we have
\begin{align*}
    \textnormal{SNR}(W) = \frac{(\mathbb{E}[W])^2}{\textnormal{Var}(W)}=\frac{(\mathbb{E}[W])^2}{\mathbb{E}[W^2]-[\mathbb{E}(W)]^2} = \frac{1}{1-\frac{(\mathbb{E}[W])^2}{\mathbb{E}[W^2]}}.
\end{align*}

\noindent\textbf{Step 1. Decomposition into sign and magnitude.}  
Define the sign and magnitude of $W$ as
\[
S \triangleq \operatorname{sign}(W) \in \{-1,+1\}, \qquad M \triangleq  |W| \ge 0,
\]
so that $W = S \cdot M$. Let $P \triangleq  \mathbb{P}(S=1)$. Then
\[
\mathbb{E}[W] = \mathbb{E}[SM] = \mathbb{E}\big[\, \mathbb{E}[S M \mid M] \, \big] = \mathbb{E}[ M \, \mathbb{E}[S \mid M] ].
\]
Define $q(m) \triangleq  \mathbb{P}(S=1 \mid M=m)$. Then
\[
\mathbb{E}[S \mid M=m] = 2q(m)-1,
\]
so that
\[
\mathbb{E}[W] = \mathbb{E}[ M (2q(M)-1) ], \qquad \mathbb{E}[W^2] = \mathbb{E}[M^2], \qquad \operatorname{Var}(W) = \mathbb{E}[M^2] - (\mathbb{E}[ M (2q(M)-1) ])^2.
\]

\noindent\textbf{Step 2. Bounding the mean via Cauchy-Schwarz.}  
By the Cauchy-Schwarz inequality, we have
\[
(\mathbb{E}[W])^2 = (\mathbb{E}[ M (2q(M)-1) ])^2 \le \mathbb{E}[M^2] \, \mathbb{E}[(2q(M)-1)^2].
\]
Hence,
\begin{align}
\label{JS}
   \text{SNR}(W) &= \frac{(\mathbb{E}[W])^2}{\mathbb{E}[M^2] - (\mathbb{E}[W])^2} \notag \\
   &\le \frac{\mathbb{E}[M^2] \, \mathbb{E}[(2q(M)-1)^2]}{\mathbb{E}[M^2] - \mathbb{E}[M^2]  \mathbb{E}[(2q(M)-1)^2]} \\
   &= \frac{\mathbb{E}[(2q(M)-1)^2]}{1 - \mathbb{E}[(2q(M)-1)^2]}. \notag
\end{align}
This shows that the maximum SNR depends only on the distribution of $S$.

By the assumption that for $w_1,w_2>0$,
\begin{align*}
    \frac{P_{W}(w_1)}{P_{W}(w_2)}=
   \frac{P_{W}(-w_1)}{P_{W}(-w_2)}
   \Leftrightarrow 
   \frac{P_{W}(w_1)}{P_{W}(-w_1)} = c
\end{align*}
for some positive constant $c$. Therefore, we have
\begin{align*}
    \int_{0}^{\infty }P_{W}(w_1) dw_1 = c 
    \int_{0}^{\infty }P_{W}(-w_1) dw_1 = c \int_{-\infty }^0 P_{W}(w_1) dw_1.
\end{align*}
Using the facts that $\int_{-\infty}^{\infty }P_{W}(w_1) dw_1=1$ and $\int_{0}^{\infty }P_{W}(w_1) dw_1=P$, we have $c = \frac{P}{1-P}$. Furthermore, 
\begin{align*}
    q(M) = \mathbb{P}(S=1 \mid M=m)=\frac{\mathbb{P}(S=1, M=m)}{\mathbb{P}(M=m)}=\frac{P_W(m)}{P_W(m)+P_W(-m)}=P.
\end{align*}
Therefore, the upper bound becomes
\begin{align*}
    \frac{\mathbb{E}[(2q(M)-1)^2]}{1 - \mathbb{E}[(2q(M)-1)^2]} =\frac{\mathbb{E}[(2q(M)-1)]^2}{1 - \mathbb{E}[(2q(M)-1)]^2} =\frac{(2P-1)^2}{4P(1-P)}.
\end{align*}

\noindent\textbf{Step 3. Reduction to a Bernoulli variable.}  
In the optimal case, write $W = M \cdot S$ with $M$ constant. Then
\[
\text{SNR}(W) = \frac{(\mathbb{E}[S])^2}{\operatorname{Var}(S)}.
\]
Since $S \in \{-1,+1\}$ with $\mathbb{P}(S=1) = P$, we have
\[
\mathbb{E}[S] = 2P-1, \qquad \operatorname{Var}(S) = 1 - (2P-1)^2 = 4P(1-P).
\]
Hence, the maximal SNR is
\[
\text{SNR}(W) = \frac{(2P-1)^2}{4P(1-P)}.
\]
This shows that the upper bound of $\text{SNR}(W)$ is achieved when $W$ follows a two-point distribution. Moreover, the equality in (\ref{JS}) holds only if $M = c_0 \cdot (2q(M)-1)$ for some constant $c_0$. Since $q(M) = P$, this condition is satisfied precisely when $M$ is a constant. This completes the proof. \hfill ${\color{red}\blacksquare}$ \\

\color{black}
\noindent
\textbf{Proof of Theorem \ref{Thm:Appendix}.} In this proof, we mainly consider two cases: (1) $\psi_{\gamma_{ij}}$'s are well specified as $\psi_{\gamma_{ij}^\star}$'s, which are fixed during the optimization task and (2) $\psi_{\gamma_{ij}^\star}$'s are unknown and $\psi_{\gamma}$ is a continuous function of $\gamma$ but $\frac{\partial \psi_{\gamma}(k)}{\partial \gamma}$ does not depend on $k$.

% but the functional form with respect to $\gamma_{ij}$ is specified correctly.

\noindent
\textbf{Case 1:} If $\psi_{\gamma_{ij}}$'s are well specified as $\psi_{\gamma_{ij}^\star}$'s, then $\mathcal{L}(\bm \theta)$ becomes
\begin{align*}
\mathcal{L}(\bm \theta) 
=  & \sum_{1 \le i < j \le n} \sum_{l=1}^{L} \Bigg[
\phi\big(\operatorname{sign}(y_{ij}^{(l)})(\theta_i-\theta_j) \big) +\psi_{\gamma_{ij}^\star}(y_{ij}^{(l)}) \\
&- \log \sum_{k \in \Upsilon(K)} \exp \Big( \phi(\operatorname{sign}(k) (\theta_i-\theta_j)) + \psi_{\gamma_{ij}^\star}(k) \Big)
\Bigg].
\end{align*}
Define
$$
p_{ij}(k) = \frac{\exp\big(\phi(\operatorname{sign}(k)\gamma_{ij}) + \psi_{\gamma_{ij}^\star}(k)\big)}
{\sum_{k' \in \Upsilon(K)} \exp\big(\phi(\operatorname{sign}(k')\gamma_{ij}) + \psi_{\gamma_{ij}^\star}(k')\big)}.
$$
Then, the partial derivative of the log-likelihood $\mathcal{L}(\bm \theta) $ with respect to $\theta_i$ is
\begin{align*}
    \frac{\partial \mathcal{L}(\bm \theta)}{\partial \theta_i} 
= &\sum_{j \neq i} \sum_{l=1}^{L} \Bigg[
\phi'\big(\operatorname{sign}(y_{ij}^{(l)}) (\theta_i-\theta_j)\big) \operatorname{sign}(y_{ij}^{(l)}) 
- \sum_{k \in \Upsilon(K)} p_{ij}(k) \, \phi'\big(\operatorname{sign}(k) (\theta_i-\theta_j)\big) \operatorname{sign}(k)
\Bigg] \\
= &\sum_{j \neq i}\phi'\big(\theta_i-\theta_j\big) \sum_{l=1}^{L} \Bigg[\operatorname{sign}(y_{ij}^{(l)}) 
- \sum_{k \in \Upsilon(K)} p_{ij}(k) \,  \operatorname{sign}(k)
\Bigg] \\
= & \sum_{j \neq i}\phi'\big(\theta_i-\theta_j\big) \sum_{l=1}^{L} 
\Bigg[\operatorname{sign}(y_{ij}^{(l)}) 
- \sum_{k =1}^K p_{ij}(k)+\sum_{k =-K}^{-1} p_{ij}(k)
\Bigg] \\
= &  \sum_{j \neq i}2\phi'\big(\theta_i-\theta_j\big)\sum_{l=1}^{L} 
\Bigg[\frac{\operatorname{sign}(y_{ij}^{(l)})+1}{2} 
- \frac{\exp(2\phi(\theta_i-\theta_j))}{1+\exp(2\phi(\theta_i-\theta_j))}
\Bigg],
\end{align*}
where the second equality follows from the fact that $\phi'(\cdot)$ is an even function and the last equality follows from the fact that
\begin{align*}
     \sum_{k =1}^K p_{ij}(k) = &
     \frac{\sum_{k =1}^K\exp\big(\phi(\operatorname{sign}(k)\gamma_{ij}) + \psi_{\gamma_{ij}^\star}(k)\big)}
{\sum_{k' \in \Upsilon(K)} \exp\big(\phi(\operatorname{sign}(k')\gamma_{ij}) + \psi_{\gamma_{ij}^\star}(k')\big)} \\
= &\frac{\exp(\phi(\gamma_{ij}) )\sum_{k =1}^K\exp\big( \psi_{\gamma_{ij}^\star}(k)\big)}
{[\exp(\phi(\gamma_{ij}))+\exp(\phi(-\gamma_{ij}))]\sum_{k'=1}^K \exp\big( \psi_{\gamma_{ij}^\star}(k')\big)} \\
= & \frac{\exp(2\phi(\gamma_{ij}))}{1+\exp(2\phi(\gamma_{ij}))}.
\end{align*}
Clearly, the derivative $\frac{\partial \mathcal{L}(\bm \theta)}{\partial \theta_i}$ is independent of the value of $K$. Consequently, the solution $\widehat{\bm{\theta}}$ that satisfies the following system of equations is also invariant to $K$.
\begin{align*}
 \frac{\partial \mathcal{L}(\bm \theta)}{\partial \theta_1}\Big|_{\theta_1=\widehat{\theta}_1} = 0 \quad 
  \frac{\partial \mathcal{L}(\bm \theta)}{\partial \theta_2}\Big|_{\theta_2=\widehat{\theta}_2} = 0 \quad \cdots
  \frac{\partial \mathcal{L}(\bm \theta)}{\partial \theta_n}
       \Big|_{\theta_n=\widehat{\theta}_n}= 0.
\end{align*}
Note that when $\phi = x/2$, $\widehat{\bm \theta}$ is equivalent to the solution for the BTL model.

\noindent\textbf{Case 2}: Suppose $\psi_{\gamma_{ij}^\star}$ is unknown and $\psi_{\gamma}$ is a continuous function of $\gamma$ with the functional form being correctly specified. Additionally, $\frac{\partial \psi_{\gamma}(k)}{\partial \gamma}$ does not depend on $k$. Define
\begin{align*}
    p(k,\gamma) = \frac{\exp\big(\phi(\operatorname{sign}(k)\gamma) + \psi_{\gamma}(k)\big)}
{\sum_{k' \in \Upsilon(K)} \exp\big(\phi(\operatorname{sign}(k')\gamma) + \psi_{\gamma}(k')\big)}
\end{align*}

Note that $\partial \psi_{\gamma_{ij}}/\partial \theta_i = \partial \psi_{\gamma_{ij}}/\partial \gamma_{ij}$. Then, the partial derivative of the log-likelihood $\mathcal{L}(\bm \theta) $ with respect to $\theta_i$ is
\begin{align*}
        \frac{\partial \mathcal{L}(\bm \theta)}{\partial \theta_i} 
= &\sum_{j \neq i} \sum_{l=1}^{L} \Bigg[
\phi'\big(\operatorname{sign}(y_{ij}^{(l)}) (\theta_i-\theta_j)\big) \operatorname{sign}(y_{ij}^{(l)}) 
- \sum_{k \in \Upsilon(K)} p_{ij}(k) \, \phi'\big(\operatorname{sign}(k) (\theta_i-\theta_j)\big) \operatorname{sign}(k)
\Bigg] \\
+ & \sum_{j \neq i} \sum_{l=1}^{L} \underbrace{\Bigg[
\frac{\partial \psi_{\gamma_{ij}}(y_{ij}^{(l)})}{\partial \theta_i}
- \sum_{k \in \Upsilon(K)} p(k,\gamma_{ij}) \frac{\partial \psi_{\gamma_{ij}}(k)}{\partial \theta_i}
\Bigg]}_{=0} \\
= &\sum_{j \neq i}2\phi'\big(\theta_i-\theta_j\big)\sum_{l=1}^{L} 
\Bigg[\frac{\operatorname{sign}(y_{ij}^{(l)})+1}{2} 
- \frac{\exp(2\phi(\theta_i-\theta_j))}{1+\exp(2\phi(\theta_i-\theta_j))}
\Bigg],
\end{align*}
where the last equation follows from the Assumption that $\frac{\partial \psi_{\gamma_{ij}}(k)}{\partial \theta_i}$ does not depend on $k$. Using similar arguments as in Case 1 leads to the desired results. This completes the proof. \hfill $\blacksquare$ \\
}

\begin{theorem}[Cram\'er's Theorem; \citet{klenke2013probability}]
\label{Thm:Cramer}
Let $(X_i)_{i\geq1}$ be i.i.d. real-valued random variables such that $\mathbb{E}[e^{\lambda X_1}] < \infty$ for all $\lambda \in \mathbb{R}$ and $S_n=\sum_{i\in[n]}X_i$. Then for any $a > \mathbb{E}[X_1]$,
\[
\lim_{n \to \infty} \frac{1}{n} \log \mathbb{P}(S_n \geq a n) = - I(a),
\]
where $I(a) = \sup_{\lambda \in \mathbb{R}} \left[ a \lambda - \log \mathbb{E}[e^{\lambda X_1}] \right]$. Particularly, when $a=0$, we have
\begin{align*}
    \lim_{n \to \infty} \frac{1}{n} \log \mathbb{P}(S_n \geq 0) = - I(0) = -\sup_{\lambda \in \mathbb{R}} \left\{- \log \mathbb{E}[e^{\lambda X_1}] \right\}.
\end{align*}
\end{theorem}

Theorem~\ref{Thm:Cramer}, known as Cramér's Theorem, is a well-established result whose detailed proof can be found in Theorem 23.3 of \citet{klenke2013probability}.

\section{Proof of Lemmas}

\begin{lemma}
    \label{Lemma:RateFunction}
Suppose that $y_{ij}^{(l)} \sim G(\phi, \psi_{\gamma_{ij}^\star}, \gamma_{ij}^\star, K)$, and let the corresponding rate function be defined as
$$
I(z) = \sup_{\lambda \in \mathbb{R}} \left\{ z \lambda - \log \mathbb{E}[e^{\lambda a_{ij}^{(l)}  y_{ij}^{(l)}}] \right\}.
$$
Then, the rate function evaluated at zero satisfies
\begin{align*}
    I(0) =  &
\log\left(\frac{\cosh(\phi(\gamma_{ij}^\star))}{p+(1-p)\cosh(\phi(\gamma_{ij}^\star))}\right) \\
&- \inf_{\lambda \in \mathbb{R}} \log \left(\frac{ \frac{p\sum_{k = 1}^K e^{\psi_{\gamma_{ij}^\star}(k)} \cosh\left( \phi(\gamma_{ij}^\star) + \lambda k \right)}{\sum_{k = 1}^K e^{\psi_{\gamma_{ij}^\star}(k)}}+(1-p) \cosh\left( \phi(\gamma_{ij}^\star) \right)}{p+(1-p)\cosh(\phi(\gamma_{ij}^\star))}\right),
\end{align*}
where $p = \mathbb{E}(a_{ij}^{(l)})$. In particular, when $K = 1$, we have
$$
I(0) = 
\log\left(\frac{\cosh(\phi(\gamma_{ij}^\star))}{p+(1-p)\cosh(\phi(\gamma_{ij}^\star))}\right).
$$
\end{lemma}

\noindent
\textbf{Proof of Lemma \ref{Lemma:RateFunction}.} Note that the probability mass function $\mathbb{P}(y_{ij}^{(l)}=k)$ is given by
\[
\mathbb{P}(y_{ij}^{(l)}=k) \propto \exp\left( \phi(\operatorname{sign}(k) \cdot \gamma_{ij}^\star) + \psi_{\gamma_{ij}^\star}(k) \right) \quad k \in \Upsilon(K).
\]
Since $a_{ij}^{(l)}$ is independent of $y_{ij}^{(l)}$ and follows $\mathrm{Bernnoulli}(p)$, the moment generating function (MGF) of $X$ is:
\begin{align*}
    M(\lambda) =& \mathbb{E}[e^{\lambda a_{ij}^{(l)}y_{ij}^{(l)}}] =
p\mathbb{E}[e^{\lambda y_{ij}^{(l)}}]+1-p \\
= &p
\sum_{k = 1}^K \left( \mathbb{P}(y_{ij}^{(l)} = k) e^{\lambda k} + \mathbb{P}(y_{ij}^{(l)} = -k) e^{-\lambda k} \right) + 1-p.
\end{align*}
Substituting the expressions yields that
\begin{align*}
    M(\lambda) =&\frac{p}{\Psi_{\phi,\psi_{\gamma_{ij}^\star}}(\gamma_{ij}^\star)} \sum_{k = 1}^K e^{\psi_{\gamma_{ij}^\star}(k)} \left( e^{\phi(\gamma_{ij}^\star) + \lambda k} + e^{-\phi(\gamma_{ij}^\star) - \lambda k} \right)+1-p \\
= &\frac{2p}{\Psi_{\phi,\psi_{\gamma_{ij}^\star}}(\gamma_{ij}^\star)} \sum_{k = 1}^K \left[e^{\psi_{\gamma_{ij}^\star}(k)} \cosh(\phi(\gamma_{ij}^\star) + \lambda k)\right] +1-p.
\end{align*}
By the definition of the Cram\'er rate function, we have
\[
I(0) = - \inf_{\lambda \in \mathbb{R}} \log M(\lambda).
\]
Substituting the MGF, we have
\[
I(0) = - \inf_{\lambda \in \mathbb{R}} \log \left( \frac{2p}{\Psi_{\phi,\psi_{\gamma_{ij}^\star}}(\gamma_{ij}^\star)}  \sum_{k = 1}^K e^{\psi_{\gamma_{ij}^\star}(k)} \cosh(\phi(\gamma_{ij}^\star) + \lambda k) +1-p\right).
\]
Using the expression for $y_{ij}^{(l)}$, we have
\[
\Psi_{\phi,\psi_{\gamma_{ij}^\star}}(\gamma_{ij}^\star) = 2 \cdot  \cosh(\phi(\gamma_{ij}^\star)) \sum_{k = 1}^K e^{\psi_{\gamma_{ij}^\star}(k)}.
\]
We obtain
\begin{align*}
I(0)
&= - \inf_{\lambda \in \mathbb{R}} \log \left( \frac{ p\sum_{k = 1}^K e^{\psi_{\gamma_{ij}^\star}(k)} \cosh(\phi(\gamma_{ij}^\star) + \lambda k) }{ \cosh(\phi(\gamma_{ij}^\star)) \sum_{k = 1}^K e^{\psi_{\gamma_{ij}^\star}(k)} } +1-p\right) \\
&= 
\log \left(  \frac{\cosh(\phi(\gamma_{ij}^\star))}{p+(1-p)\cosh(\phi(\gamma_{ij}^\star))} \right)
-  Q(p,\gamma_{ij}^\star)
\end{align*}
where $ Q(p,\gamma_{ij}^\star)$ is defined as
\begin{align*}
  Q(p,\gamma_{ij}^\star)=  \inf_{\lambda \in \mathbb{R}} \log \left(\frac{ \frac{p\sum_{k = 1}^K e^{\psi_{\gamma_{ij}^\star}(k)} \cosh(\phi(\gamma_{ij}^\star) + \lambda k)}{\sum_{k = 1}^K e^{\psi_{\gamma_{ij}^\star}(k)}}+(1-p)\cosh(\phi(\gamma_{ij}^\star)) }{p+(1-p)\cosh(\phi(\gamma_{ij}^\star))}\right).
\end{align*}
Here, it is worth noting that since $\cosh(x) \geq 1$ for any $x \in \mathbb{R}$. Note that by the Jensen's inequality, we have
\begin{align*}
    \frac{\sum_{k = 1}^K e^{\psi_{\gamma_{ij}^\star}(k)} \cosh(\phi(\gamma_{ij}^\star) + \lambda k)}{\sum_{k = 1}^K e^{\psi_{\gamma_{ij}^\star}(k)}}\ge
    \cosh\left(
\phi(\gamma_{ij}^\star)+\lambda \cdot \frac{\sum_{k = 1}^K k\cdot e^{\psi_{\gamma_{ij}^\star}(k)}}{\sum_{k = 1}^K e^{\psi_{\gamma_{ij}^\star}(k)}}
    \right) = \cosh \left(\phi(\gamma_{ij}^\star)+\lambda\mathbb{E}(X_{\gamma_{ij}^\star})\right).
\end{align*}
Therefore, if $X_{\gamma_{ij}^\star} \sim \text{Geo}(\psi_{\gamma_{ij}^\star},K)$ is not a degenerate distribution (mass on a single point), we always have
\begin{align*}
    Q(p,\gamma_{ij}^\star) > \inf_{\lambda \in \mathbb{R}} \log \left(\frac{ p \cdot\cosh \left(\phi(\gamma_{ij}^\star)+\lambda\mathbb{E}(X_{\gamma_{ij}^\star})\right)+(1-p)\cosh(\phi(\gamma_{ij}^\star)) }{p+(1-p)\cosh(\phi(\gamma_{ij}^\star))}\right)
    =0.
\end{align*}
To sum up, we have
\begin{align*}
\text{When $K=1$: }\, & \inf_{\lambda \in \mathbb{R}}  \frac{p\sum_{k = 1}^K e^{\psi_{\gamma_{ij}^\star}(k)} \cosh(\phi(\gamma_{ij}^\star) + \lambda k)}{\sum_{k = 1}^K e^{\psi_{\gamma_{ij}^\star}(k)}} =  \inf_{\lambda \in \mathbb{R}} p \cosh(\phi(\gamma_{ij}^\star) + \lambda ) = p,\\
\text{When $K \geq 2$: }\, & 
\inf_{\lambda \in \mathbb{R}}  \frac{p\sum_{k = 1}^K e^{\psi_{\gamma_{ij}^\star}(k)} \cosh(\phi(\gamma_{ij}^\star) + \lambda k)}{\sum_{k = 1}^K e^{\psi_{\gamma_{ij}^\star}(k)}} >p.
\end{align*}
Particularly, when $K=1$, the rate function $I(0)$ becomes
\begin{align*}
    I(0) =& \log \left(  \cosh(\phi(\gamma_{ij}^\star)) \right)-
    \inf_{\lambda \in \mathbb{R}} \log \left( p\cosh(\phi(\gamma_{ij}^\star) + \lambda)+(1-p)\cosh(\phi(\gamma_{ij}^\star)) \right) \\
    = &\log \left(  \cosh(\phi(\gamma_{ij}^\star)) \right)-\log(p+(1-p)\cosh(\phi(\gamma_{ij}^\star)) \\
    = & \log\left(\frac{\cosh(\phi(\gamma_{ij}^\star))}{p+(1-p)\cosh(\phi(\gamma_{ij}^\star))}\right),
\end{align*}
where the last equality follows by taking $\lambda = -\phi(\gamma_{ij}^\star)$. This completes the proof. \hfill ${\color{red}\blacksquare}$. \\

\begin{lemma}
\label{Lemma:Convergence}
Let $N\in\mathbb{N}$ be fixed. For each $i=1,\dots,N$ let $A_i(L)\ge0$ and $B_i(L)>0$ be functions of $L$.
If $\frac{A_i(L)}{B_i(L)}\xrightarrow{L\to\infty} 0$ for every $i$, then
\[
\frac{\sum_{i=1}^N A_i(L)}{\sum_{i=1}^N B_i(L)}\xrightarrow{L\to\infty} 0.
\]
\end{lemma}

\noindent \textbf{Proof of Lemma \ref{Lemma:Convergence}.} Fix $\varepsilon>0$. For each $i$ there exists $L_i$ such that $A_i(L)/B_i(L)<\varepsilon$ for all $L\ge L_i$. Set $L_0=\max_{1\le i\le N}L_i$. Then for $L\ge L_0$ we have $A_i(L)/B_i(L)<\varepsilon$ for all $i$. Writing
\[
\frac{\sum_{i=1}^N A_i(L)}{\sum_{i=1}^N B_i(L)}
=\sum_{i=1}^N \frac{B_i(L)}{\sum_{j=1}^N B_j(L)}\cdot\frac{A_i(L)}{B_i(L)},
\]
and noting that the weights $w_i(L)=B_i(L)/\sum_{j=1}^N B_j(L)$ are nonnegative and sum to~1, we obtain for $L\ge L_0$
\[
\frac{\sum_{i=1}^N A_i(L)}{\sum_{i=1}^N B_i(L)}\le\sum_{i=1}^N w_i(L)\varepsilon=\varepsilon.
\]
Since $\varepsilon>0$ is arbitrary, the result follows. \hfill ${\color{red}\blacksquare}$. \\

\begin{lemma}
    \label{Lemma:SumVarRate}
    For $i,j\in[n]$ and $l \in [L]$, suppose that $y_{ij}^{(l)} \sim G\big(\phi, \psi_{\gamma_{ij}^\star}, \gamma_{ij}^\star, K\big)$ with $\theta_i^\star > \theta_j^\star$ for $i < j$, and let $ X_{\gamma_{ij}^\star} \sim \mathrm{Geo}(\psi_{\gamma_{ij}^\star}, K) $ for $i\neq j$ and $K \geq 2$. Then, for every $i<j$, we have
    \begin{align*}
  \mathbb{P}\left(-\sum_{l=1}^L a_{ij}^{(l)}y_{ij}^{(l)} \geq 0\right) \xrightarrow{L\rightarrow\infty} 0 \mbox{ and }
  \mathbb{P}\left(-\sum_{l=1}^L a_{ij}^{(l)}\sign(y_{ij}^{(l)}) \geq 0\right) \xrightarrow{L\rightarrow\infty} 0.
    \end{align*}
    In addition, we have
    \begin{align*}
       \lim_{L\rightarrow\infty} \frac{\mathbb{P}\left(-\sum_{l=1}^L a_{ij}^{(l)}\sign(y_{ij}^{(l)}) \geq 0\right)}{ \mathbb{P}\left(-\sum_{l=1}^L a_{ij}^{(l)}y_{ij}^{(l)} \geq 0\right)} =0.
    \end{align*}
\end{lemma}

\noindent
\textbf{Proof of Lemma \ref{Lemma:SumVarRate}.}  The proof of Lemma \ref{Lemma:SumVarRate} mainly uses the result of Theorem \ref{Thm:Cramer} and Lemma \ref{Lemma:RateFunction}. It suffices to verify the conditions of using Theorem \ref{Thm:Cramer}.

First, since $y_{ij}^{(l)}$ and $\sign(y_{ij}^{(l)})$ are both bounded, we have 
\begin{align*}
    \mathbb{E}(e^{-\lambda a_{ij}^{(l)}y_{ij}^{(l)} })&=
    p \mathbb{E}(e^{-\lambda y_{ij}^{(l)} })+(1-p) <\infty,  \\
    \mathbb{E}(e^{-\lambda a_{ij}^{(l)}\sign(y_{ij}^{(l)}) })&=
    p \mathbb{E}(e^{-\lambda \sign(y_{ij}^{(l)} )})+(1-p) <\infty.
\end{align*}
Therefore, by Lemma \ref{Lemma:RateFunction}, we have
\begin{align*}
   & \lim_{L\rightarrow\infty}\frac{1}{L}
    \log\left[
    \mathbb{P}\left(-\sum_{l=1}^L a_{ij}^{(l)}y_{ij}^{(l)} \geq 0\right) \right]\\
    = & -\log\left(\frac{\cosh(\phi(\gamma_{ij}^\star))}{p+(1-p)\cosh(\phi(\gamma_{ij}^\star))}\right) 
+ \inf_{\lambda \in \mathbb{R}} \log \left(\frac{ \frac{p\sum_{k = 1}^K e^{\psi_{\gamma_{ij}^\star}(k)} \cosh\left( \phi(\gamma_{ij}^\star) + \lambda k \right)}{\sum_{k = 1}^K e^{\psi_{\gamma_{ij}^\star}(k)}}+(1-p) \cosh\left( \phi(\gamma_{ij}^\star) \right)}{p+(1-p)\cosh(\phi(\gamma_{ij}^\star))}\right) \\
\triangleq &-I_1<0,
\end{align*}
and 
\begin{align*}
\lim_{L\rightarrow\infty}\frac{1}{L}
    \log\left[
    \mathbb{P}\left(-\sum_{l=1}^L a_{ij}^{(l)}\sign(y_{ij}^{(l)}) \geq 0\right) \right] = -\log\left(\frac{\cosh(\phi(\gamma_{ij}^\star))}{p+(1-p)\cosh(\phi(\gamma_{ij}^\star))}\right) 
    \triangleq -I_2 . 
\end{align*}
This indicates that
\begin{align*}
      \mathbb{P}\left(-\sum_{l=1}^L a_{ij}^{(l)}y_{ij}^{(l)} \geq 0\right) \xrightarrow{L\rightarrow\infty} 0 \mbox{ and }
  \mathbb{P}\left(-\sum_{l=1}^L a_{ij}^{(l)}\sign(y_{ij}^{(l)}) \geq 0\right) \xrightarrow{L\rightarrow\infty} 0.
\end{align*}
There exists a large $C$ such that for $L \geq C$
\begin{align*}
    \log\left[
    \mathbb{P}\left(-\sum_{l=1}^L a_{ij}^{(l)}y_{ij}^{(l)} \geq 0\right) \right]  \geq L(-I_1 -\varepsilon) \mbox{ and }
    \log\left[
    \mathbb{P}\left(-\sum_{l=1}^L a_{ij}^{(l)}\sign(y_{ij}^{(l)}) \geq 0\right) \right]  \leq L(-I_2 +\varepsilon).
\end{align*}
Since $I_1 < I_2$, we can select $\varepsilon$ sufficiently small so that, for all $L \geq C$,
\begin{align*}
\frac{\mathbb{P}\left(-\sum_{l=1}^L a_{ij}^{(l)}\sign(y_{ij}^{(l)}) \geq 0\right)}{\mathbb{P}\left(-\sum_{l=1}^L a_{ij}^{(l)}y_{ij}^{(l)} \geq 0\right) } \leq \frac{\exp(L(-I_2+\varepsilon))}{\exp(L(-I_1-\varepsilon))}=
\exp(-L(I_2-I_1-2\varepsilon))\xrightarrow{L\rightarrow \infty}0.
\end{align*}
This completes the proof. \hfill ${\color{red}\blacksquare}$\\

\bibliographystyle{authoryear}

\begin{thebibliography}{}

\bibitem[Agresti, 1992]{agresti1992analysis}
Agresti, A. (1992).
\newblock Analysis of ordinal paired comparison data.
\newblock {\em Journal of the Royal Statistical Society: Series C (Applied Statistics)}, 41(2):287--297.

\bibitem[B{\"o}ckenholt and Dillon, 1997]{bockenholt1997some}
B{\"o}ckenholt, U. and Dillon, W.~R. (1997).
\newblock Some new methods for an old problem: Modeling preference changes and competitive market structures in pretest market data.
\newblock {\em Journal of Marketing Research}, 34(1):130--142.

\bibitem[Bradley and Terry, 1952]{bradley1952rank}
Bradley, R.~A. and Terry, M.~E. (1952).
\newblock Rank analysis of incomplete block designs: I. the method of paired comparisons.
\newblock {\em Biometrika}, 39(3/4):324--345.

\bibitem[Buhlmann and Huber, 1963]{buhlmann1963pairwise}
Buhlmann, H. and Huber, P.~J. (1963).
\newblock Pairwise comparison and ranking in tournaments.
\newblock {\em The Annals of Mathematical Statistics}, 34(2):501--510.

\bibitem[Busa-Fekete et~al., 2013]{busa2013top}
Busa-Fekete, R., Szorenyi, B., Cheng, W., Weng, P., and H{\"u}llermeier, E. (2013).
\newblock Top-k selection based on adaptive sampling of noisy preferences.
\newblock In {\em International Conference on Machine Learning}, pages 1094--1102. PMLR.

\bibitem[Caron et~al., 2014]{caron2014bayesian}
Caron, F., Teh, Y.~W., and Murphy, B. (2014).
\newblock Bayesian nonparametric plackett-luce models for the analysis of clustered ranked data.
\newblock {\em Annal of Applied Statistics}, 8:1145--1181.

\bibitem[Chen et~al., 2022a]{chen2022optimal}
Chen, P., Gao, C., and Zhang, A.~Y. (2022a).
\newblock Optimal full ranking from pairwise comparisons.
\newblock {\em The Annals of Statistics}, 50(3):1775--1805.

\bibitem[Chen et~al., 2022b]{chen2022partial}
Chen, P., Gao, C., and Zhang, A.~Y. (2022b).
\newblock Partial recovery for top-k ranking: optimality of mle and suboptimality of the spectral method.
\newblock {\em The Annals of Statistics}, 50(3):1618--1652.

\bibitem[Chen et~al., 2019]{chen2019spectral}
Chen, Y., Fan, J., Ma, C., and Wang, K. (2019).
\newblock Spectral method and regularized mle are both optimal for top-k ranking.
\newblock {\em Annals of statistics}, 47(4):2204.

\bibitem[Cui et~al., 2023]{cui2023ultrafeedback}
Cui, G., Yuan, L., Ding, N., Yao, G., He, B., Zhu, W., Ni, Y., Xie, G., Xie, R., Lin, Y., et~al. (2023).
\newblock Ultrafeedback: Boosting language models with scaled ai feedback.
\newblock {\em arXiv preprint arXiv:2310.01377}.

\bibitem[Davidson, 1970]{davidson1970extending}
Davidson, R.~R. (1970).
\newblock On extending the bradley-terry model to accommodate ties in paired comparison experiments.
\newblock {\em Journal of the American Statistical Association}, 65(329):317--328.

\bibitem[Duineveld et~al., 2000]{duineveld2000log}
Duineveld, C., Arents, P., and King, B.~M. (2000).
\newblock Log-linear modelling of paired comparison data from consumer tests.
\newblock {\em Food Quality and Preference}, 11(1-2):63--70.

\bibitem[Fan et~al., 2025]{fan2025ranking}
Fan, J., Lou, Z., Wang, W., and Yu, M. (2025).
\newblock Ranking inferences based on the top choice of multiway comparisons.
\newblock {\em Journal of the American Statistical Association}, 120(549):237--250.

\bibitem[Glenn and David, 1960]{glenn1960ties}
Glenn, W.~A. and David, H.~A. (1960).
\newblock Ties in paired-comparison experiments using a modified thurstone-mosteller model.
\newblock {\em Biometrics}, 16(1):86--109.

\bibitem[Han et~al., 2022]{han2022general}
Han, R., Xu, Y., and Chen, K. (2022).
\newblock A general pairwise comparison model for extremely sparse networks.
\newblock {\em Journal of the American Statistical Association}, pages 1--11.

\bibitem[Harper and Konstan, 2015]{harper2015movielens}
Harper, F.~M. and Konstan, J.~A. (2015).
\newblock The movielens datasets: History and context.
\newblock {\em Acm transactions on interactive intelligent systems (tiis)}, 5(4):1--19.

\bibitem[Jadbabaie et~al., 2020]{jadbabaie2020estimation}
Jadbabaie, A., Makur, A., and Shah, D. (2020).
\newblock Estimation of skill distribution from a tournament.
\newblock {\em Advances in Neural Information Processing Systems}, 33:8418--8429.

\bibitem[Kendall, 1938]{kendall1938new}
Kendall, M.~G. (1938).
\newblock A new measure of rank correlation.
\newblock {\em Biometrika}, 30(1-2):81--93.

\bibitem[Li et~al., 2022]{li2022detecting}
Li, W., Rinaldo, A., and Wang, D. (2022).
\newblock Detecting abrupt changes in sequential pairwise comparison data.
\newblock {\em Advances in Neural Information Processing Systems}, 35:37851--37864.

\bibitem[Liu et~al., 2023]{liu2023lagrangian}
Liu, Y., Fang, E.~X., and Lu, J. (2023).
\newblock Lagrangian inference for ranking problems.
\newblock {\em Operations research}, 71(1):202--223.

\bibitem[Poddar et~al., 2024]{poddar2024personalizing}
Poddar, S., Wan, Y., Ivison, H., Gupta, A., and Jaques, N. (2024).
\newblock Personalizing reinforcement learning from human feedback with variational preference learning.
\newblock {\em arXiv preprint arXiv:2408.10075}.

\bibitem[Rao and Kupper, 1967]{rao1967ties}
Rao, P. and Kupper, L.~L. (1967).
\newblock Ties in paired-comparison experiments: A generalization of the bradley-terry model.
\newblock {\em Journal of the American Statistical Association}, 62(317):194--204.

\bibitem[Shah and Wainwright, 2018]{shah2018simple}
Shah, N.~B. and Wainwright, M.~J. (2018).
\newblock Simple, robust and optimal ranking from pairwise comparisons.
\newblock {\em Journal of Machine Learning Research}, 18(199):1--38.

\bibitem[Slocum et~al., 2025]{slocum2025diverse}
Slocum, S., Parker-Sartori, A., and Hadfield-Menell, D. (2025).
\newblock Diverse preference learning for capabilities and alignment.
\newblock In {\em The Thirteenth International Conference on Learning Representations}.

\bibitem[Stern, 1990]{stern1990continuum}
Stern, H. (1990).
\newblock A continuum of paired comparisons models.
\newblock {\em Biometrika}, 77(2):265--273.

\bibitem[Stern, 2011]{stern2011moderated}
Stern, S.~E. (2011).
\newblock Moderated paired comparisons: a generalized bradley--terry model for continuous data using a discontinuous penalized likelihood function.
\newblock {\em Journal of the Royal Statistical Society Series C: Applied Statistics}, 60(3):397--415.

\bibitem[Thurstone, 1994]{thurstone1994law}
Thurstone, L.~L. (1994).
\newblock A law of comparative judgment.
\newblock {\em Psychological Review}, 101(2):266.

\bibitem[Wauthier et~al., 2013]{wauthier2013efficient}
Wauthier, F., Jordan, M., and Jojic, N. (2013).
\newblock Efficient ranking from pairwise comparisons.
\newblock In {\em International Conference on Machine Learning}, pages 109--117. PMLR.

\bibitem[Xu et~al., 2025]{xu2025rate}
Xu, S., Sun, W.~W., and Cheng, G. (2025).
\newblock Rate-optimal rank aggregation with private pairwise rankings.
\newblock {\em Journal of the American Statistical Association}, pages 1--14.

\bibitem[Zhu et~al., 2023]{zhu2023principled}
Zhu, B., Jordan, M., and Jiao, J. (2023).
\newblock Principled reinforcement learning with human feedback from pairwise or k-wise comparisons.
\newblock In {\em International Conference on Machine Learning}, pages 43037--43067. PMLR.

\end{thebibliography}


\begin{thebibliography}{}

\bibitem[Chen et~al., 2019]{chen2019spectral}
Chen, Y., Fan, J., Ma, C., and Wang, K. (2019).
\newblock Spectral method and regularized mle are both optimal for top-k ranking.
\newblock {\em Annals of statistics}, 47(4):2204.

\bibitem[Davidson, 1970]{davidson1970extending}
Davidson, R.~R. (1970).
\newblock On extending the bradley-terry model to accommodate ties in paired comparison experiments.
\newblock {\em Journal of the American Statistical Association}, 65(329):317--328.

\bibitem[Gao et~al., 2023]{gao2023uncertainty}
Gao, C., Shen, Y., and Zhang, A.~Y. (2023).
\newblock Uncertainty quantification in the bradley--terry--luce model.
\newblock {\em Information and Inference: A Journal of the IMA}, 12(2):1073--1140.

\bibitem[Glenn and David, 1960]{glenn1960ties}
Glenn, W.~A. and David, H.~A. (1960).
\newblock Ties in paired-comparison experiments using a modified thurstone-mosteller model.
\newblock {\em Biometrics}, 16(1):86--109.

\bibitem[Klenke, 2013]{klenke2013probability}
Klenke, A. (2013).
\newblock {\em Probability theory: a comprehensive course}.
\newblock Springer Science \& Business Media.

\bibitem[Negahban et~al., 2012]{negahban2012iterative}
Negahban, S., Oh, S., and Shah, D. (2012).
\newblock Iterative ranking from pair-wise comparisons.
\newblock {\em Advances in neural information processing systems}, 25.

\bibitem[Rao and Kupper, 1967]{rao1967ties}
Rao, P. and Kupper, L.~L. (1967).
\newblock Ties in paired-comparison experiments: A generalization of the bradley-terry model.
\newblock {\em Journal of the American Statistical Association}, 62(317):194--204.

\bibitem[Shah and Wainwright, 2018]{shah2018simple}
Shah, N.~B. and Wainwright, M.~J. (2018).
\newblock Simple, robust and optimal ranking from pairwise comparisons.
\newblock {\em Journal of Machine Learning Research}, 18(199):1--38.

\end{thebibliography}
\putbib[Ref]
\end{bibunit}
\end{document}